\documentclass[journal]{IEEEtran}
\usepackage{times}
\usepackage{epsfig}
\usepackage{graphicx}
\usepackage{amsmath}
\usepackage{array}
\usepackage{graphicx}
\usepackage{amssymb}
\usepackage{multirow}
\usepackage{adjustbox}
\usepackage{tabularx}
\usepackage{xcolor,colortbl}
\usepackage{booktabs}
\usepackage{xspace}
\usepackage{url}

\usepackage{algorithmic}
\usepackage[ruled, linesnumbered, noend]{algorithm2e}
\usepackage{xcolor}

\usepackage{amssymb}% http://ctan.org/pkg/amssymb
\usepackage{pifont}% http://ctan.org/pkg/pifont
\usepackage{soul}
\usepackage[caption=false]{subfig}
\usepackage{caption}
\usepackage{enumitem}
\usepackage{footnote}
\usepackage{tablefootnote}
\usepackage{longtable}

\newcommand{\blue}[1]{\textcolor{blue}{#1}}
\newcommand{\red}[1]{\textcolor{red}{#1}}

\newcommand{\cmark}{\ding{51}}%

\makeatletter
\DeclareRobustCommand\onedot{\futurelet\@let@token\@onedot}
\def\@onedot{\ifx\@let@token.\else.\null\fi\xspace}

\def\etal{\emph{et al}\onedot}

\definecolor{Gray}{gray}{0.9}
\definecolor{GrayLine}{gray}{0.7}

\makeatother
\usepackage[noadjust]{cite}

\usepackage{tcolorbox,varwidth}
\newtcbox{\tightredbox}{ colframe=red,colback=white,arc=0.5pt,boxsep=0pt,left=0pt,right=0pt,top=0pt,bottom=0pt,boxrule=0.5pt,varwidth upper}
\newtcbox{\tightbluebox}{ colframe=blue,colback=white,arc=0.5pt,boxsep=0pt,left=0pt,right=0pt,top=0pt,bottom=0pt,boxrule=0.5pt,varwidth upper}

\usepackage[linewidth=1pt]{mdframed}
\mdfsetup{skipabove=0pt,skipbelow=1pt}
\mdfdefinestyle{RedBox}{%
    linecolor=red,
    innerlinewidth=3pt,
    innertopmargin=0pt,
    innerbottommargin=0pt,
    innerrightmargin=0pt,
    innerleftmargin=0pt,
    leftmargin = 0pt,
    rightmargin = 0pt,
    outermargin = 0pt,
        }
        
\mdfdefinestyle{BlueBox}{%
    linecolor=blue,
    innerlinewidth=3pt,
    innertopmargin=0pt,
    innerbottommargin=0pt,
    innerrightmargin=0pt,
    innerleftmargin=0pt,
    leftmargin = 0pt,
    rightmargin = 0pt,
    outermargin = 0pt,
        }
\ifCLASSINFOpdf
\else
\fi
\begin{document}
\title{A Multi-purpose Realistic Haze Benchmark with Quantifiable Haze Levels and Ground Truth}

\author{Priya~Narayanan, %~\IEEEmembership{Member,~IEEE,}
        Xin~Hu*, %~\IEEEmembership{Member,~IEEE,}
        Zhenyu~Wu*, %~\IEEEmembership{Member,~IEEE,}
        Matthew~D~Thielke, %~\IEEEmembership{Member,~IEEE,}
        John~G~Rogers, %~\IEEEmembership{Senior Member,~IEEE,}
        Andre~V~Harrison, %~\IEEEmembership{Member,~IEEE,}
        John A D’Agostino,
        James D Brown,
        Long~P~Quang, %~\IEEEmembership{Member,~IEEE,}
        James R Uplinger, %~\IEEEmembership{Member,~IEEE,}
        Heesung~Kwon, %~\IEEEmembership{Member,~IEEE,}
        and~Zhangyang~Wang%~\IEEEmembership{Member,~IEEE,}% <-this % stops a space
%\thanks{M. Shell was with the Department
%of Electrical and Computer Engineering, Georgia %Institute of Technology, Atlanta,
%GA, 30332 USA e-mail: (see %http://www.michaelshell.org/contact.html).}% <-this % %stops a space
%\thanks{J. Doe and J. Doe are with Anonymous %University.}% <-this % stops a space
%\thanks{Manuscript received April 19, 2005; revised %August 26, 2015.}}

\thanks{P. Narayanan, M. D. Thielke, J. G. Rogers, A. V. Harrison, L. D. Quang, J. R. Uplinger and H. Kwon are with DEVCOM Army Research Laboratory, Adelphi,
MD, 20783 USA e-mail: priya.narayanan.civ@mail.mil}% <-this % stops a space
\thanks{X. Hu is with Tulane University, New Orleans, LA 70118}% <-this % stops a space
\thanks{Z. Wu is with Texas A \& M University, College Station, TX 77843 }% <-this % stops a space
\thanks{X. Hu and Z. Wu have contributed equally to this paper}% <-this % stops a space
\thanks{Z. Wang is with The University of Texas at Austin, Austin, TX 78712 }% <-this % stops a space
\thanks{J. A. D'Agostino and J.D Brown are with DEVCOM Chemical and Biological Center, Aberdeen Proving Ground, MD 21010 }% <-this % stops a space
\thanks{Manuscript received Sept XX, 2021; revised XX XX, XXX.}}


\markboth{Journal of \LaTeX\ Class Files,~Vol.~14, No.~8, August~2015}%
{Shell \MakeLowercase{\textit{et al.}}: Bare Demo of IEEEtran.cls for IEEE Journals}



\maketitle

\begin{abstract}
Imagery collected from outdoor visual environments is often degraded due to the presence of dense smoke or haze. A key challenge for research in scene understanding in these degraded visual environments (DVE) is the lack of representative benchmark datasets. These datasets are required to evaluate state-of-the-art vision algorithms (e.g., detection and tracking) in degraded settings. 
In this paper, we address some of these limitations by introducing the first realistic hazy image benchmark, from both aerial and ground view, with paired haze-free images, and in-situ haze density measurements. This dataset was produced in a controlled environment with professional smoke generating machines that covered the entire scene, and consists of images captured from the perspective of both an unmanned aerial vehicle (UAV) and an unmanned ground vehicle (UGV). We also evaluate a set of representative state-of-the-art dehazing approaches as well as object detectors on the dataset. The full dataset presented in this paper, including the ground truth object classification bounding boxes and haze density measurements, is provided for the community to evaluate their algorithms at: \textit{\url{https://a2i2-archangel.vision}}. A subset of this dataset has been used for the ``Object Detection in Haze'' Track of CVPR UG2 2022 challenge at \textit{\url{http://cvpr2022.ug2challenge.org/track1.html}}.
\end{abstract}


% \fi
\IEEEpeerreviewmaketitle

\begin{figure*}[!htb]
    \captionsetup[subfigure]{labelformat=empty,justification=centering,farskip=1pt,captionskip=1pt}
    \centering
    \subfloat
    {
       \includegraphics[width=0.235\linewidth]{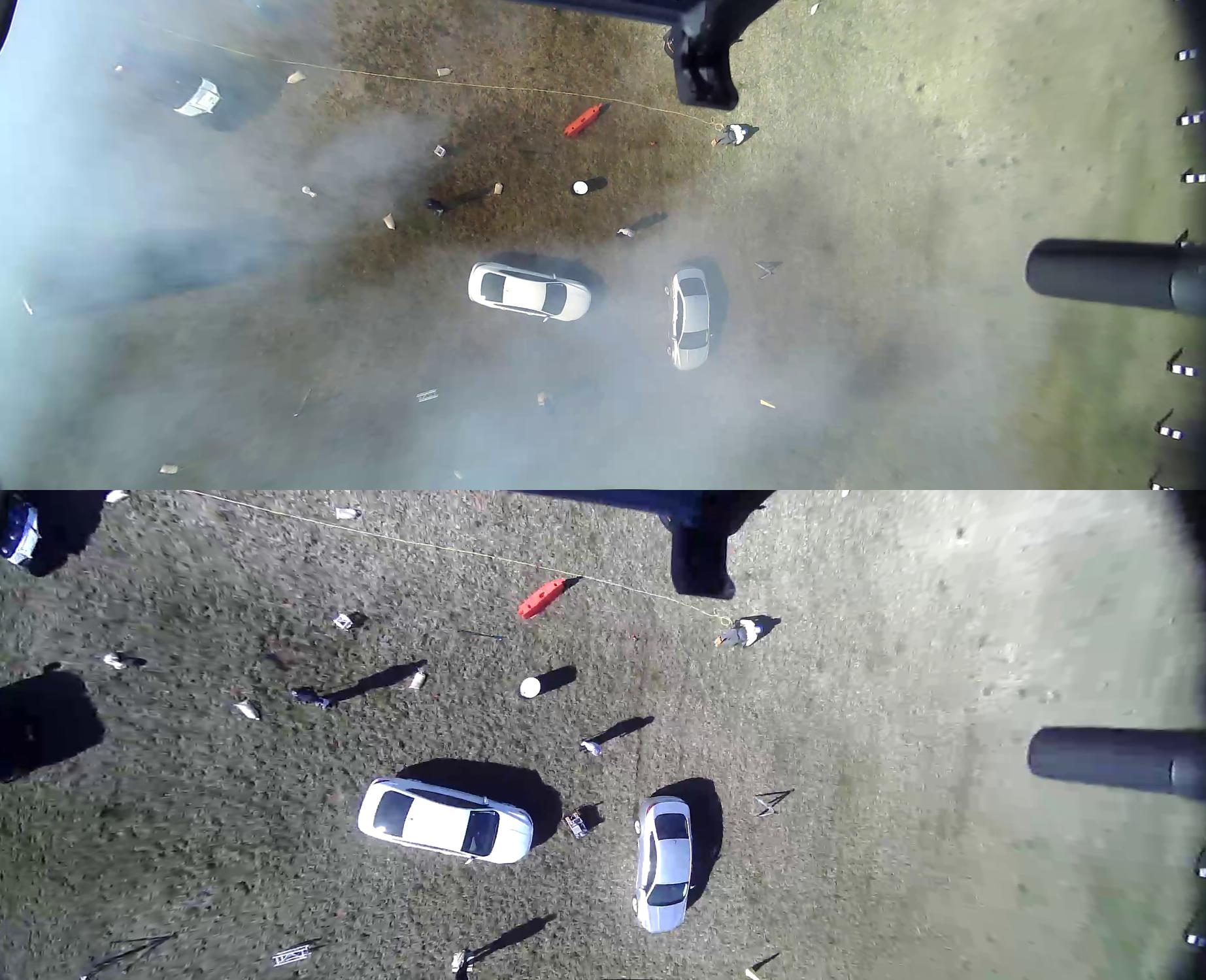}
    }
    \subfloat
    {
       \includegraphics[width=0.235\linewidth]{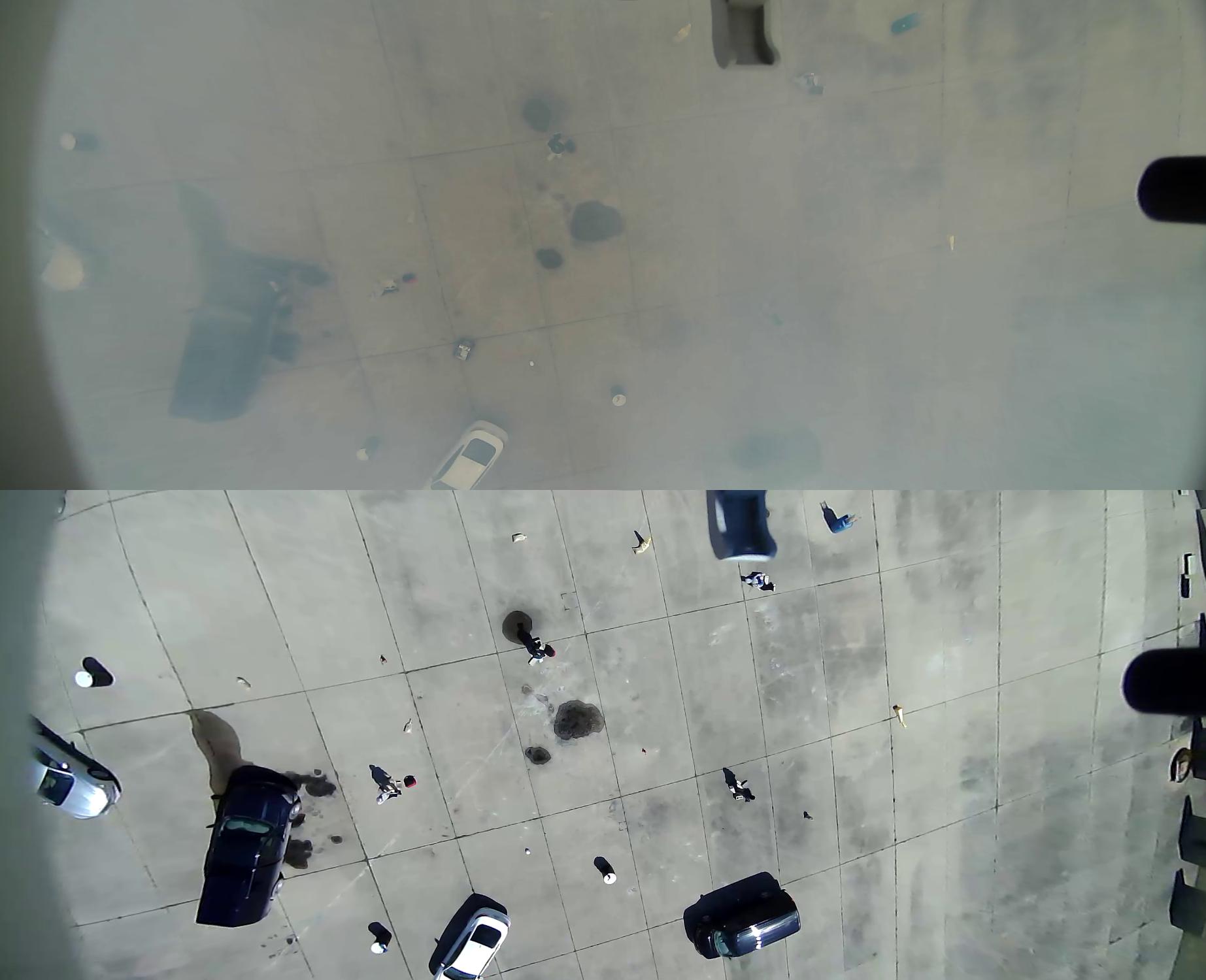}
    }
    \subfloat
    {
       \includegraphics[width=0.235\linewidth]{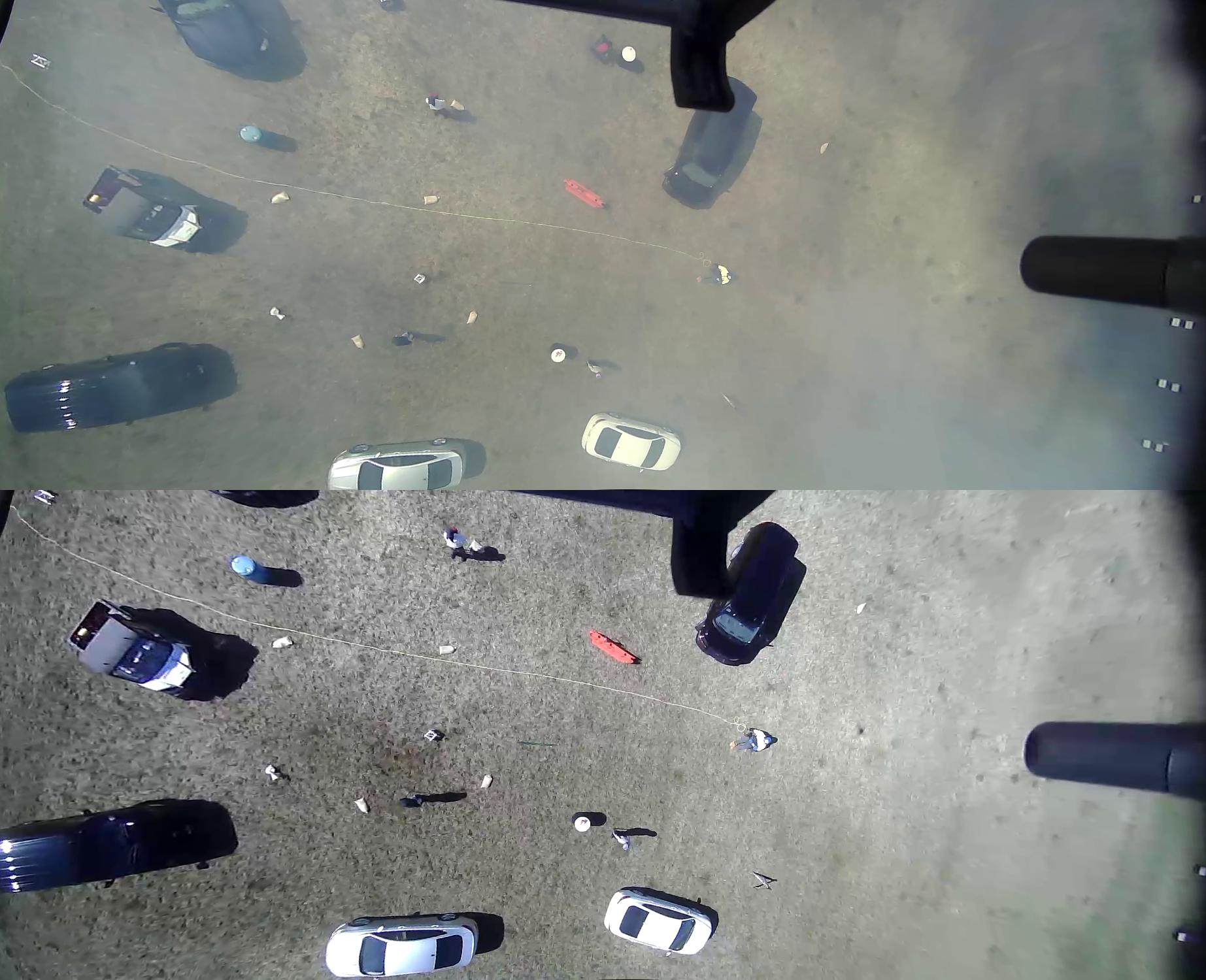}
    }
    \subfloat
    {
       \includegraphics[width=0.235\linewidth]{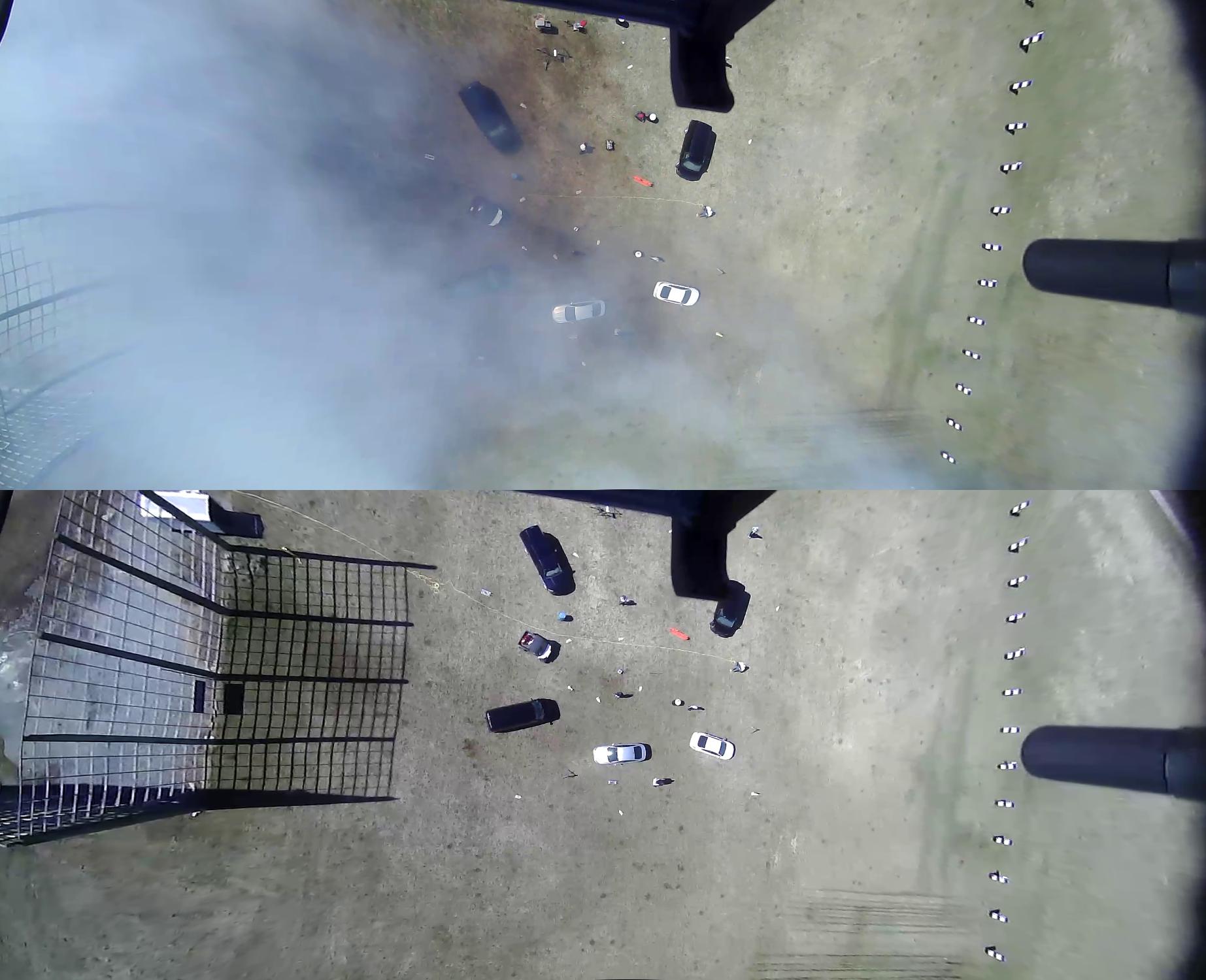}
    }
    \\
    \vspace{+0.5em}
    \subfloat
    {
       \includegraphics[width=0.235\linewidth]{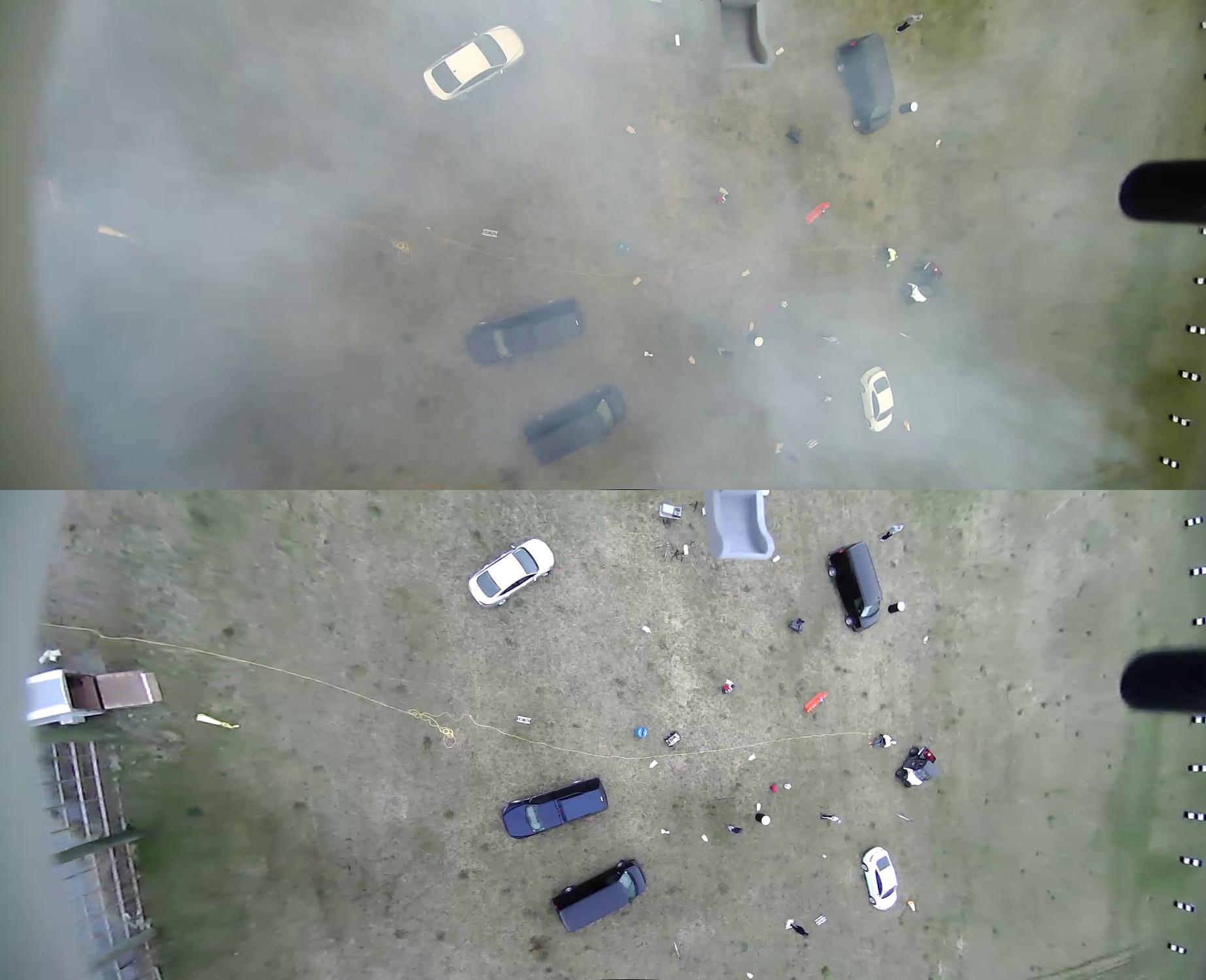}
    }
    \subfloat
    {
       \includegraphics[width=0.235\linewidth]{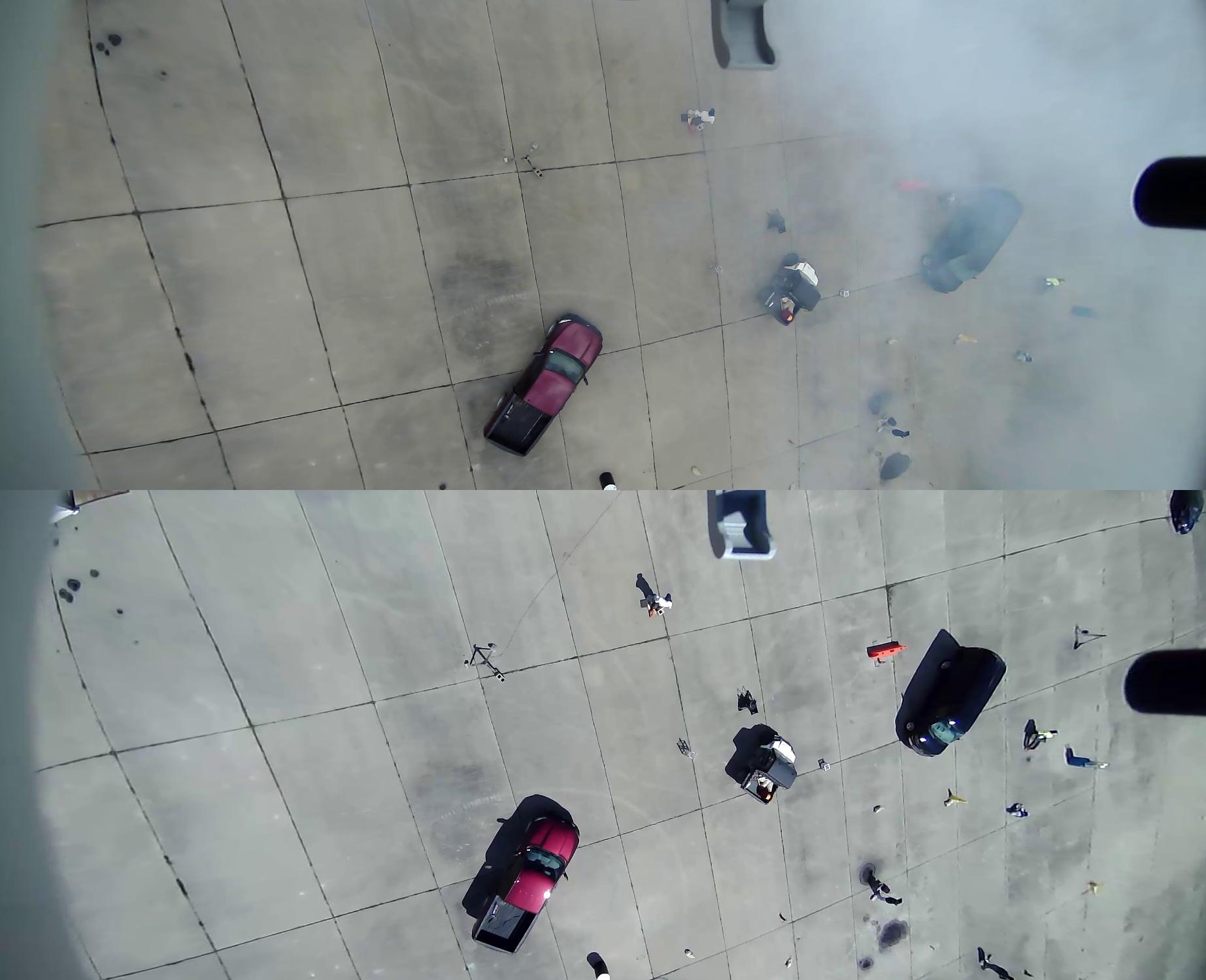}
    }
    \subfloat
    {
       \includegraphics[width=0.235\linewidth]{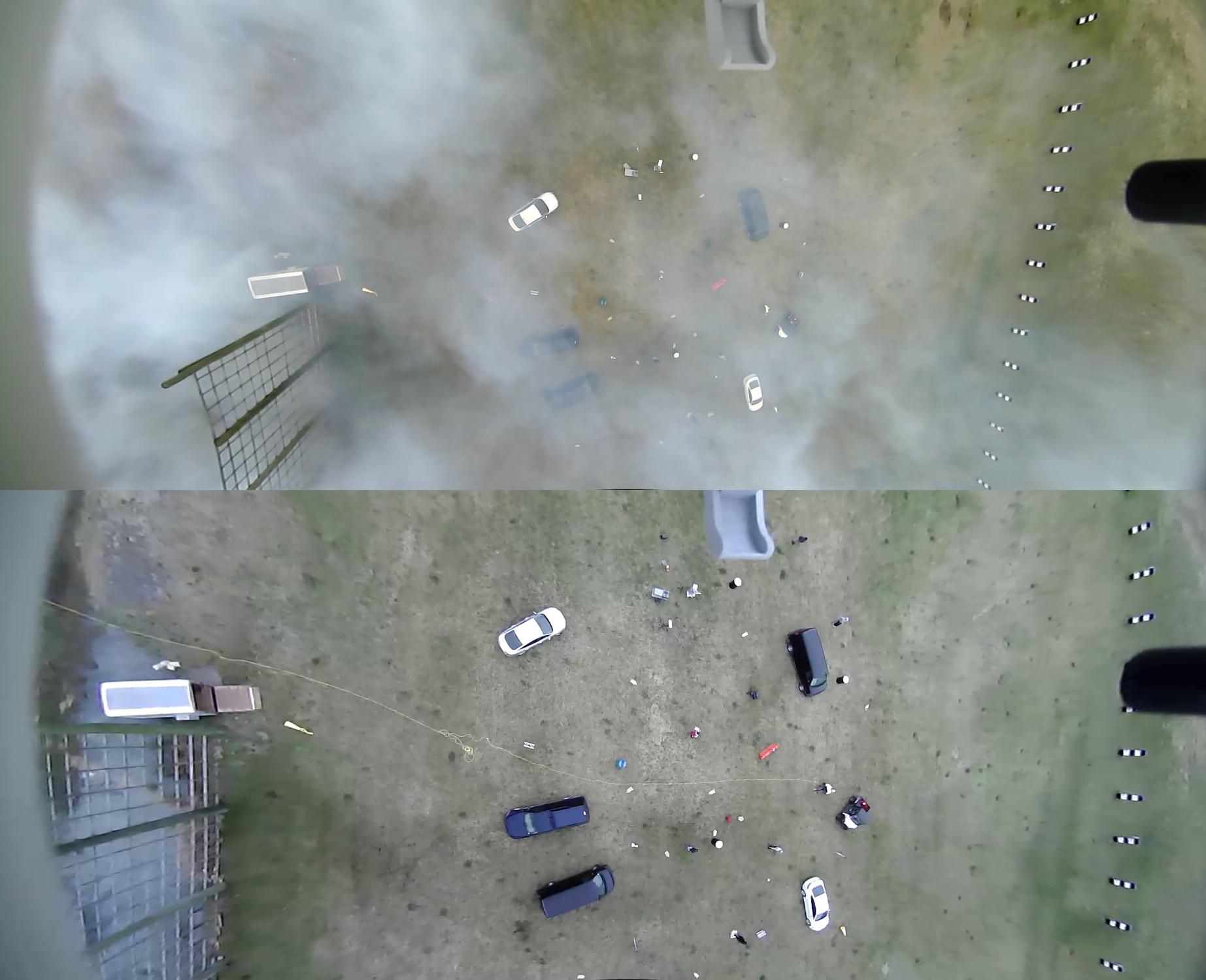}
    }
    \subfloat
    {
       \includegraphics[width=0.235\linewidth]{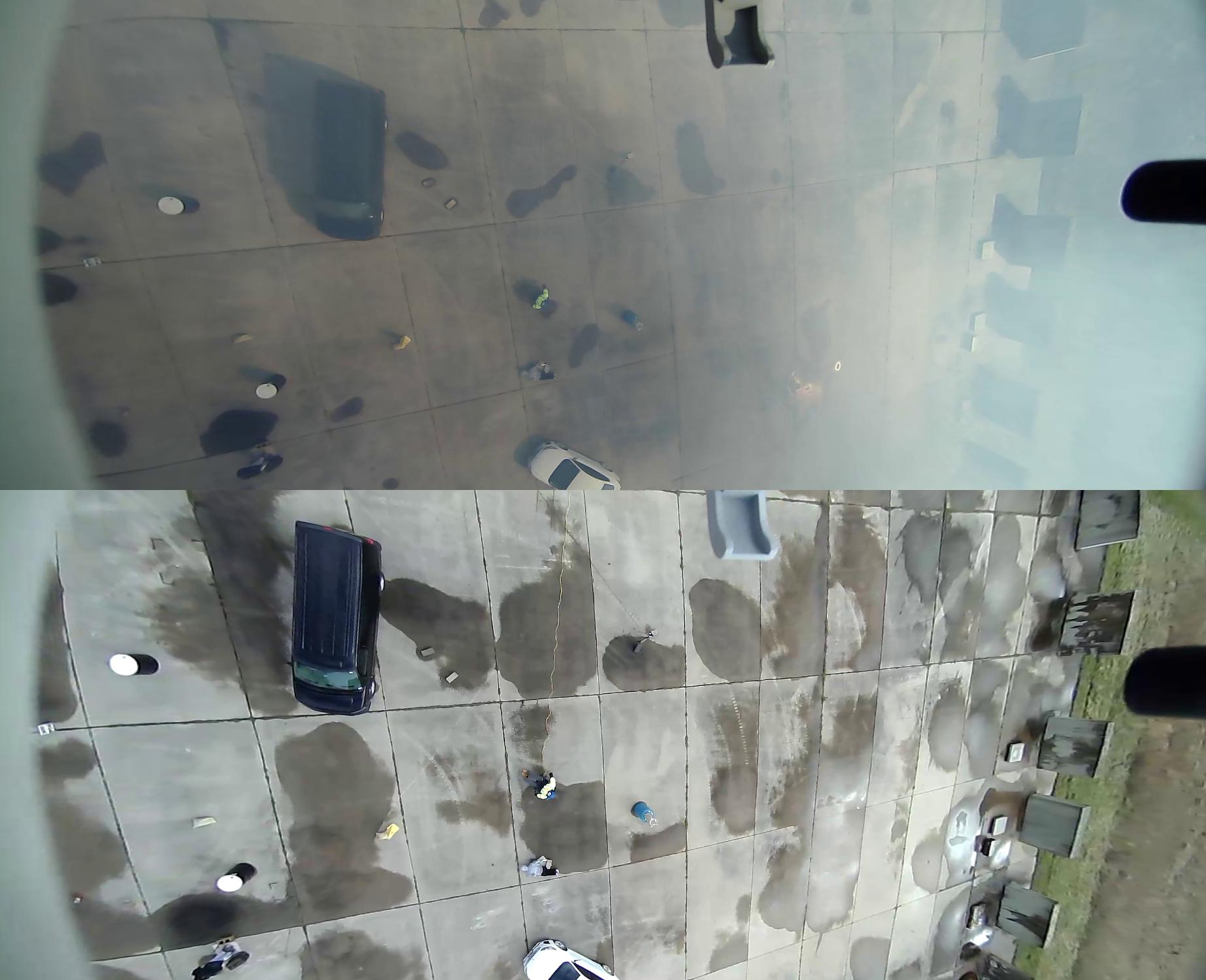}
    }
    \vspace{+0.5em}
    \\
        \subfloat
    {
       \includegraphics[width=0.235\linewidth]{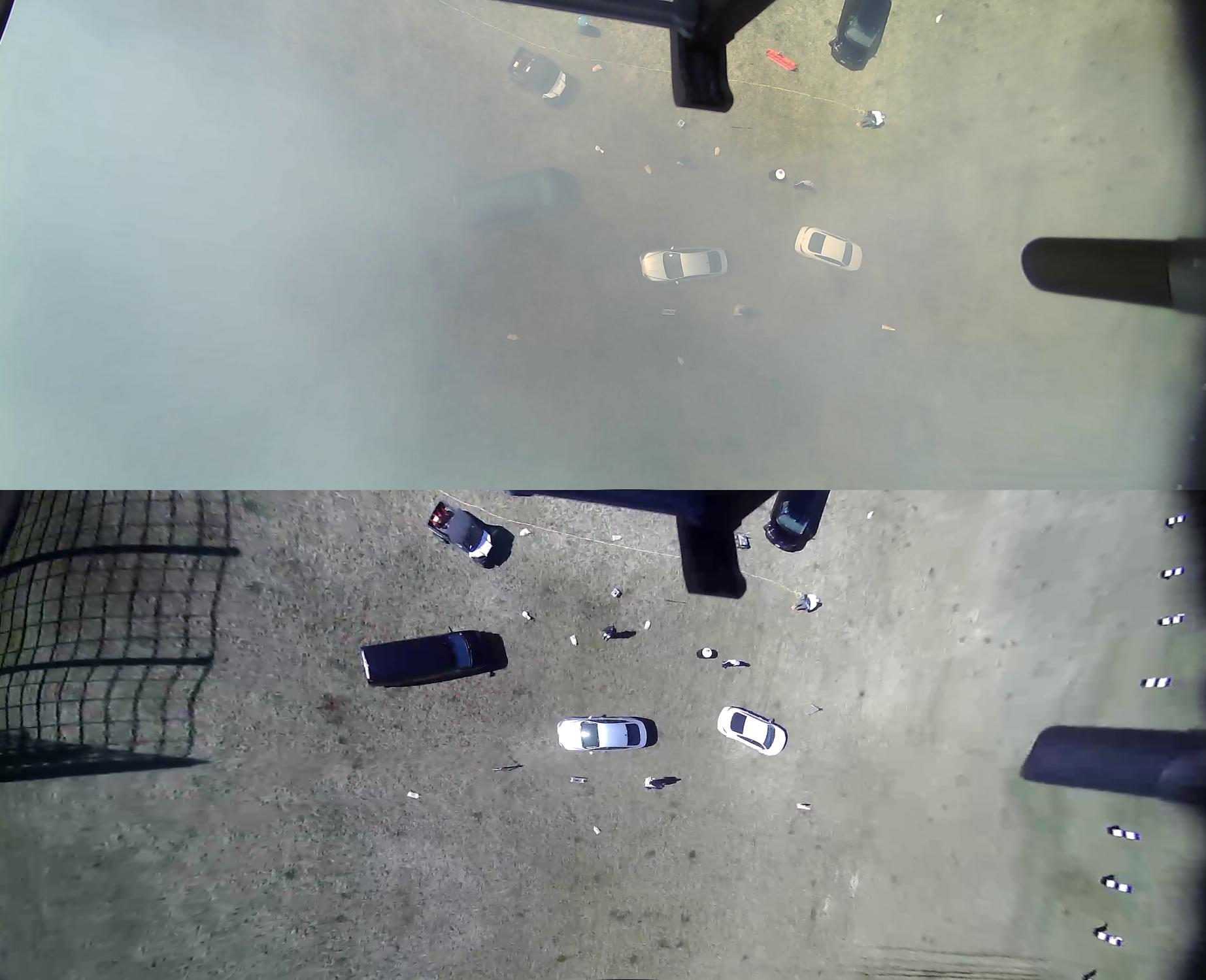}
    }
    \subfloat
    {
       \includegraphics[width=0.235\linewidth]{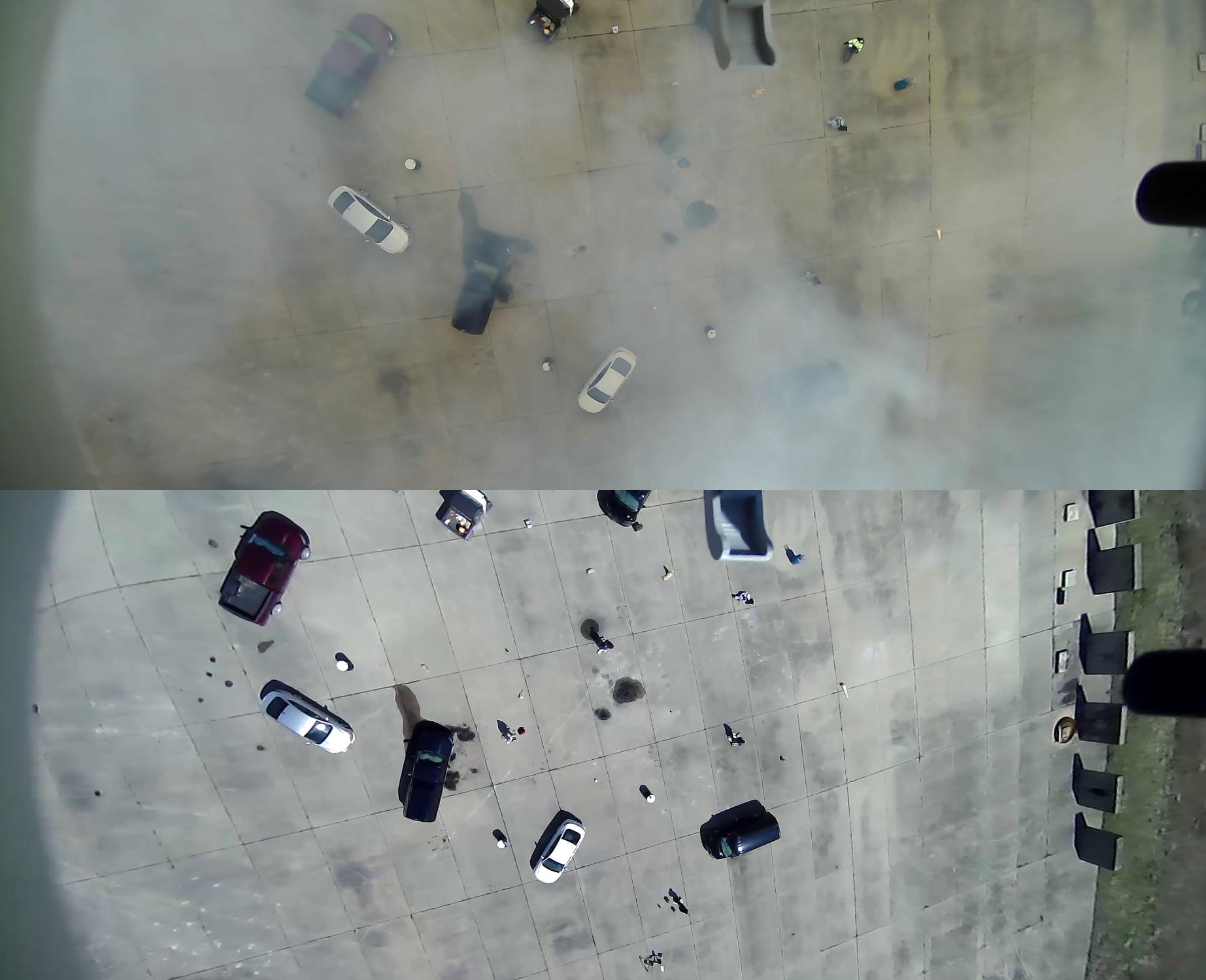}
    }
    \subfloat
    {
       \includegraphics[width=0.235\linewidth]{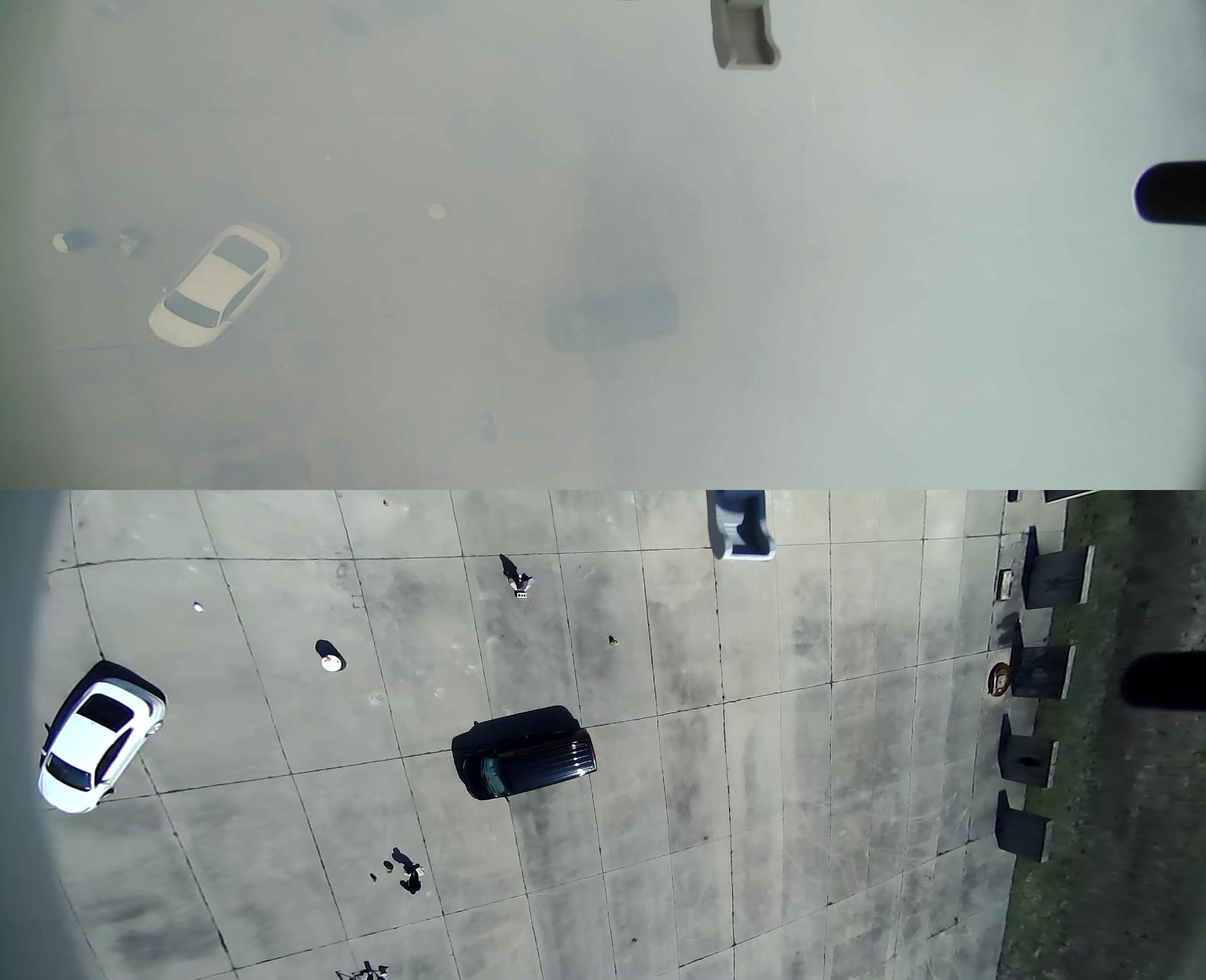}
    }
    \subfloat
    {
       \includegraphics[width=0.235\linewidth]{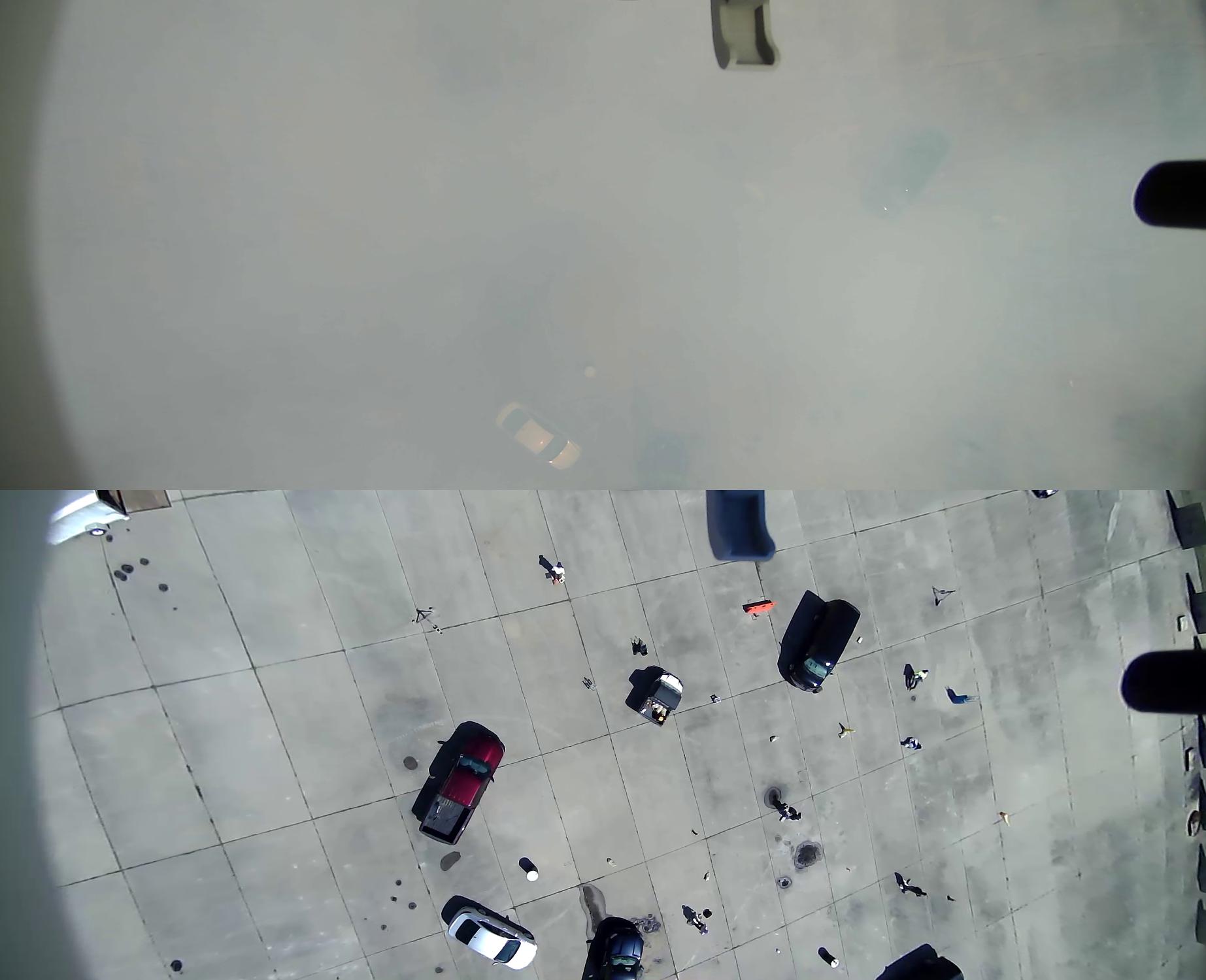}
    }
    \\
    \vspace{+0.5em}
        \subfloat
    {
       \includegraphics[width=0.235\linewidth]{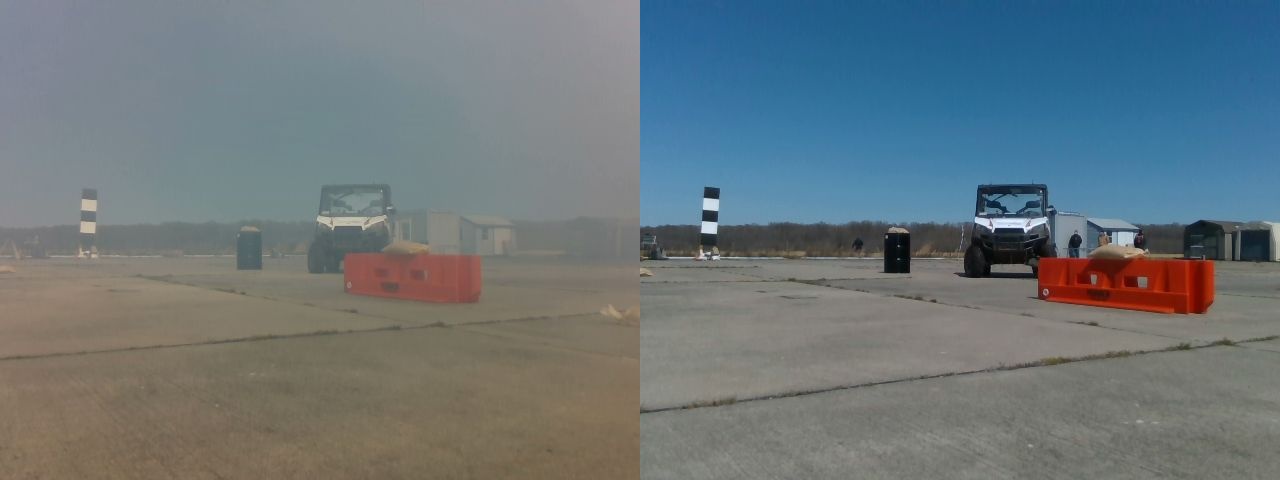}
    }
    \subfloat
    {
       \includegraphics[width=0.235\linewidth]{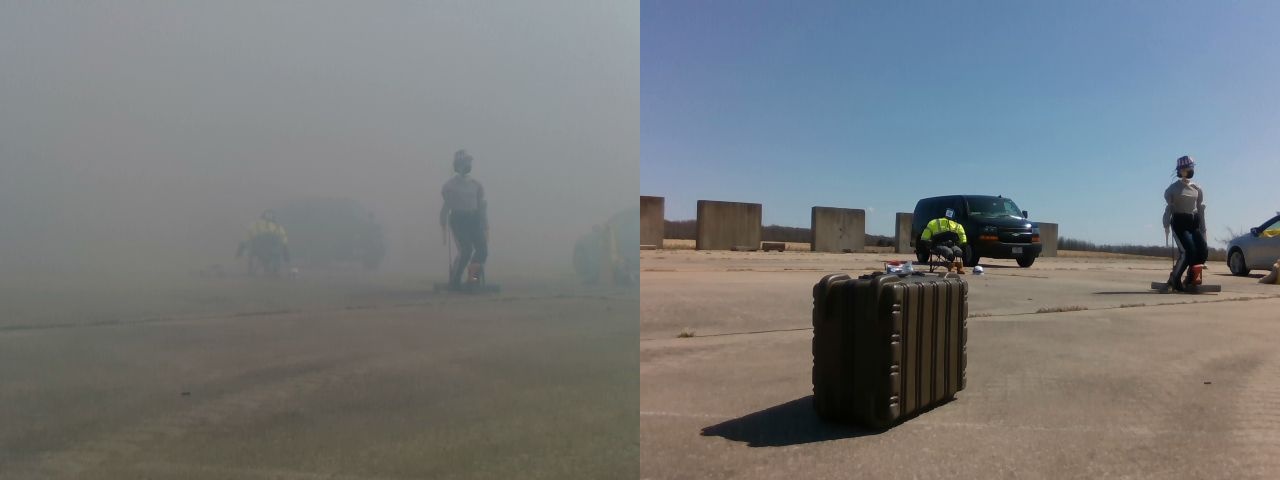}
    }
    \subfloat
    {
       \includegraphics[width=0.235\linewidth]{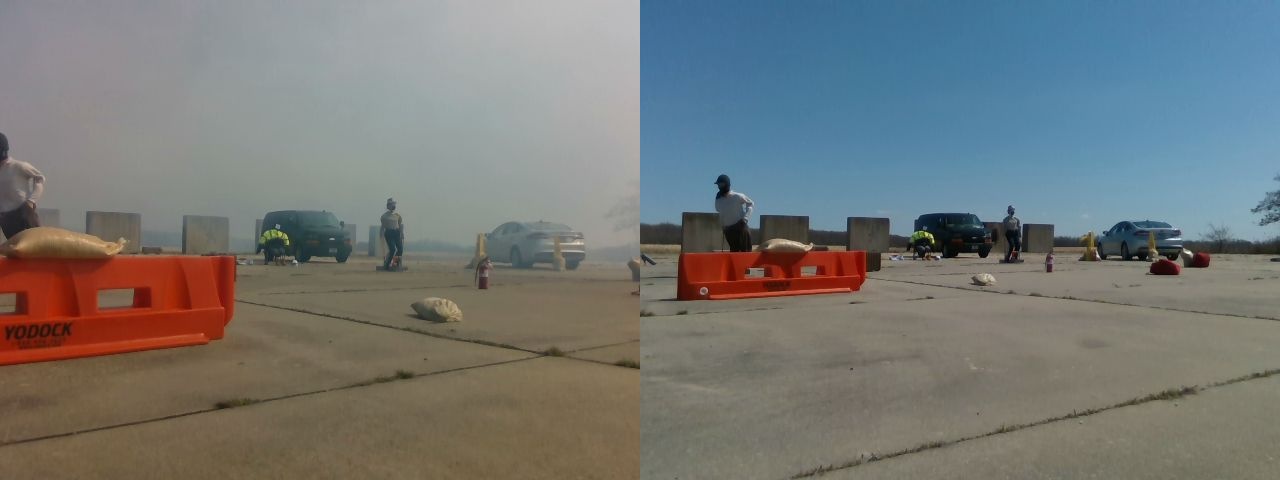}
    }
    \subfloat
    {
       \includegraphics[width=0.235\linewidth]{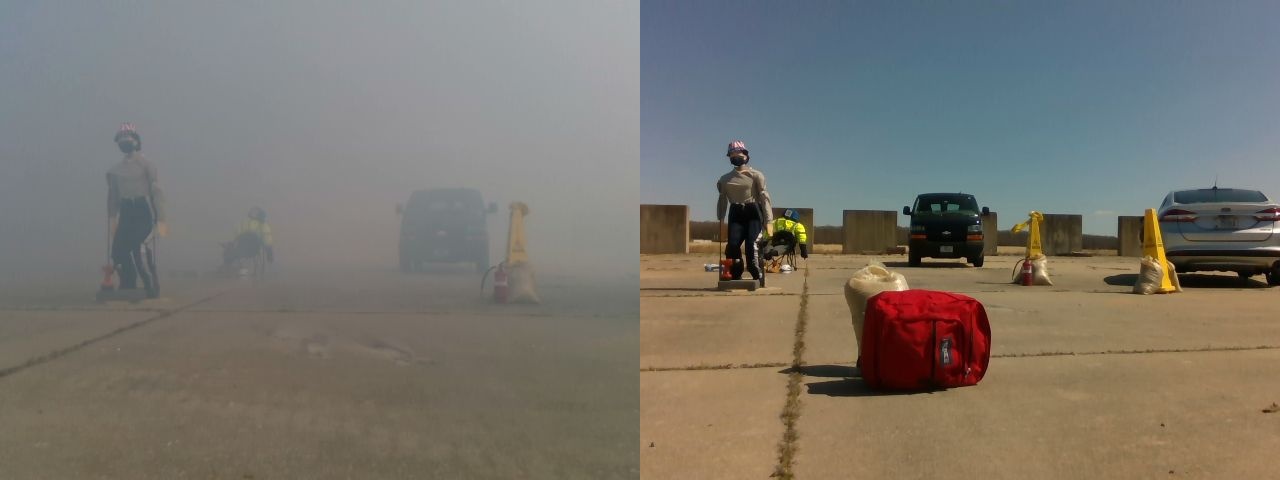}
    }
    \caption{Examples of paired hazy and haze-free images in A2I2-UAV (top three rows) and A2I2-UGV (bottom one row) from A2I2-Haze.
    }
    \label{fig:a2i2_teaser}
\end{figure*}

\section{Introduction }
\IEEEPARstart{S}{cene} understanding in outdoor environments for applications such as intelligence, surveillance, and reconnaissance (ISR), and autonomous vehicle navigation is extremely challenging in the presence of smoke, and other adverse weather conditions due to haze, fog, and mist. These atmospheric phenomena with smoke particles or microscopic water droplets significantly interfere with the operations of onboard vision systems, often resulting in imagery with non-linear noise, blur, reduced contrast levels, and color dimming issues. These visual artifacts, generated from uncontrolled and potentially dynamic outdoor environments or other DVE effects, pose major challenges in many components of semantic scene understanding, including image enhancement, image restoration, object localization, and object classification.  

To address these challenges, a key requirement is benchmarks to accurately evaluate the performance of these algorithms relative to different quantifiable haze levels. This is beyond the reach and scope of most of the existing curated haze datasets such as RESIDE~\cite{li2019benchmarking}, NH-Haze~\cite{NH-Haze_2020}, and REVIDE~\cite{REVIDE}. Furthermore, these datasets are inadequate to quantitatively and fairly compare computer vision algorithms on hazy vs. haze-free imagery on a specific scene. They cannot isolate and measure the effect of haze and have a shortfall in scene object diversity (Discussed in Section II).

In this study, we leverage the US Army’s unique capability to produce and measure smoke/obscurant to generate haze in a controlled fashion. We collect imagery of target objects such as civilian vehicles, mannequins, and man-made obstacles from UAVs and UGVs. We also collect metadata such as altitude and local haze density from moving and stationary sensors. We then develop an image dataset with metadata for realistic, accurate, and fine-grained algorithm evaluation in hazy DVEs. In summary, the contributions of this paper are as follows:

a) We present A2I2-Haze, the first realistic haze dataset with in-situ smoke measurement aligned to aerial and ground imagery. This multi-purpose dataset has paired hazy and haze-free imagery to allow fine-grained evaluation of low-level vision (dehazing) and high-level vision (detection) tasks. Exemplar images are shown in Fig.~\ref{fig:a2i2_teaser}.

b) We conduct a comprehensive study and evaluation on state-of-the-art single image dehazing and object detection algorithms using this non-synthetic benchmark dataset.
\section{Previous Work }
This section summarizes some of the key findings from previous DVE studies. In subsection A, we provide an overview of the hazy DVE datasets that are publicly available for benchmarking the performance of computer vision algorithms. We then survey current state-of-the-art single-image dehazing methods and object detection techniques used for evaluating DVE datasets in subsections B and C, respectively.
\subsection{Haze Datasets}
Several of the currently available hazy DVE datasets (Table 1), such as 3R~\cite{zhang2020nighttime}, HazeRD~\cite{Zhang:HazeRD:ICIP17b}, 4K~\cite{4K_Xiao_2022} and RESIDE~\cite{li2019benchmarking} were synthetically generated using scattering models to simulate hazy conditions. However, while these simulation techniques are useful in generating large-scale training datasets comprising both hazy and reference haze-free images, they often use unrealistic parameters and assumptions such as homogeneity. Further, while synthetically simulating haze ensures all other conditions in the scene are preserved, the technique still poses challenges such as transferring the knowledge to real-world target domains. 

More recently, a few real-world hazy datasets such as MRFID~\cite{liu2020end}, BeDDE~\cite{zhao2020dehazing}, Dense-Haze~\cite{ancuti2019dense}, O-Haze~\cite{O-HAZE_2018}, I-Haze~\cite{I-HAZE_2018}, REVIDE~\cite{REVIDE}, and NH-Haze~\cite{NH-Haze_2020} have been published, with images captured in the presence of haze that is generated using professional-grade haze machines. Among the six datasets listed above, NH-Haze is the most representative of a realistic haze scene with non-homogeneous hazy and haze-free pairs. MRFID, BeDDE, Dense-Haze, NH-Haze and O-Haze are datasets with outdoor scenes that make them relevant to ISR and autonomous UGV applications. However, these datasets also pose several limitations. The smoke generators and the sensors for data collection in the MRFID, BeDDE, Dense-Haze, and (NH, O, and I)-Haze datasets are typically mounted in a fixed location and do not capture the spatial variability of the scene. Furthermore, all these datasets are limited to ground view imagery. Moreover, they do not provide a measurement of haze or smoke transmissibility for training and evaluation. The main distinction between our A2I2-Haze dataset from previous efforts is that our dataset provides ground and aerial imagery with haze-free reference images taken from mobile sensors in an outdoor environment. We also provide highly synchronized qualitative and quantitative measurements of smoke transmissibility, using human assessments and in-situ measurements from ground sensors. This will be discussed in detail in subsequent sections.
\begin{table}[ht]
\caption{Properties of A2I2-Haze relative to other public dehazing benchmarks. A/G stands for aerial-/ground-view. In/Out stands for in/outdoor. NH stands for non-homogeneous. HM stands for haze measurement. Details of A2I2-Haze are in the first paragraph of Section IV.}
\centering
\resizebox{\columnwidth}{!}{
\begin{tabular}{c|c|c|c|c|c|c}
\hline
\multirow{2}{*}{Datasets}  &   \multicolumn{6}{c}{Attributes}  \\
\cline{2-7}
& A/G & In/Out & NH & Syn/Real & \#Images & HM  \\
\hline
3R~\cite{zhang2020nighttime} & G & O & & S & $2,750$ & \\
HazeRD~\cite{Zhang:HazeRD:ICIP17b} & G & O & & S & $33$ & \\
O-Haze~\cite{O-HAZE_2018} & G & O & & R & $45$ & \\
I-Haze~\cite{I-HAZE_2018} & G & I & & R & $35$ & \\
4K~\cite{4K_Xiao_2022} & G & O & & S & $10,000$ & \\ 
REVIDE~\cite{REVIDE} & G & I & & R & 1982 & \\
RESIDE~\cite{li2019benchmarking} & G & I+O & & S+R & $100,076$ & \\
BeDDE~\cite{zhao2020dehazing} & G & O & & R & $208$ & \\
MRFID~\cite{liu2020end} & G & O & & R & $800$\tablefootnote{MRFID contains images from 200 clear outdoor scenes. For each of the clear images, there are 4 images of the same scene containing different densities.} & \\
Dense-Haze~\cite{ancuti2019dense} & G & O & & R & $33$ & \\
NH-Haze~\cite{NH-Haze_2020} & G & O & \cmark & R & $55$ & \\
{\bf A2I2-Haze} & {\bf A+G} & {\bf O} & {\bf \cmark} & {\bf R} & {\bf 1,033} & {\bf \cmark} \\
\hline
\end{tabular}
}
\label{table:nonlin}
\end{table}

\subsection{Single Image Dehazing}
Based on \textit{Atmospheric scattering model} (ASM)~\cite{mccartney1976optics,nayar1999vision,narasimhan2003contrast}, a hazy image $I$ can be represented as:
\begin{equation}
    I(x)=t(x)J(x)+(1-t(x))A,
\end{equation}
where $J$, $t$, and $A$ denote the latent haze-free image, transmission map, and
global atmospheric light, respectively.
Dehazing using ASM involves estimating the transmission map $t(x)$ and global atmospheric light $A$. Existing approaches for dehazing can be broadly categorized as prior-based methods and learning-based methods.
\subsubsection{Prior-based Methods}
Dehazing methods based on priors~\cite{peng2007image,rao2005algorithms,berman2016non,fattal2008single,fattal2014dehazing,he2010single,tan2008visibility,zhu2015fast} first estimate transmission maps by exploiting the statistical properties of clean images and then obtain dehazed results using the scattering model. 
Tan~\etal~\cite{tan2008visibility} proposed an adaptive contrast enhancement method for haze removal by maximizing the local contrast of hazy images. 
He~\etal~\cite{he2010single} put forward an approach with a dark channel prior (DCP) that assumes the existence of at least one channel for every pixel whose value is close to zero. 
Zhu~\etal~\cite{zhu2015fast} proposed a color attenuation prior for haze removal by estimating the scene depth using a linear model. 
Fattal~\cite{fattal2014dehazing} proposed a color-line prior for the transmission map estimation by exploiting the regularity in natural images wherein small image patches lay in a one-dimensional distribution in the RGB color space.
%observed that pixels of image patches lay in a one-dimensional distribution in RGB color space and proposed a color-line prior for the transmission map estimation.
%
Berman~\etal~\cite{berman2016non} proposed a non-local prior based on a key observation that pixels in a given cluster are often non-local. Thus, colors of a haze-free image could be approximated by a few hundred distinct colors. 
Despite some promising results, prior-based approaches have limitations in performance because some of the strong assumptions of these hand-crafted priors often do not hold in a real-world environment.

\subsubsection{Learning-based Methods}
With the availability of large-scale paired data and powerful CNNs, learning-based dehazing methods have become popular in recent years.
MSCNN~\cite{ren2016single} estimated transmission map of the hazy images in a coarse-to-fine manner, where the coarse-scale net produced a holistic map based on the whole image and the fine-scale net refined it locally.
DehazeNet~\cite{cai2016dehazenet} proposed an end-to-end network that learned and estimated the mapping relations between hazy image patches and their corresponding medium transmissions.
DCPDN~\cite{zhang2018densely} embedded the atmospheric scattering model into the network so that it could jointly learn to estimate the transmission map, atmospheric light, and dehazing. 
Dehaze-cGAN~\cite{li2018single} estimated the clean image based on an encoder-decoder-based conditional generative adversarial network (cGAN). It also introduced the VGG features and $\ell_1$ regularized gradient prior to improving realism.
AOD-Net~\cite{li2017aod} recovered hazy images by reformulating the physical scattering model and developing a lightweight CNN to extract clean images. This approach has been extended to an end-to-end video dehazing and detection network~\cite{li2018end}. 
GFN~\cite{ren2018gated} introduced a multi-scale gated fusion end-to-end encoder-decoder network. This network obtains dehazed images by gating the important features of three inputs derived from the original hazy image by applying White Balance (WB), Contrast Enhancing (CE), and Gamma Correction (GC).
EPDN~\cite{qu2019enhanced} formulated the dehazing task as an image-to-image translation problem and proposed an enhanced pix2pix network to solve it. This work also introduced Perceptual Index (PI) as a metric to evaluate the dehazing quality from the perceptual perspective.
MSBDN~\cite{dong2020multi} used a boosting strategy and error feedback for progressive restoration.
4KDehazing~\cite{zheng2021ultra} proposed an ultra-high-definition image dehazing method via multi-guided bilateral learning. CNNs were used to build an affine bilateral grid that maintained detailed edges and textures.
Domain-invariant single image dehazing~\cite{shyam2021towards} recovered images affected by unknown haze distribution using spatially-aware channel attention to ensure feature enhancement and consistency between recovered and neighboring pixels.
MSRL-DehazeNet~\cite{yeh2019multi} embedded a multi-scale deep residual learning into an image decompoosition-guided framework, for single image haze removal.
DM$^2$F-Net~\cite{deng2019deep} presented a deep multi-model fusion network to attentively integrate multiple models to separate layers
for single-image dehazing.
LAP-Net~\cite{li2019lap} learned different levels of haze with different supervisions so that each stage could focus on a region with specific haze level.
Despite the improved performance, learning-based methods also have limitations. They require a large amount of data (often paired hazy and haze-free images), and they don't have debugging capabilities on failure cases.

\subsection{Object Detection}
In the deep learning era, object detectors can be grouped into two genres: ``two-stage detectors'' and ``one-stage detectors''. 
One-stage detectors classify and localize objects in a single-shot for dense prediction. Two-stage detectors have a region proposal module for sparse prediction. Compared to one-stage detectors, two-stage detectors usually achieve better accuracy but lower speed. RCNN series (RCNN~\cite{girshick2014rich}, Fast-RCNN~\cite{girshick2015fast}, Faster-RCNN~\cite{ren2016faster}, R-FCN~\cite{dai2016r}, and Mask-RCNN~\cite{he2017mask}) are the most representative of two-stage detectors. YOLO~\cite{redmon2016you,redmon2017yolo9000,redmon2018yolov3,bochkovskiy2020yolov4,ge2021yolox}, SSD~\cite{liu2016ssd}, RetinaNet~\cite{lin2017focal}, CenterNet~\cite{zhou2019objects}, and FCOS~\cite{tian2019fcos} are some of the state-of-the-art one-stage detectors. 
\paragraph{Architecture} The anatomy of an object detector includes a backbone, a neck, and a head. The backbone~\cite{he2016deep,howard2017mobilenets} serves as a feature extractor that starts from low-level structures to high-level semantics. The neck, composed of several bottom-up and top-down paths, is adopted to collect feature maps from different stages (scales). The most representative necks include FPN~\cite{lin2017feature}, PAN~\cite{liu2018path}, BiFPN~\cite{tan2020efficientdet}, and NAS-FPN~\cite{ghiasi2019fpn}. The head makes predictions in a multi-scale fashion. Head could either decouples object localization and classification (in two-stage detectors) or simultaneously makes the predictions for localization and classification (in one-stage detectors). In the head, anchors typically give the objects' prior location, shape and size. Recently, anchor-free one-stage detectors such as CenterNet~\cite{zhou2019objects}, FCOS~\cite{tian2019fcos}, and YOLOX~\cite{ge2021yolox} have gained popularity.
\paragraph{Label Assignment} Label assignment defines classification and regression targets for each anchor/grid cell. Traditional assigning strategies utilize local-view information, such as Intersection-over-Union (IoU)~\cite{ren2016faster,redmon2016you} or Centerness~\cite{tian2019fcos}. DeTR~\cite{carion2020end} is the first work that revisits label assignments from a global view, by considering one-to-one assignments using the Hungarian algorithm. To obtain the optimal global assignment under the one-to-many situation, OTA~\cite{ge2021ota} formulates label assignment as an Optimal Transport problem that defines each ground truth (together with the background) as the supplier and each anchor/grid cell as the demander. Thus, the best assignment could be obtained by solving the optimal transport in a Linear Programming form.
\begin{figure}
    \centering
    \includegraphics[width=1\linewidth]{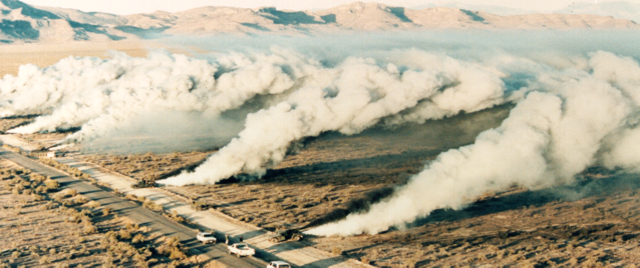}
    \caption{M56E1 smoke-generating system for producing large scale obscuration.}
    \label{fig:smoke_generator}
\end{figure}

\begin{figure}[!htb]
    \captionsetup[subfigure]{labelformat=empty,justification=centering,farskip=1pt,captionskip=1pt}
    \centering
    \subfloat{
        \includegraphics[width=0.95\linewidth]{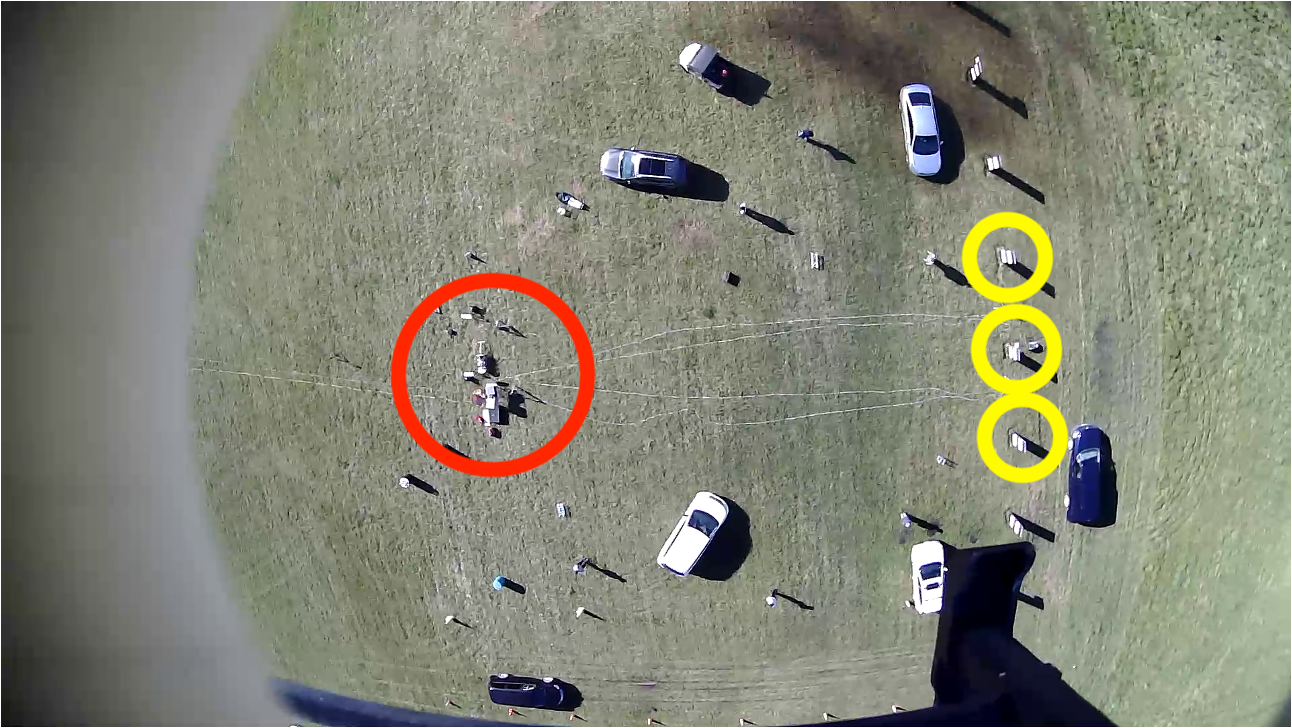}
    }
    % \subfloat
    % {
    %   \includegraphics[width=0.475\linewidth]{figures/data_collection/data_collection1.JPG}
    % }
    \\
    \subfloat
    {
       \includegraphics[width=0.475\linewidth]{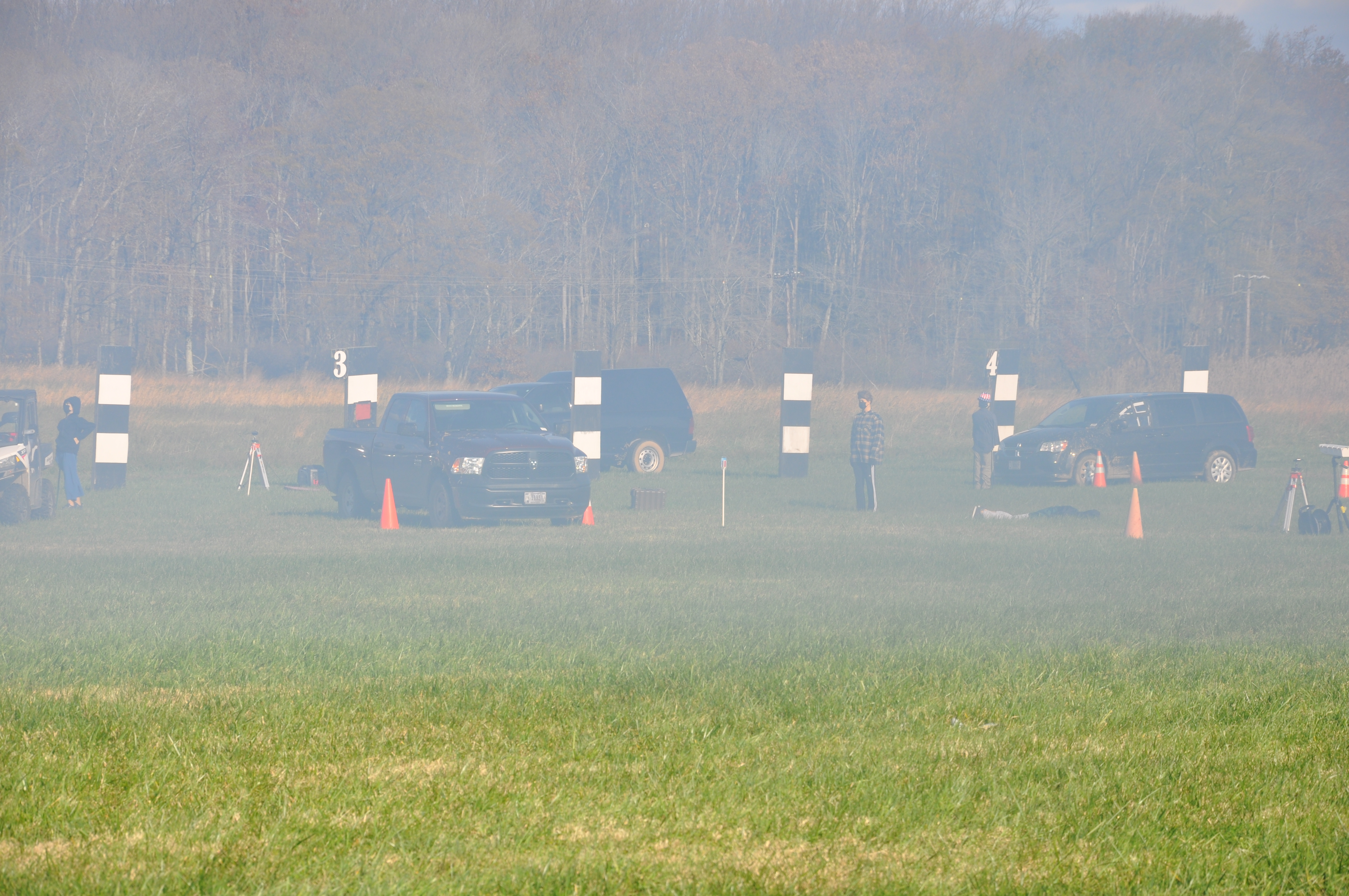}
    }
    % \subfloat
    % {
    %   \includegraphics[width=0.475\linewidth]{figures/data_collection/data_collection3.JPG}
    % }
    \subfloat
    {
      \includegraphics[width=0.475\linewidth]{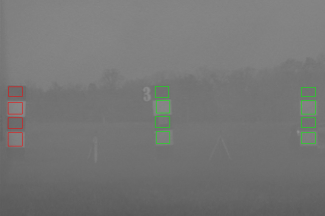}
    }
    % \subfloat
    % {
    %   \includegraphics[width=0.475\linewidth]{figures/data_collection4.JPG}
    % }
    \caption{Example images demonstrating the data collection process. The top figure shows a bird-view layout of the testing site. The yellow circles show target contrast boards, and the red circle shows an infrared and a visible camera. The bottom two figures show target contrast boards (the black and white pylons) used for local haze density measurement.}
    \label{fig:data_collection}
\end{figure}

\section{A2I2-Haze Dataset}
The A2I2-Haze Dataset consists of paired haze and haze-free aerial and ground imagery acquired by a small UAS and UGV, respectively. The data set is synchronized with in-situ smoke measurements as well as altitude data acquired from the flight controller. A2I2-Haze were captured in two separate settings – a) grass field as the background, and b) ground with concrete slabs as the background. For the aerial dataset, after each trial, target objects were re-positioned to generate different configuration of objects. The images with good pairing were then selected from the trials. However, the annotated ground dataset is only available from one configuration of target objects at this time. This section provides a detailed description of the data collection procedure and the post-processing pipeline.

\subsection{Data Collection}
Data collection was conducted at the DEVCOM Chemical and Biological Center's (CBC) M-field test range utilizing the world's most comprehensive obscurant generation facility and assessment technologies to measure the smoke concentration accurately. The facility has a platoon of six M56E1 smoke-generating systems (shown in Fig.~\ref{fig:smoke_generator}) that provide large area obscuration by disseminating obscurants either simultaneously or separately while stationary or mobile. The current onboard obscuration capabilities include visual obscuration using fog oil, infrared obscuration using graphite, and radar obscuration using carbon fiber. These obscurants thus can degrade the perception capabilities of threat weapons and Reconnaissance, Intelligence, Surveillance, Targeting, and Acquisition (RISTA) operating in the Visual, Infra-Red (IR), and Millimeter Wave (MMW) portions of the electromagnetic spectrum. Specifically, the obscurant materials are designed to effectively absorb and scatter energy, thus making target acquisition difficult.

As part of the Phase 1 DVE data collection, we obtained a wide range of imagery of targets in the DEVCOM CBC M-Field test area using visual obscurants generated by using fog oil in the M56E1 Smoke-Generating System. The target objects included vehicles, mannequins, and man-made obstacles encountered during UGV maneuvers, such as traffic cones, barriers, and barricades. The ground truth for smoke concentration was measured using laser-based transmissometers. These instruments measured transmittance through the smoke cloud at a $625$nm wavelength and provided a quantitative measure of the clouds' effectiveness in attenuating visible light. The transmissometers' light sources were Z-Laser, S3 series diode modules (Edmund Optics; Barrington, NJ) with output wavelength and power of $625$nm and $5$mW, respectively.

The laser power received at the detector was measured before obscuration, providing a baseline to compare the effect of the fog oil obscurant. During the series of trials, the transmittance was reduced from $100\%$ to below $1\%$, providing a range of obscurant concentrations to correlate with sensor performance. This variation in transmittance (and thus obscurant concentration) is critical as real-world scenarios will have varying haze concentrations. All measurements were time-synchronized with the UAV and UGV sensors to relate the obscuration properties to the sensor's response.

In addition to the transmissometer-based obscuration measurement, we used target boards, as shown in Fig.~\ref{fig:data_collection}, as the second technique for haze measurement. The black and white pylons in the bottom two figures show target contrast boards used for haze density local measurement.

\subsubsection{UAV Dataset}
Aerial video of the DVE releases was captured by Deep Purple 3 (DP3), a custom UAS developed and operated by DEVCOM CBC. DP3 was primarily designed to carry sensor payloads weighing up to 6 lbs using the Array Configured of Remote Networked Sensors (ACoRNS) interface system. In addition to sensor payloads, DP3 can also be equipped with a variety of camera systems. For the A2I2-Haze data collection, a pair of cameras were mounted in the nosecone to capture both longwave Infrared (LWIR) and visible spectrum. DP3 was equipped with a FLIR Boson 640 ($95^\circ$ FOV) for LWIR data capture. Calibrated visible spectrum data was captured using an ELP-USBFHD01M-L21 with a 2.1mm lens, approximately $120^\circ$ field of view, and an OV2710 sensor. Cameras mounted on the nosecone could be rotated to face $0^\circ$, $45^\circ$, $90^\circ$ relative to the body of the UAS, with $90^\circ$ being a nadir pointing the camera, and $0^\circ$ in line with the nose of the UAS. For the A2I2 collection, cameras were clocked at either $90^\circ$ or $45^\circ$. The UAS is flying in a lawn-mowing pattern based on pre-programmed GPS coordinates (aka, waypoints). The flight plan for the collection was defined using a survey grid in Mission Planner to generate a cross-hatch pattern over a predefined area. The area was selected using an arbitrary rectangle within the Mission Planner, with the bounds of the rectangle selected to keep the area of interest within view of the cameras as much as possible over a range of altitudes. The survey grid was cross-hatched with a 10m lane spacing. In order to maintain consistency between altitudes, the survey grid was copied, and the altitude was increased in 5m increments from 15-50m. UAS was commanded to face toward the center of the grid at all points during its flight. The GPS coordinate for the center of the grid was determined by the UAS by manually placing the UAS at the point of interest and using its GPS to determine the center of the grid. 
\subsubsection{UGV Dataset}\label{sec:ugv_dataset_generation}
In addition to the aerial vehicle dataset collection described above, a ground vehicle was also used to capture a second viewpoint from inside the haze effect on the surface. The ground vehicle used to collect this dataset is a Clearpath Husky mobile robot which is equipped with a variety of cameras and lidar sensing modalities. The visual dataset for this paper was collected from the color image from the left camera of a Carnegie Robotics MultisenseSL stereo + lidar sensor, which was mounted in a forward-facing position at the front of the robot at the height of 0.6 meters. This sensor package also generated a frame-synchronized second monochrome image from the right camera, a depth image, and a lidar point cloud from the included Hokuyo 2D lidar scanner, which is actuated through rotation about the forward axis to cover a 3D hemispherical volume. A FLIR Boson infrared camera was also used to capture thermal images of the scene. The robot also collected point clouds from an Ouster OS1-64 lidar which is used with a robot mapping engine based on OmniMapper~\cite{trevor2014omnimapper} to localize the UGV in a global frame of reference. The trajectory of the UGV was teleoperated from a remote location via live sensor feedback to collect these datasets. In this paper, only the visual imagery is analyzed; however, the analysis of additional sensor modalities is planned for future work and can be made available to other interested researchers.

\subsection{Data Augmentation and Synchronization }
The raw data collected by sensors were subjected to a series of data augmentation and synchronization procedures in order to enhance their quality. The high mobility of UAV-mounted cameras has brought additional challenges compared to traditional datasets, such as variations in altitude, view angles, and weather conditions. NDFT~\cite{wu2019delving} named these variations as UAV-specific nuisances, which constitute a large number of fine-grained domains. NDFT shows that these nuisances can be used to train a cross-domain object detector that stays robust to many fine-grained domains. As part of A2I2-Haze, we collected these UAV-specific nuisances as metadata that are synchronized with UAV's clock.

\begin{figure}[!htb]
    \captionsetup[subfigure]{labelformat=empty,justification=centering,farskip=1pt,captionskip=1pt}
    \centering
    \subfloat
    {
    \includegraphics[width=0.45\linewidth]{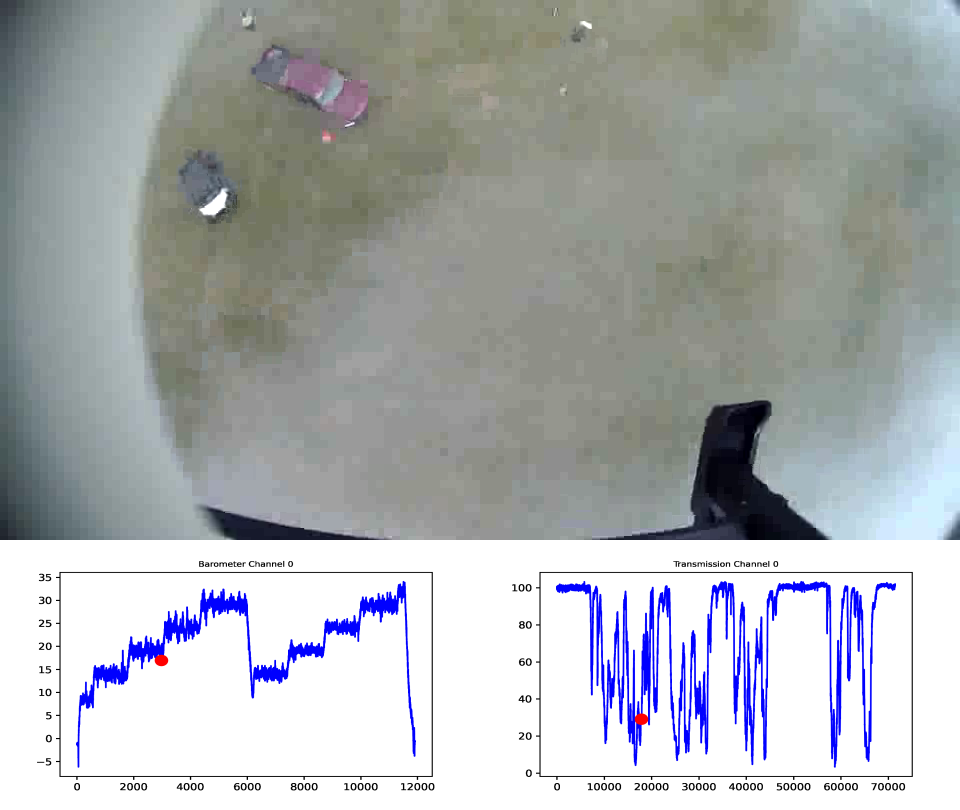}
    }
    \subfloat{
    \includegraphics[width=0.45\linewidth]{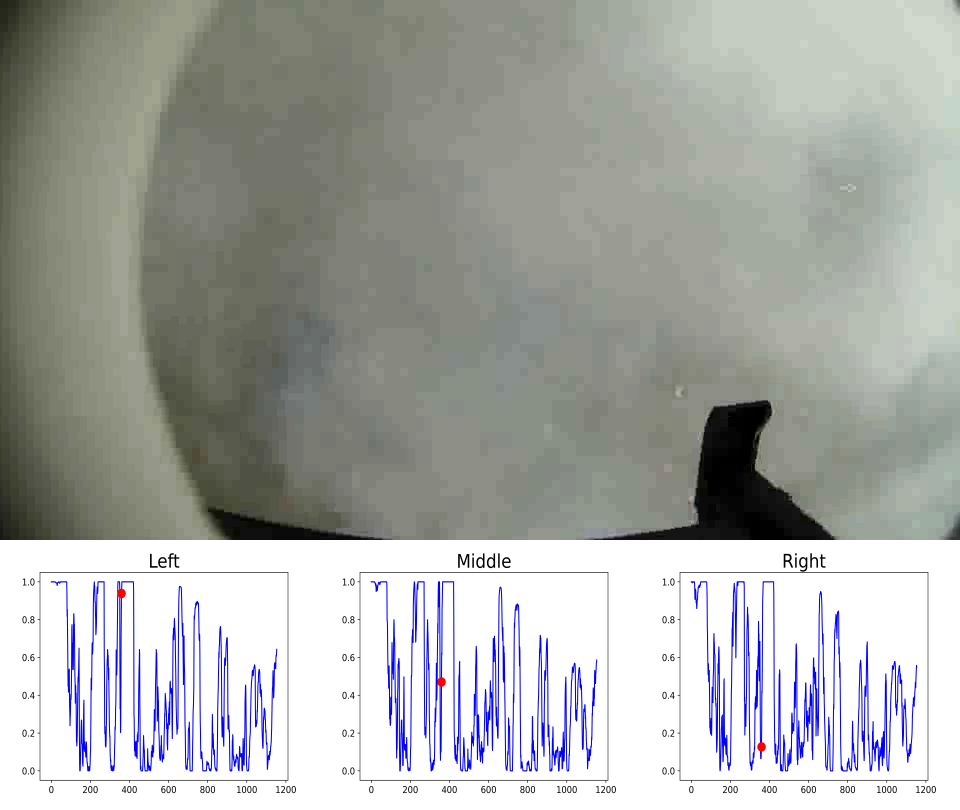}
    }
    \caption{UAV visual data synchronized with collected metadata. The left figure shows the visual data synchronized with the barometer and transmissometer metadata. The barometer records the height data of the UAV, and the transmissometer records the laser transmission rate, which measures the haze density. The right figure shows the synchronization of visual data with contrast on three target boards, the second technique that measures the haze density.}
    \label{fig:sync_uav}
\end{figure}

\begin{figure}[!htb]
    \captionsetup[subfigure]{labelformat=empty,justification=centering,farskip=1pt,captionskip=1pt}
    \centering
    \includegraphics[width=0.8\linewidth]{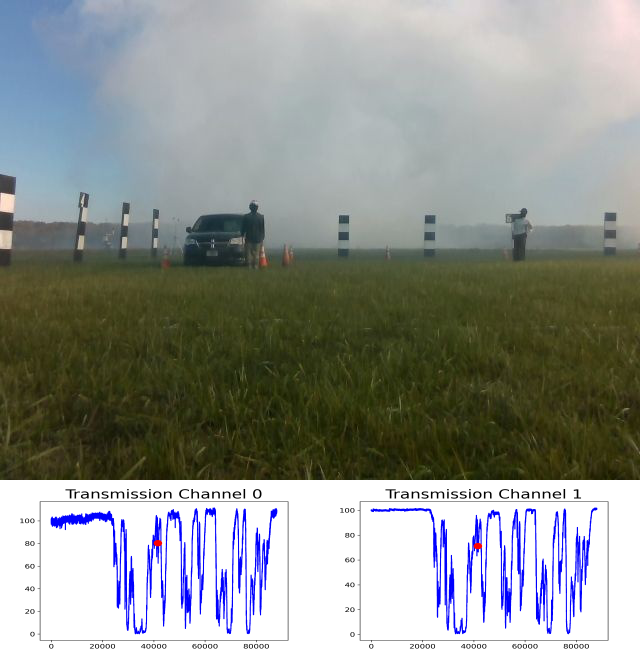}
    \caption{UGV visual data synchronized with contrast on target board, which is the second technique that measures the haze density.}
    \label{fig:sync_ugv}
\end{figure}

Fig.~\ref{fig:sync_uav} shows the UAV visual data synchronized to the metadata, including the altitude measured by UAV's barometer, and haze density measured by two independent techniques (section III A)- a) using laser transmission rate on a transmissometer and b) using visual contrasts on three target boards. Similar techniques were used for synchronizing the UGV dataset to haze density measurement, as shown in Fig.~\ref{fig:sync_ugv}.

% \subsection{Haze/Haze-Free Pairing for UAV Dataset (\zhenyu{Zhenyu})}
% Subsection text here.

\subsection{Labeling and pairing of UAV Dataset}
Quantitative evaluation of low-level dehazing and high-level detection tasks require paired (hazy and haze-free) images with annotated target objects. 

The detection task requires accurate data curation using 2D bounding box annotation tools to define regions of interest. Annotation was performed by external annotators from Engineering and Computer Simulations (ECS) using their labeling platform. Ten object classes were provided as an ontology for labeling. The objects classes were: Sedan, Van, Pickup Truck, Utility Task Vehicle (UTV), Mannequin, Unmanned Ground Vehicle (UGV), Barrel, Jersey Barrier, Aluminum Truss, and Red backpacks. The annotators provided rectangular bounding boxes for each object. They also provided the subjective haze level for each bounding box as light, medium, or heavy.

Dehazing task requires paired hazy and haze-free images. Due to UAV's high flying altitude and rapid movement speed, we sought scene-level pairing rather than pixel-level pairing. We proposed a ``Coarse-to-Fine Matching'' strategy to achieve scene-level pairing. Given two video sequences (hazy and haze-free), we first cut each video into short clips of 2 seconds. We then manually selected hazy and haze-free clips that were similar at the scene level. Lastly, for each selected clip pair ($V_s$ and $V_t$), we used Alg.~\ref{algo:coarse2fine} to find the best pair-able frames with the least amount of relative translation or rotation. Since $V_s$ and $V_t$ were describing the same scene, we hypothesize that the homography matrix (computed from keypoint matching by LoFTR~\cite{sun2021loftr}) is mostly composed of rotational and translational transformation. 
The distance of the homography matrix to an identity matrix served as a metric that measures scene-level similarity. We assume that the best pair-able frames have the smallest distance. We manually check the returned paired hazy and haze-free images by the proposed ``Coarse-to-Fine Matching'' strategy. If it fails, we manually find the best pair-able images given the selected hazy and haze-free clips that are similar at the scene level. Such a final human intervention could guarantee the correctness even in the presence of dense haze.

\begin{algorithm}
\caption{Coarse-to-Fine Matching}\label{algo:coarse2fine}
\SetAlgoLined
$N,M \gets \text{Size}(V_s), \text{Size}(V_t)$ \\
$d_{min} \gets \infty$\\
\For {$i \gets 1$ \KwTo $N$} {
    \For{$j \gets 1$ \KwTo $M$}{
        $I_s, I_t \gets V_s[i], V_t[j]$ \\
        \tcp{Keypoints Matching}
        $p_s, p_t \gets {\small \text{LoFTR}(I_s, I_t)}$ \\
        $\mathbf{M} \gets {\small \text{HomographyMatrix}(p_s,p_t))}$ \\
        $d \gets \text{sqrt}(\text{Trace}( (\mathbf{M}-\mathbf{I})^\intercal*(\mathbf{M}-\mathbf{I})))$ \\
        \If{$d < d_{min}$}{
            $I_s^*, I_t^* \gets I_s, I_t$
        }
    }
}
\Return $I_s^*, I_t^*$
\end{algorithm}
\subsection{Labeling and Pairing for UGV Dataset }
Labeling the UGV dataset presents unique challenges due to the difficulty of determining object bounding boxes in hazy images to a human labeler. To address these issues, a 3D object labeling scheme was developed to assist with the labeling process. Initial bounding boxes were generated by projecting the 3D bounding volumes of each object visible in each image, given the UGV's known observation point. To determine the UGV's global position and orientation for each captured image, the Monte-Carlo particle filter-based localization package \emph{AMCL}~\cite{ros} from ROS~\cite{quigley2009ros} was used to localize into a 2D map which was generated for each obstacle configuration using OmniMapper~\cite{trevor2014omnimapper}. The lidar data from the Ouster OS1-64 was used with the UGV's odometry and internal IMU to generate this map. The projections of the 3D object bounding volumes were then refined by hand with LabelMe~\cite{labelme}.

Two UGV trajectories were collected for each object configuration: one with the haze effect and the other without the haze effect. A 2D map was generated for each obstacle configuration and used to localize the UGV with Monte-Carlo localization as described in Section~\ref{sec:ugv_dataset_generation}. This global localization was used to tag each image with the UGV's position and orientation at the capture time. This position and orientation information was used with a Hungarian algorithm to find the optimal pairing between haze and haze-free images, which minimizes the error metric $E=10\Delta_o + \Delta_p$ where $\Delta_o$ is the difference in orientation and $\Delta_p$ is the difference in position. This metric weighs orientation more heavily as changes in orientation have a more significant effect on image viewpoint.

\begin{figure*}[!htb]
    \captionsetup[subfigure]{labelformat=empty,justification=centering,farskip=1pt,captionskip=1pt}
    \centering
    \begin{mdframed}[style=RedBox,nobreak=true]
    \subfloat
    {
       \includegraphics[width=0.32\linewidth]{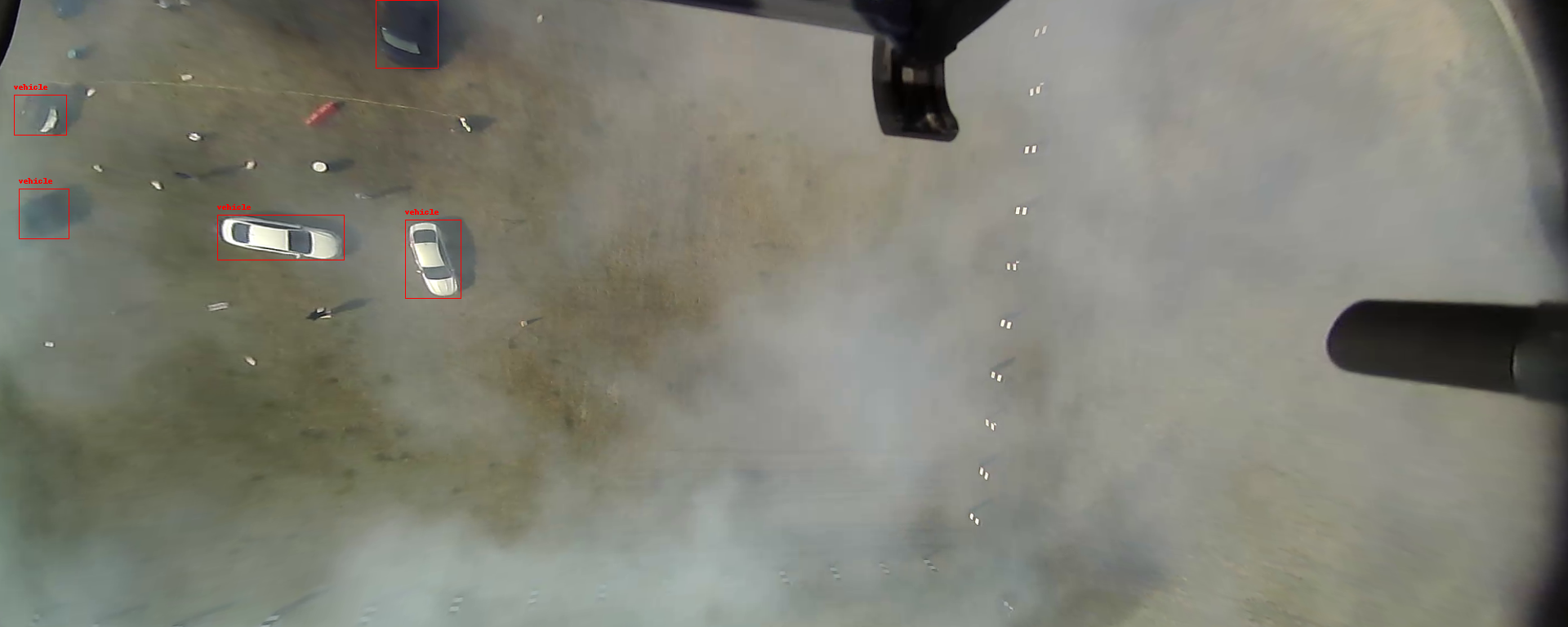}
    }
    \subfloat
    {
       \includegraphics[width=0.32\linewidth]{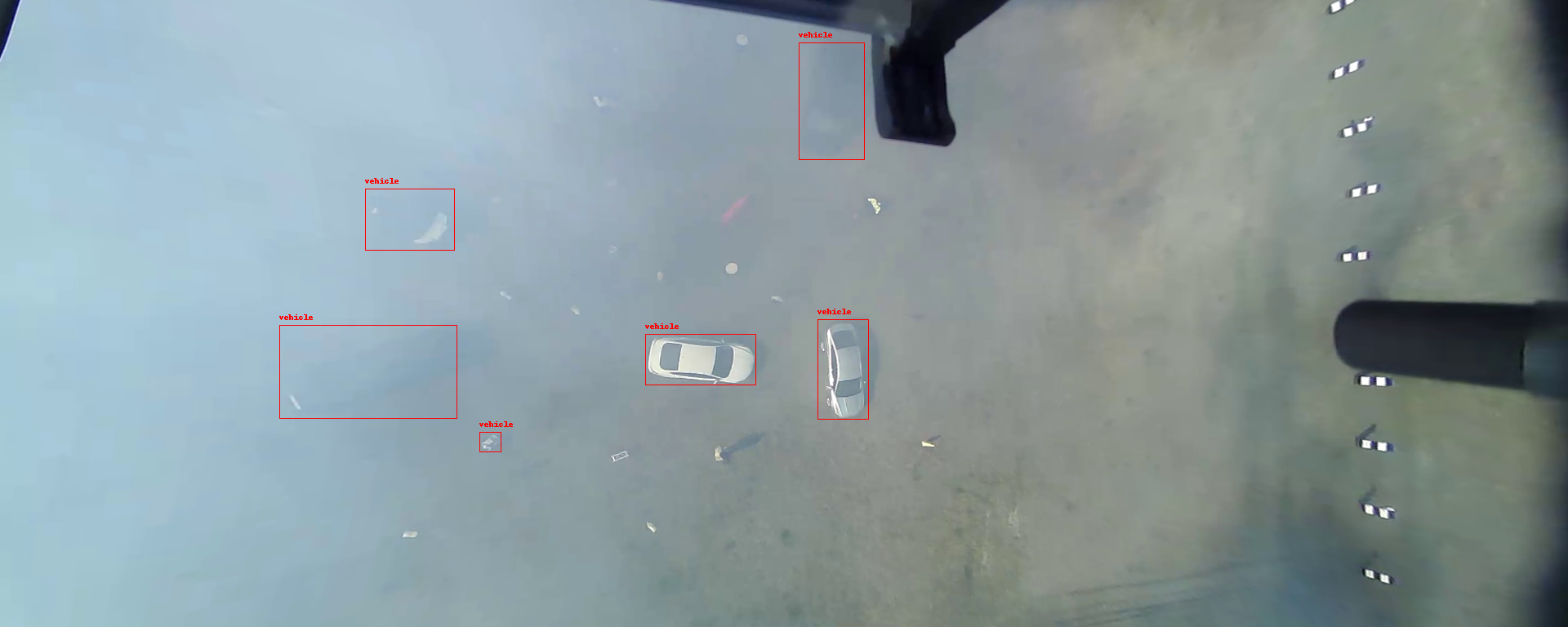}
    }
    \subfloat
    {
       \includegraphics[width=0.32\linewidth]{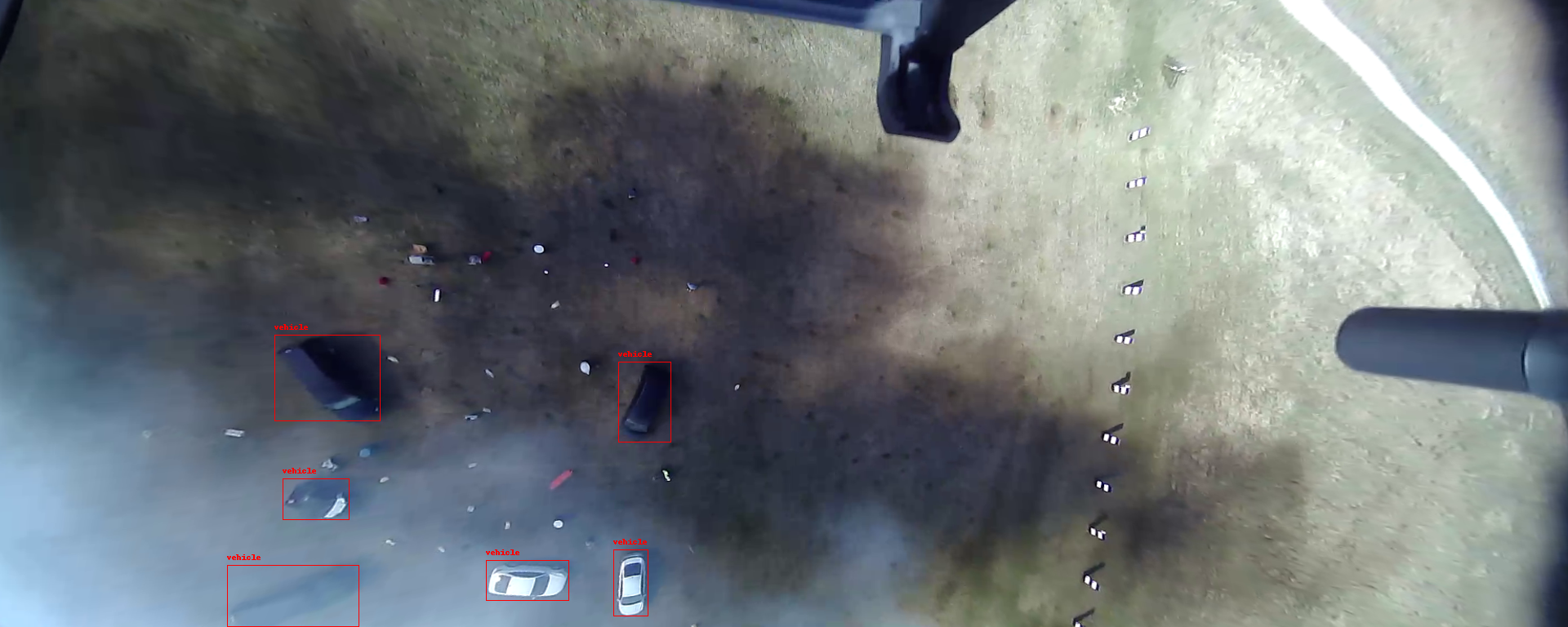}
    }
    \end{mdframed}
    \begin{mdframed}[style=BlueBox,nobreak=true]
    \subfloat
    {
       \includegraphics[width=0.32\linewidth]{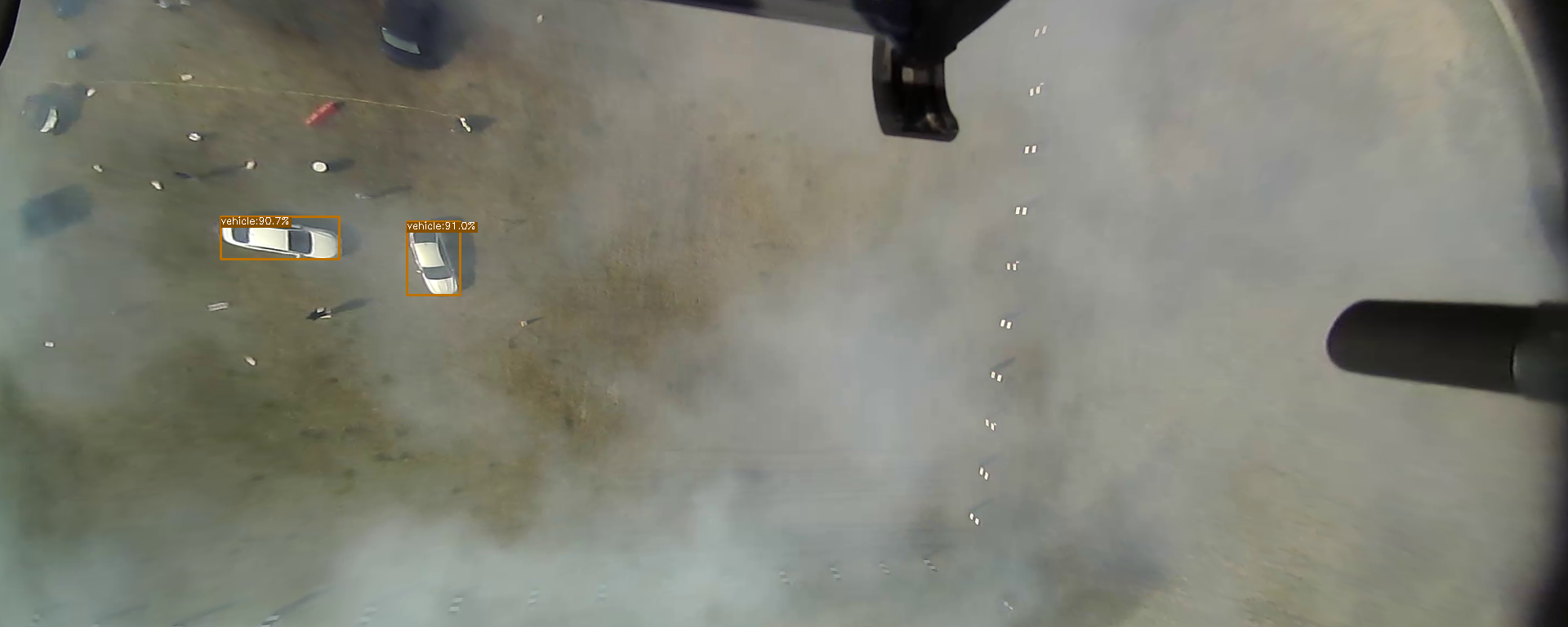}
    }
    \subfloat
    {
       \includegraphics[width=0.32\linewidth]{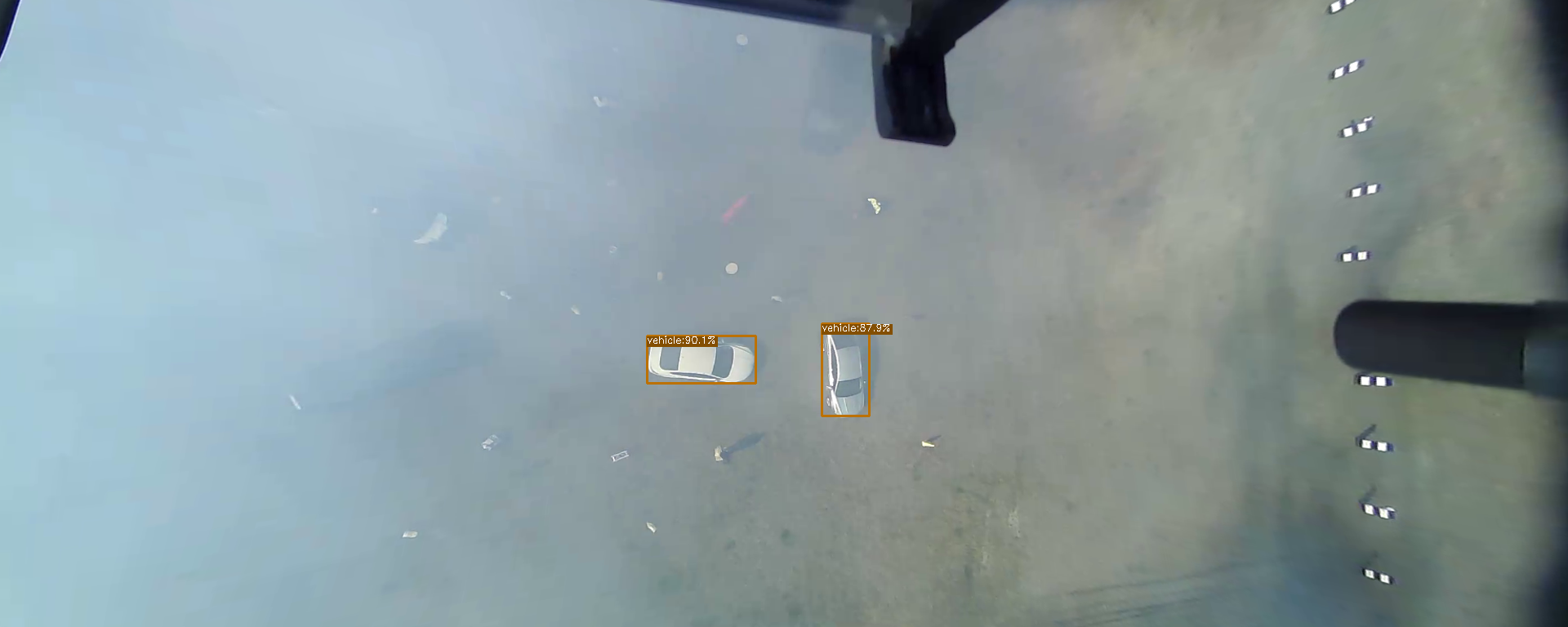}
    }
    \subfloat
    {
       \includegraphics[width=0.32\linewidth]{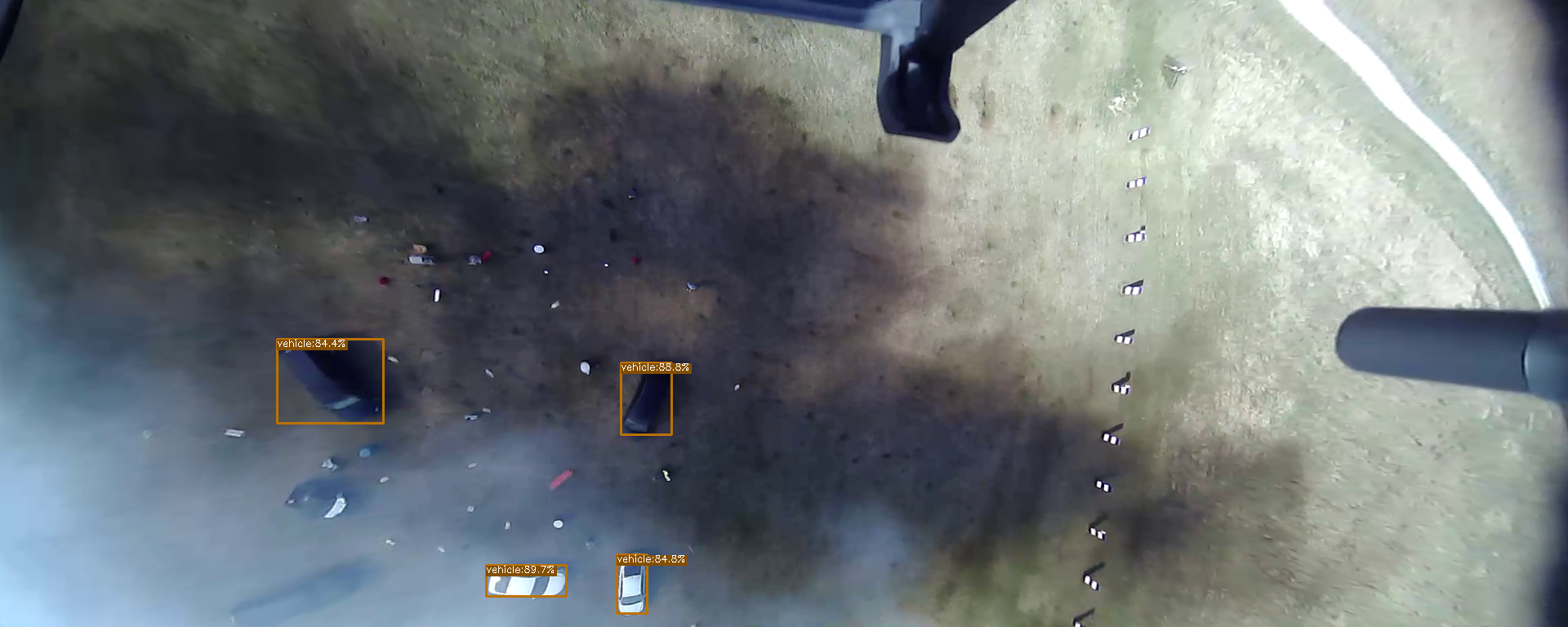}
    }
    \end{mdframed}
    \subfloat
    {
       \includegraphics[width=0.32\linewidth]{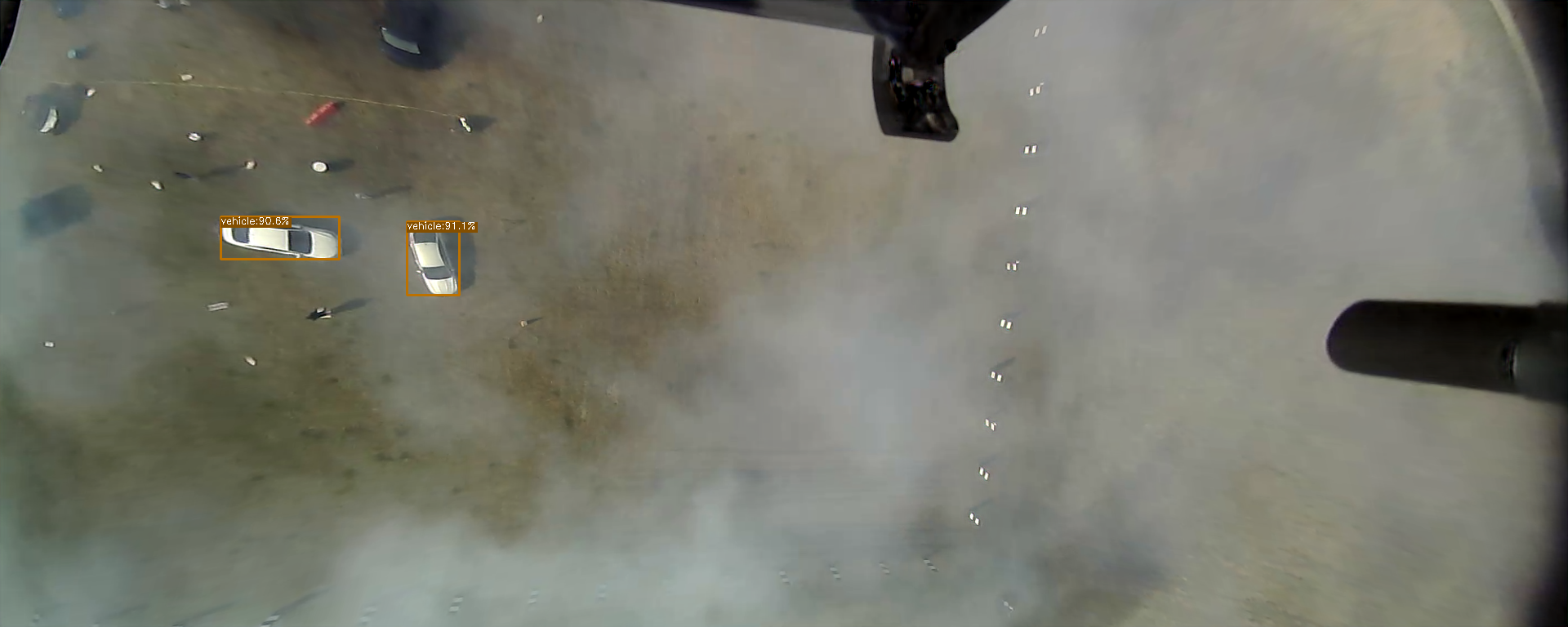}
    }
    \subfloat
    {
       \includegraphics[width=0.32\linewidth]{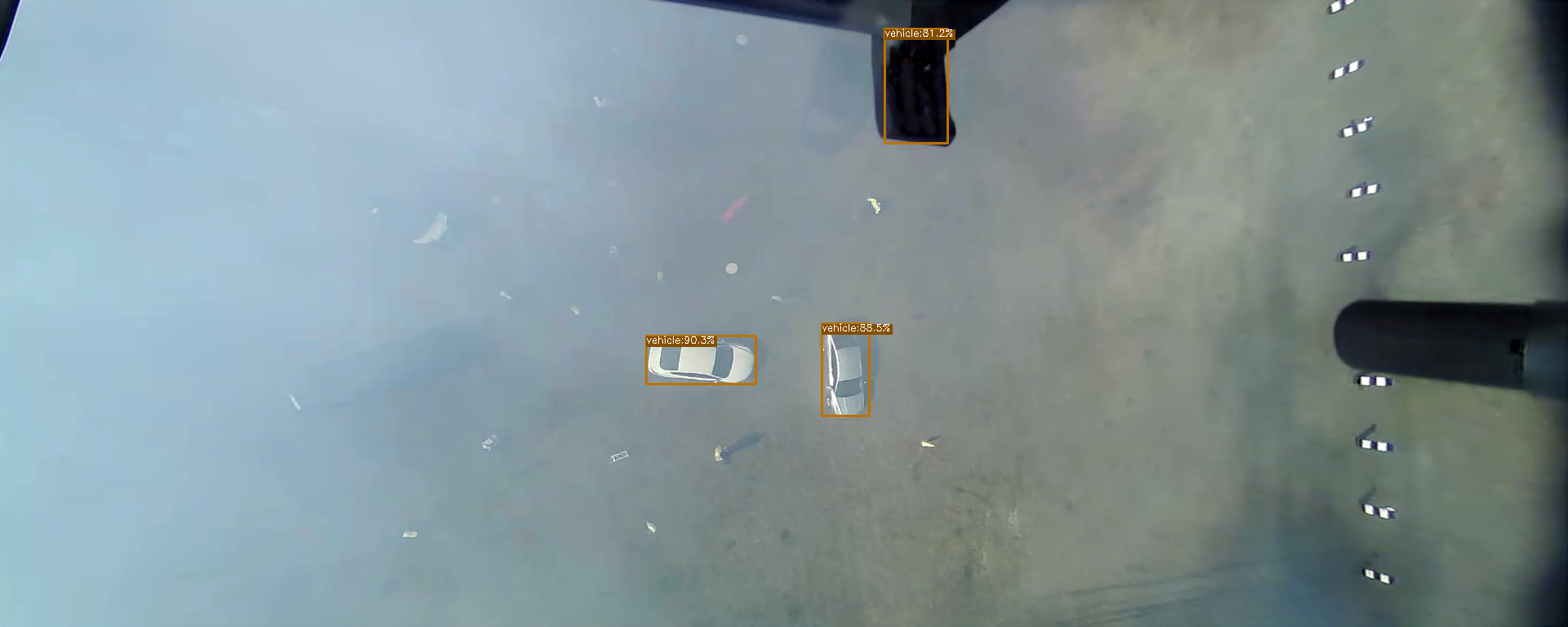}
    }
    \subfloat
    {
       \includegraphics[width=0.32\linewidth]{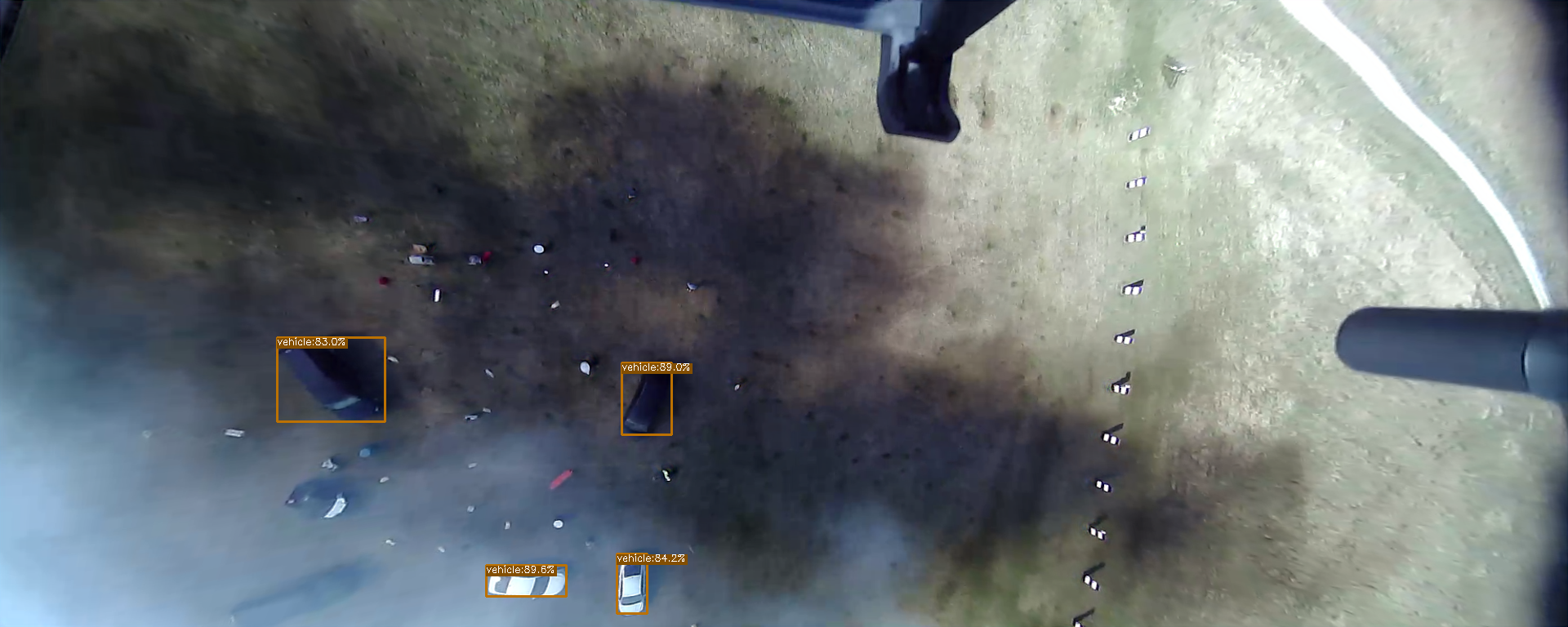}
    }
    \\
    \subfloat
    {
       \includegraphics[width=0.32\linewidth]{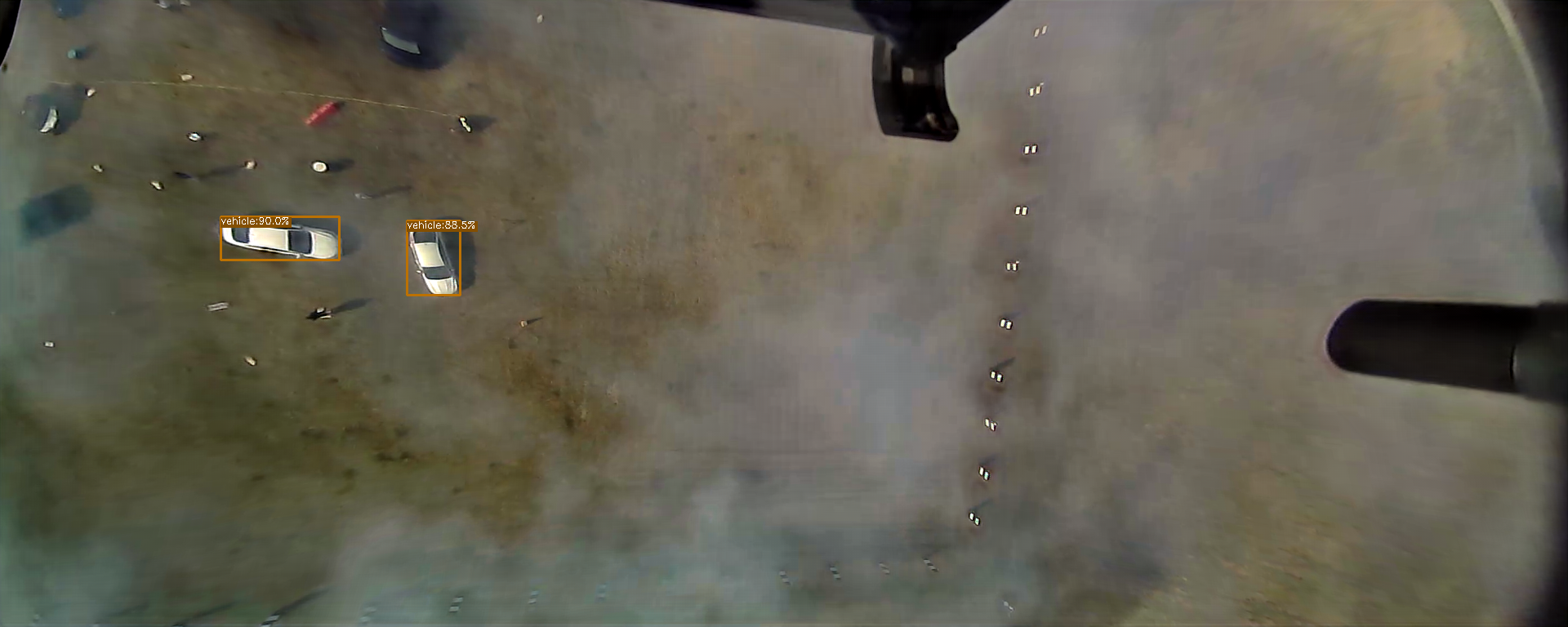}
    }
    \subfloat
    {
       \includegraphics[width=0.32\linewidth]{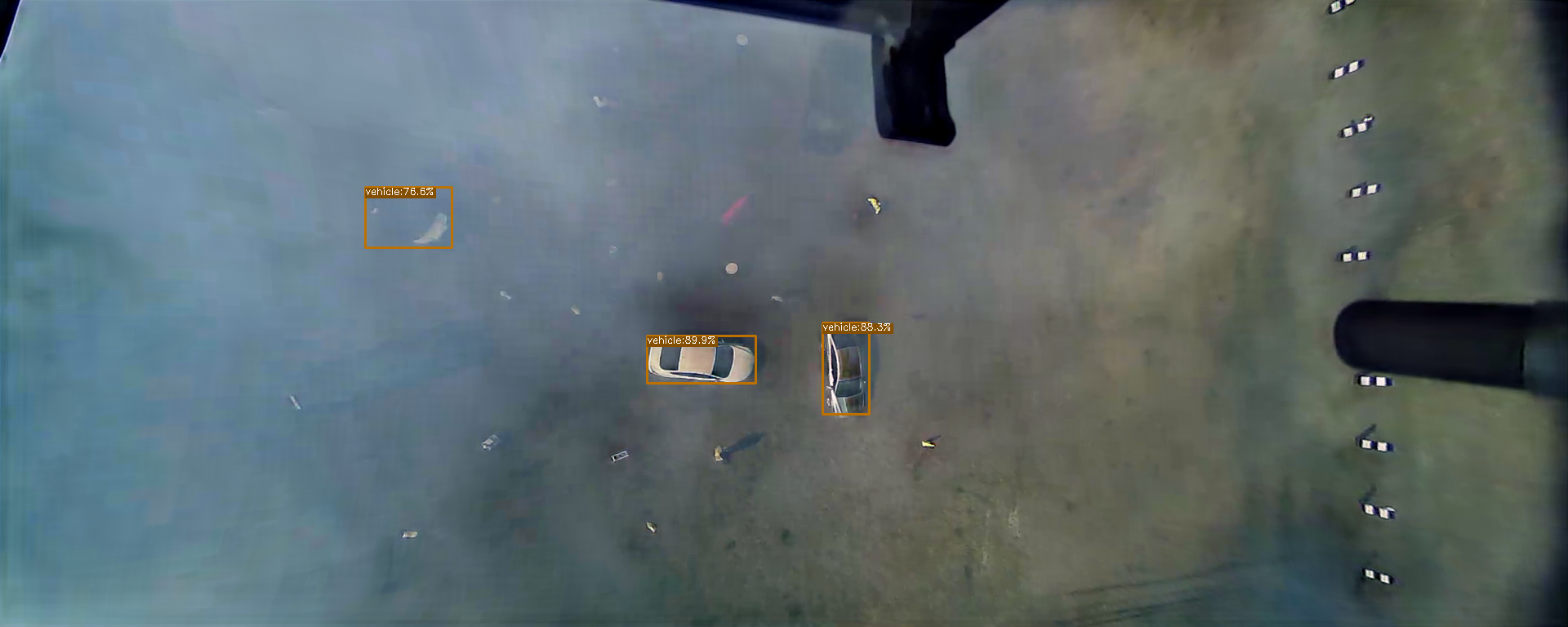}
    }
    \subfloat
    {
       \includegraphics[width=0.32\linewidth]{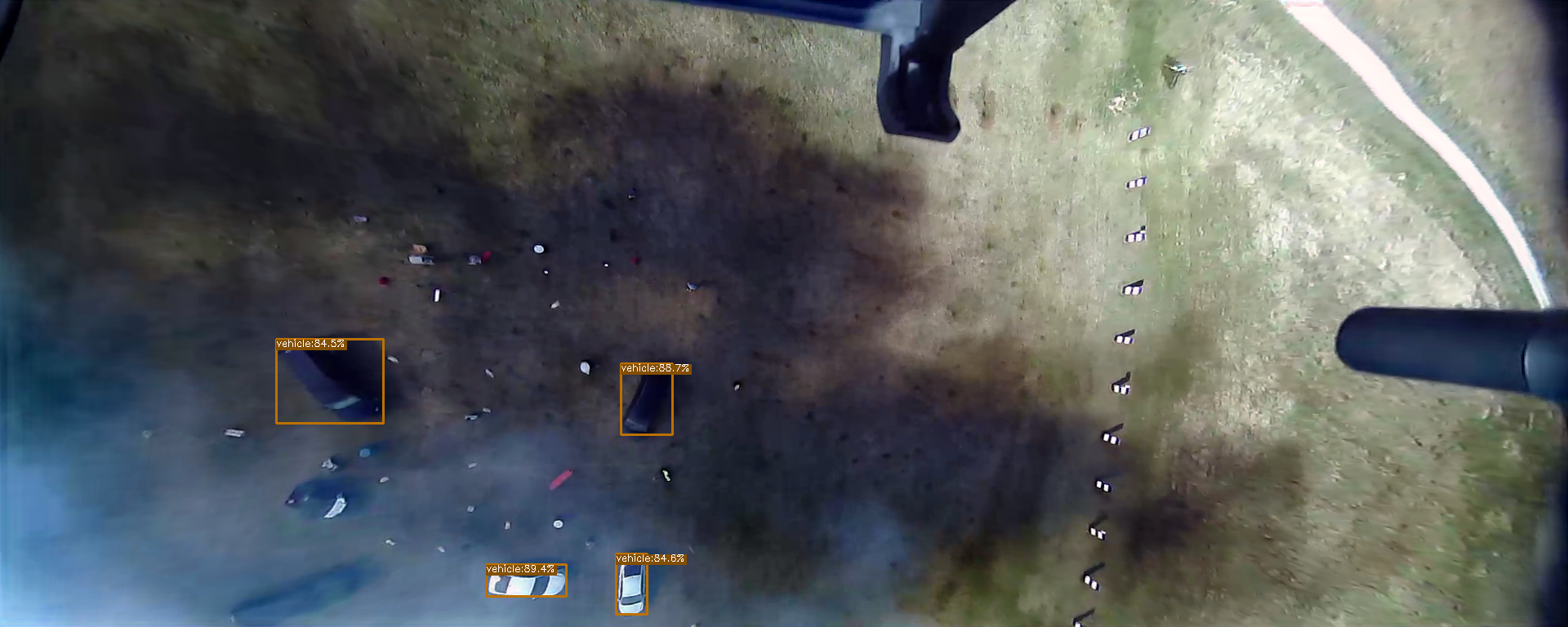}
    }
    \\
    \subfloat
    {
       \includegraphics[width=0.32\linewidth]{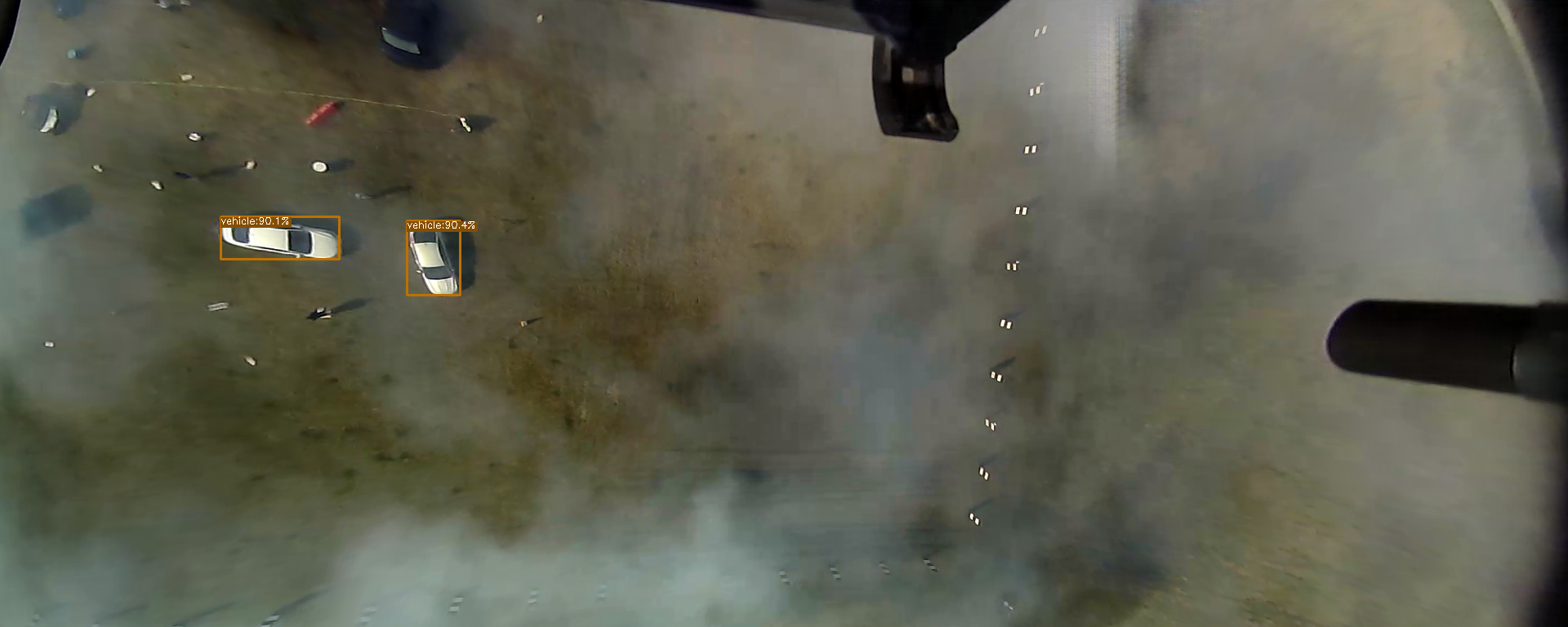}
    }
    \subfloat
    {
       \includegraphics[width=0.32\linewidth]{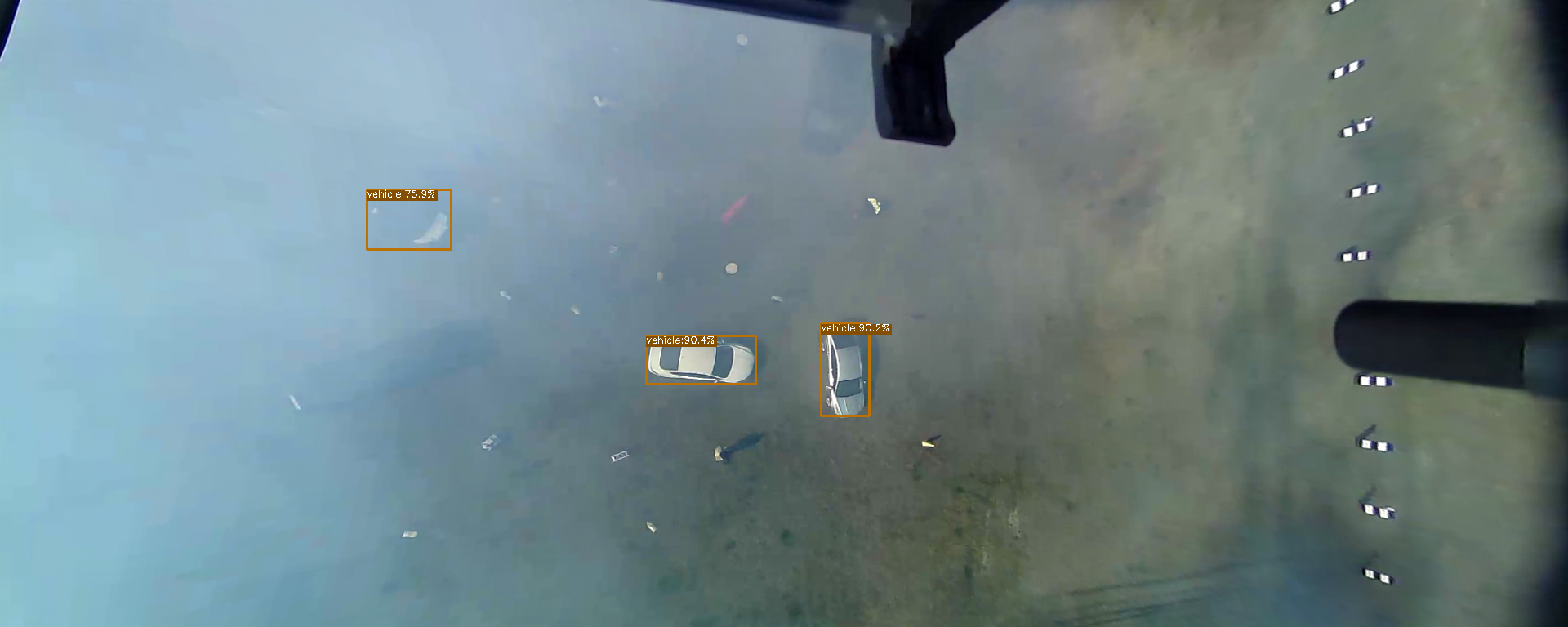}
    }
    \subfloat
    {
       \includegraphics[width=0.32\linewidth]{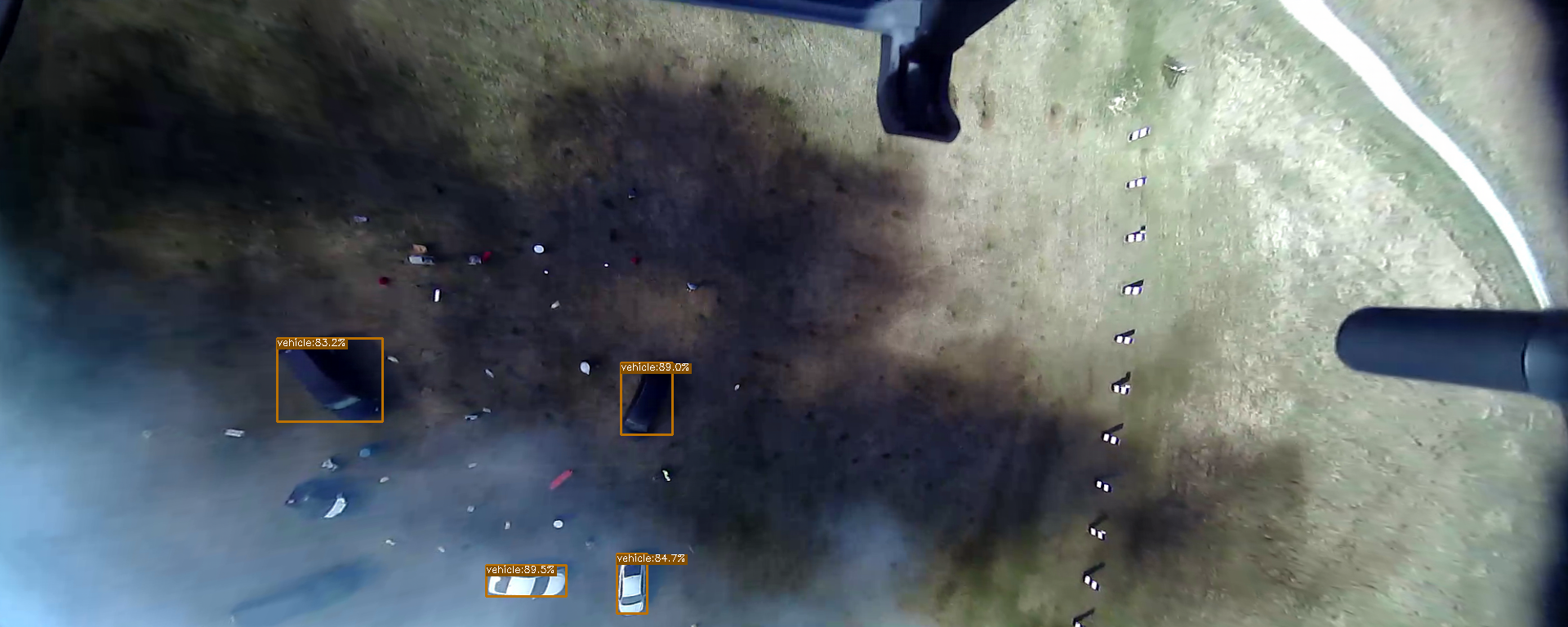}
    }
    \\
    \subfloat
    {
       \includegraphics[width=0.32\linewidth]{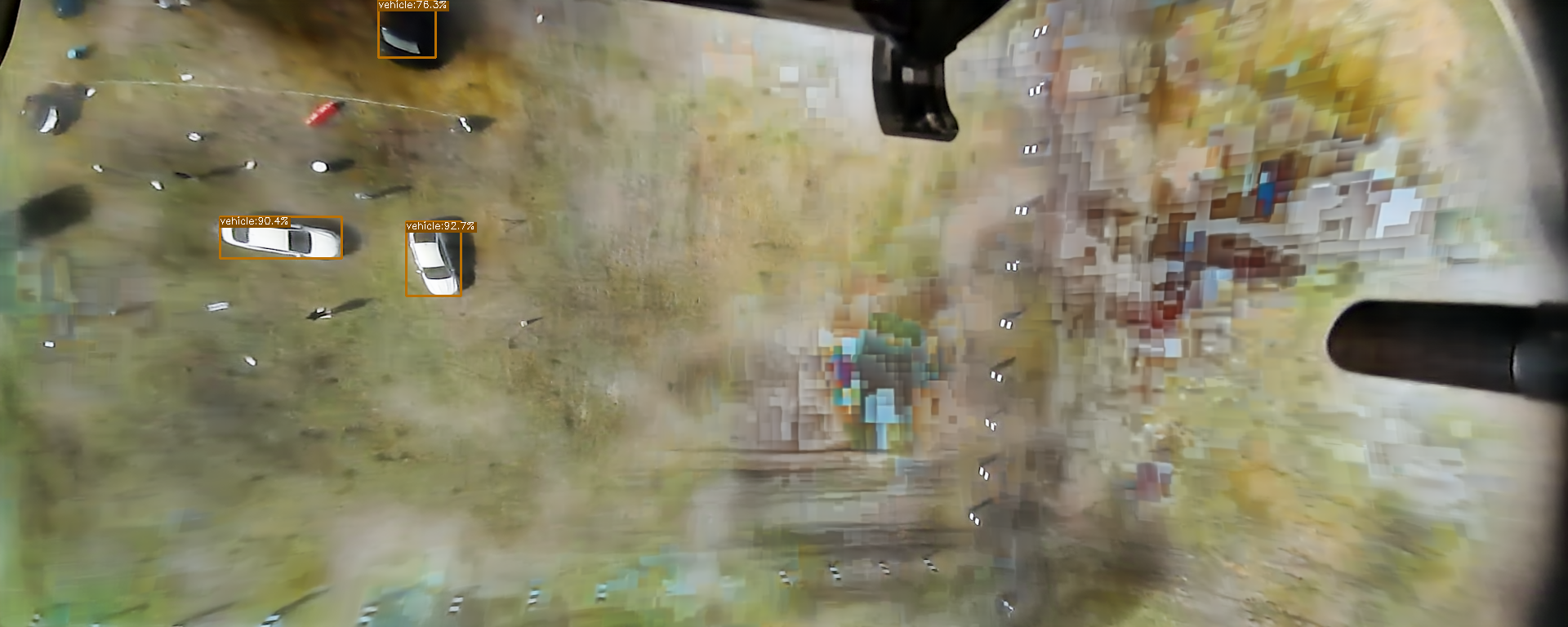}
    }
    \subfloat
    {
       \includegraphics[width=0.32\linewidth]{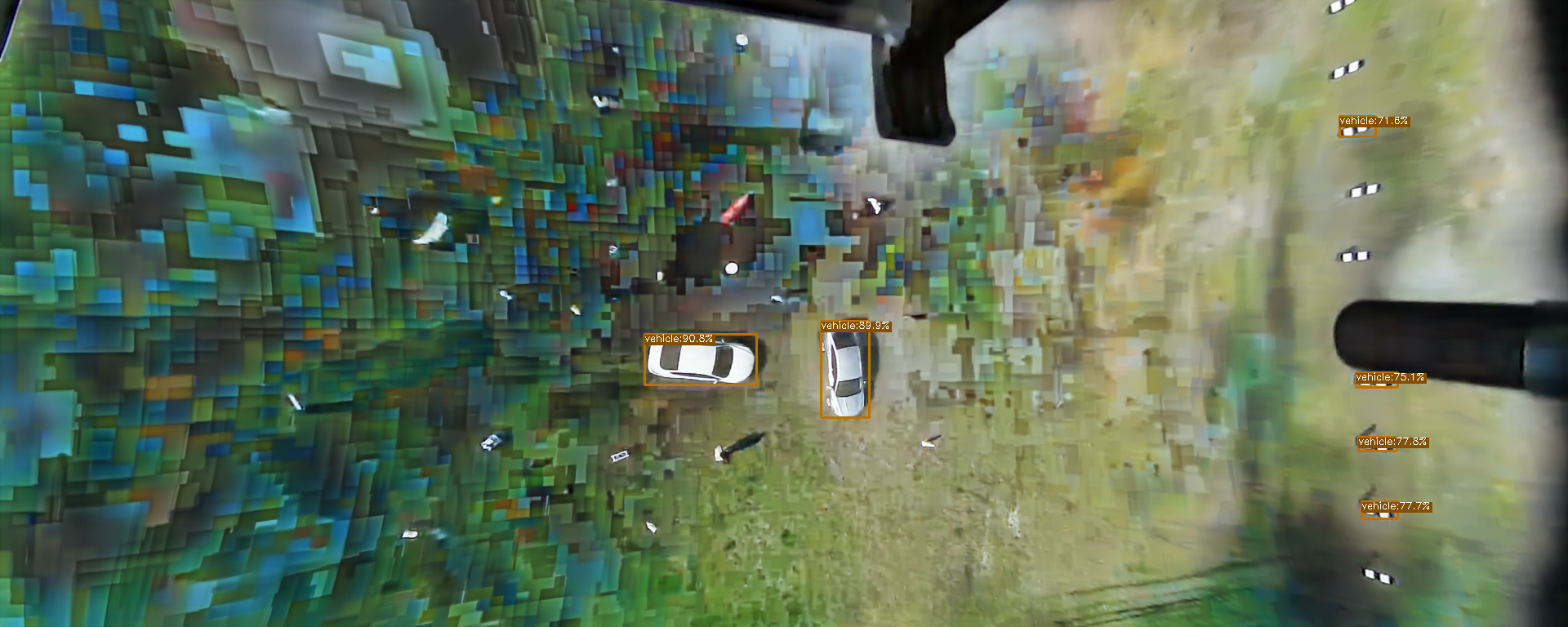}
    }
    \subfloat
    {
       \includegraphics[width=0.32\linewidth]{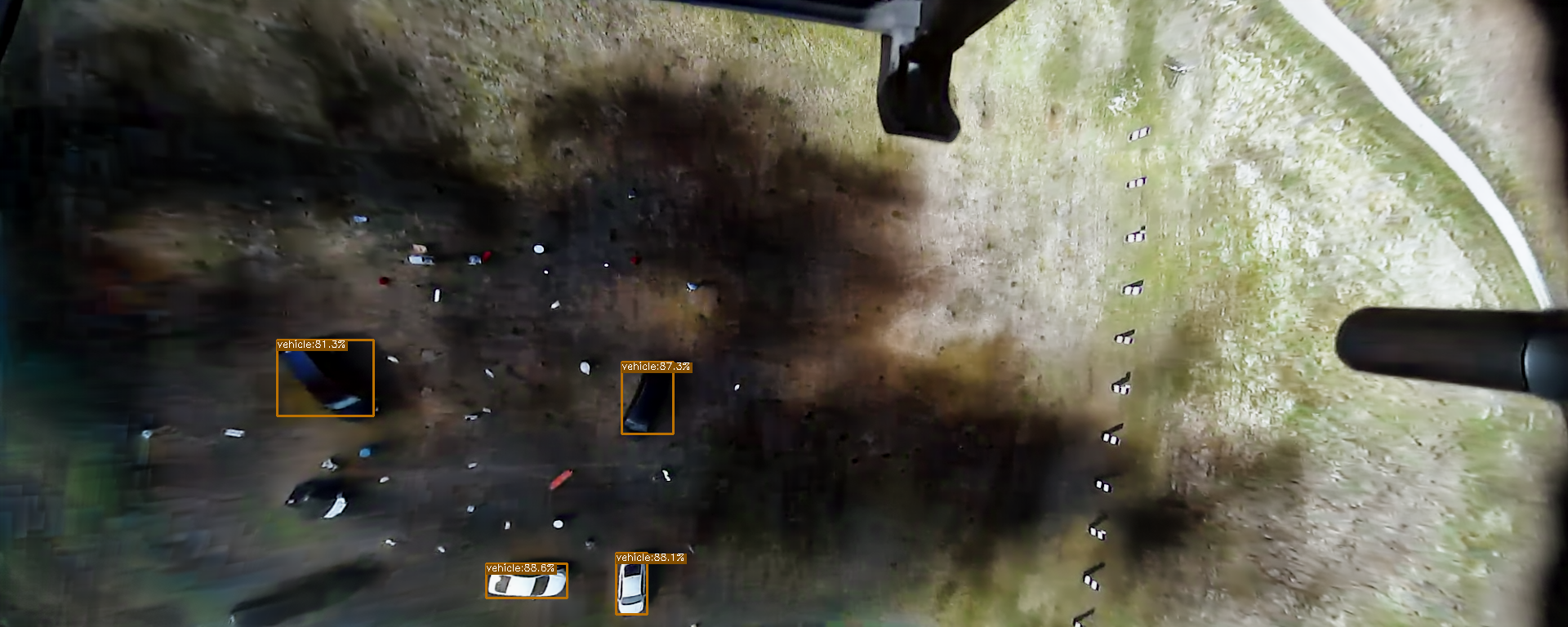}
    }
    \\
    \subfloat
    {
       \includegraphics[width=0.32\linewidth]{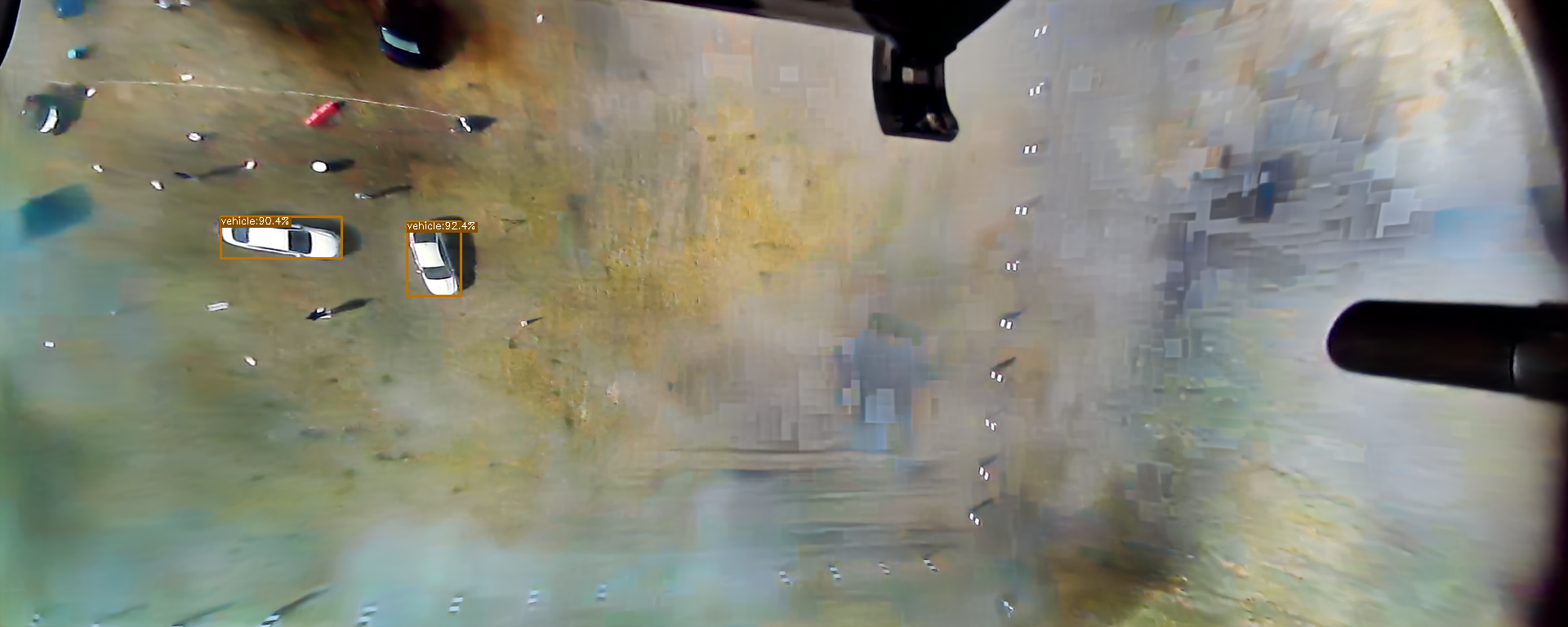}
    }
    \subfloat
    {
       \includegraphics[width=0.32\linewidth]{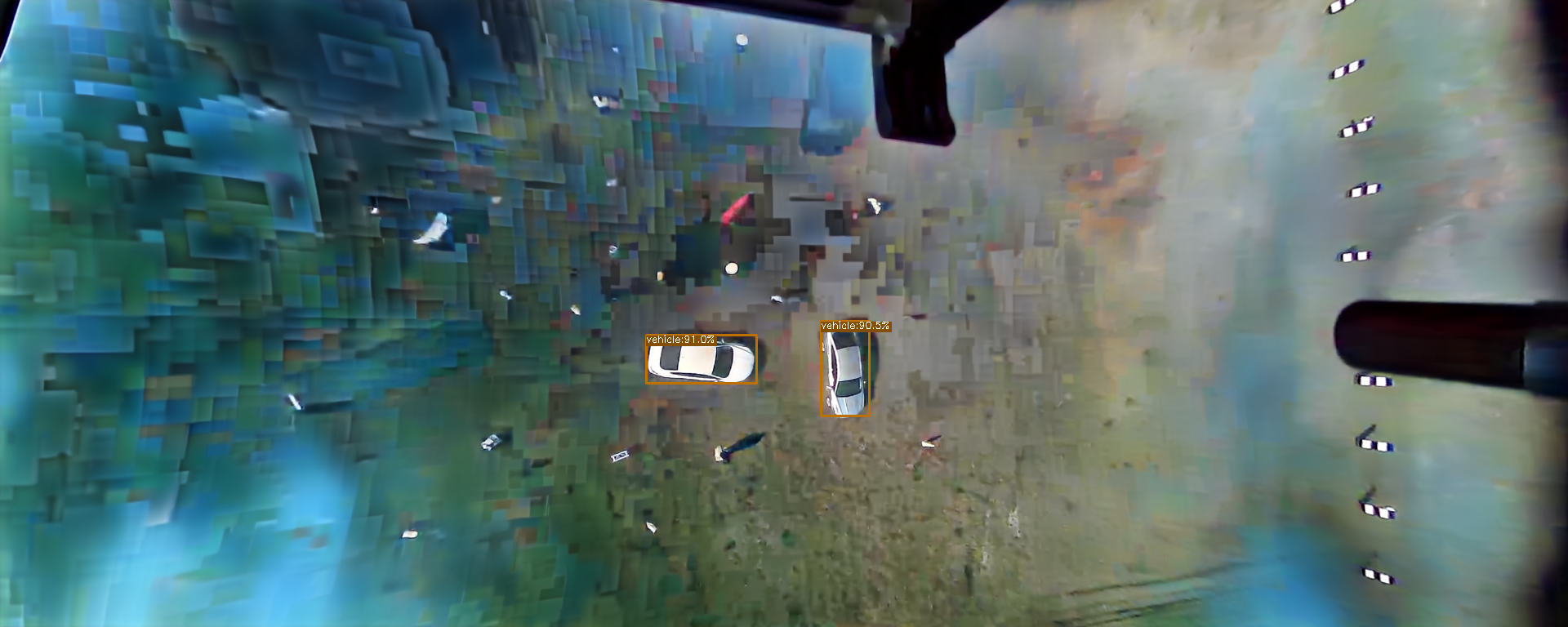}
    }
    \subfloat
    {
       \includegraphics[width=0.32\linewidth]{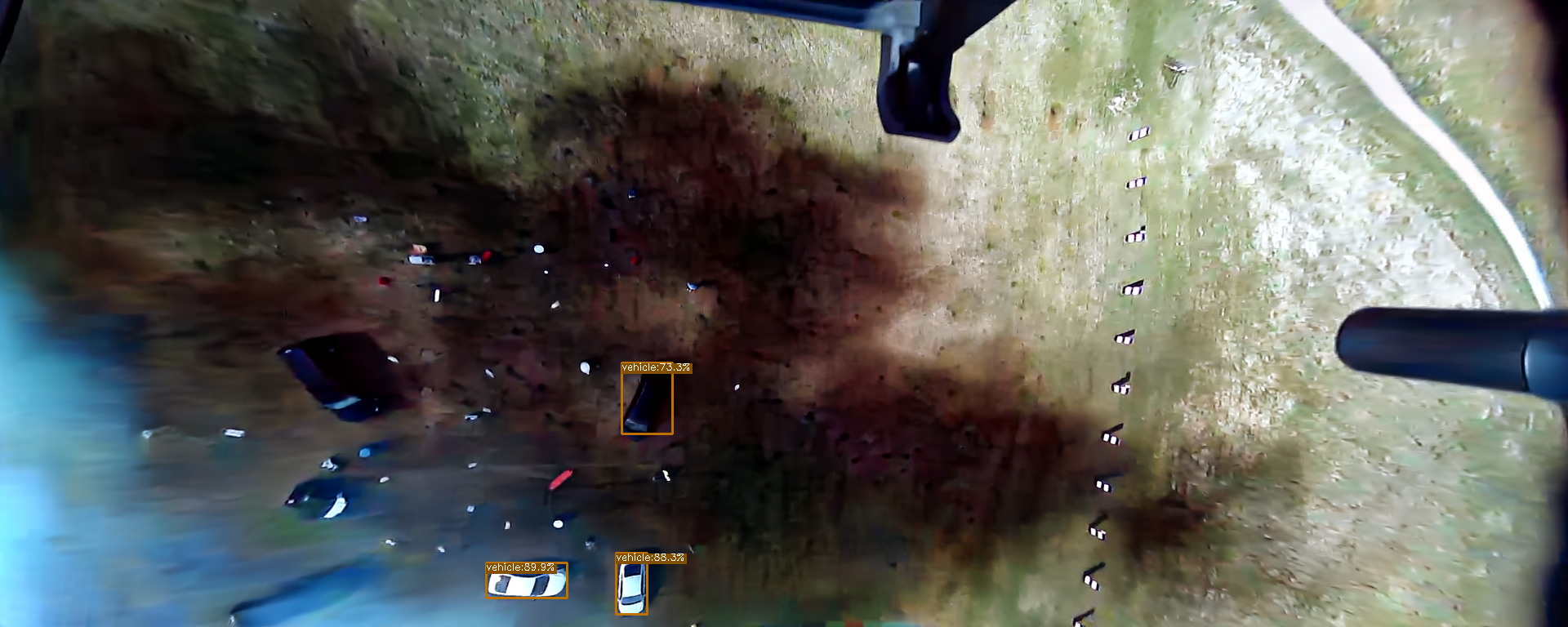}
    }
    \\
    \subfloat
    {
       \includegraphics[width=0.32\linewidth]{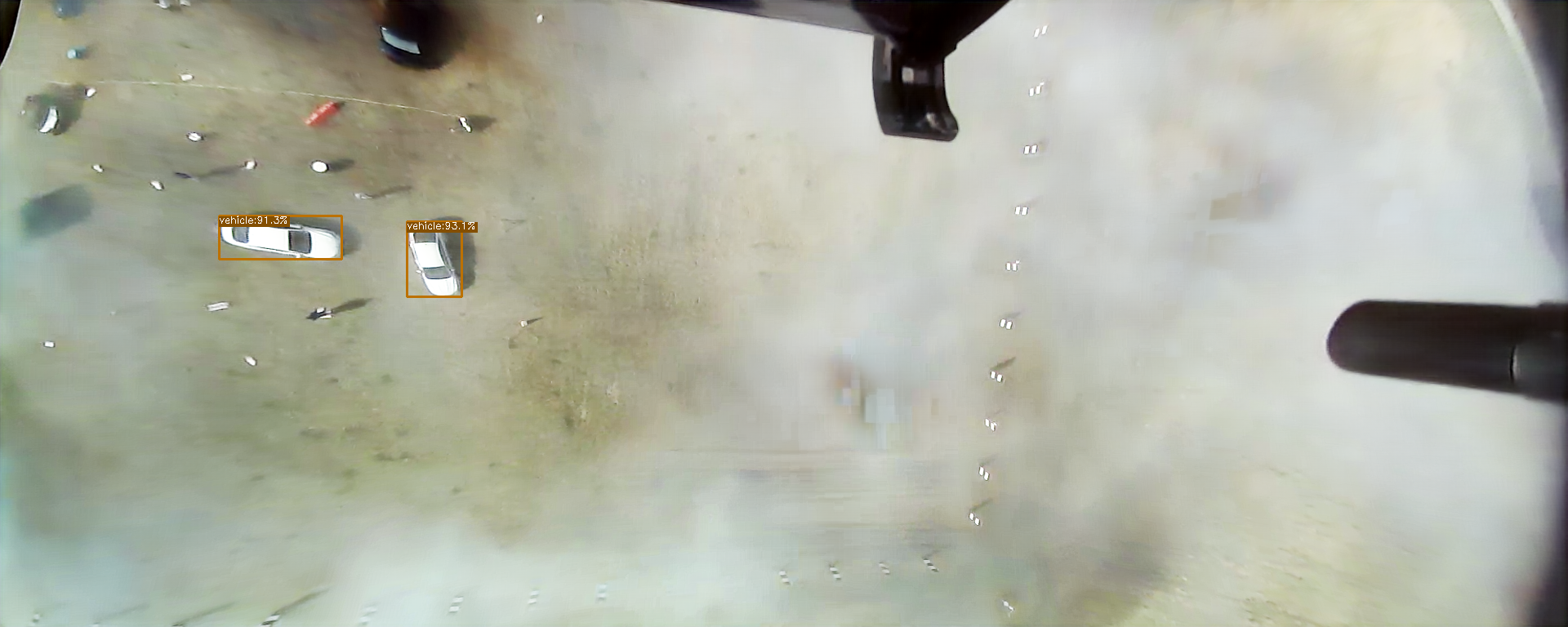}
    }
    \subfloat
    {
       \includegraphics[width=0.32\linewidth]{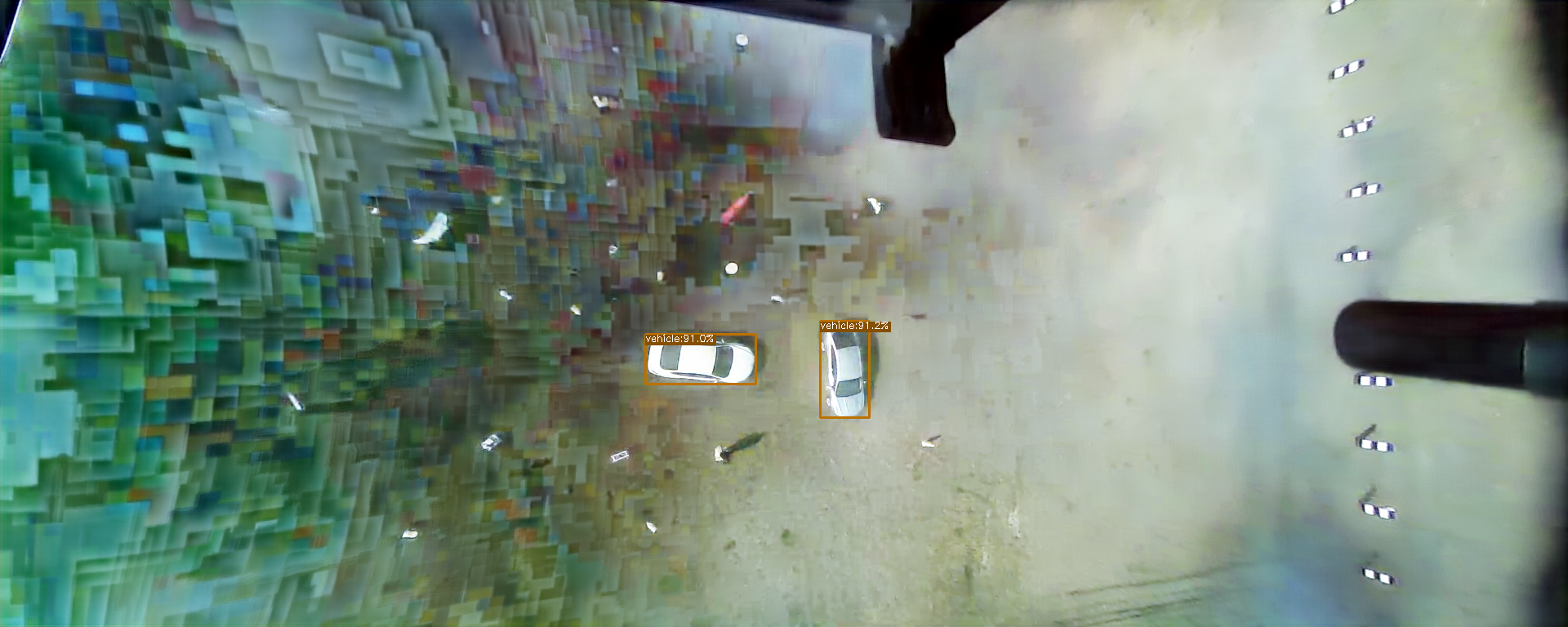}
    }
    \subfloat
    {
       \includegraphics[width=0.32\linewidth]{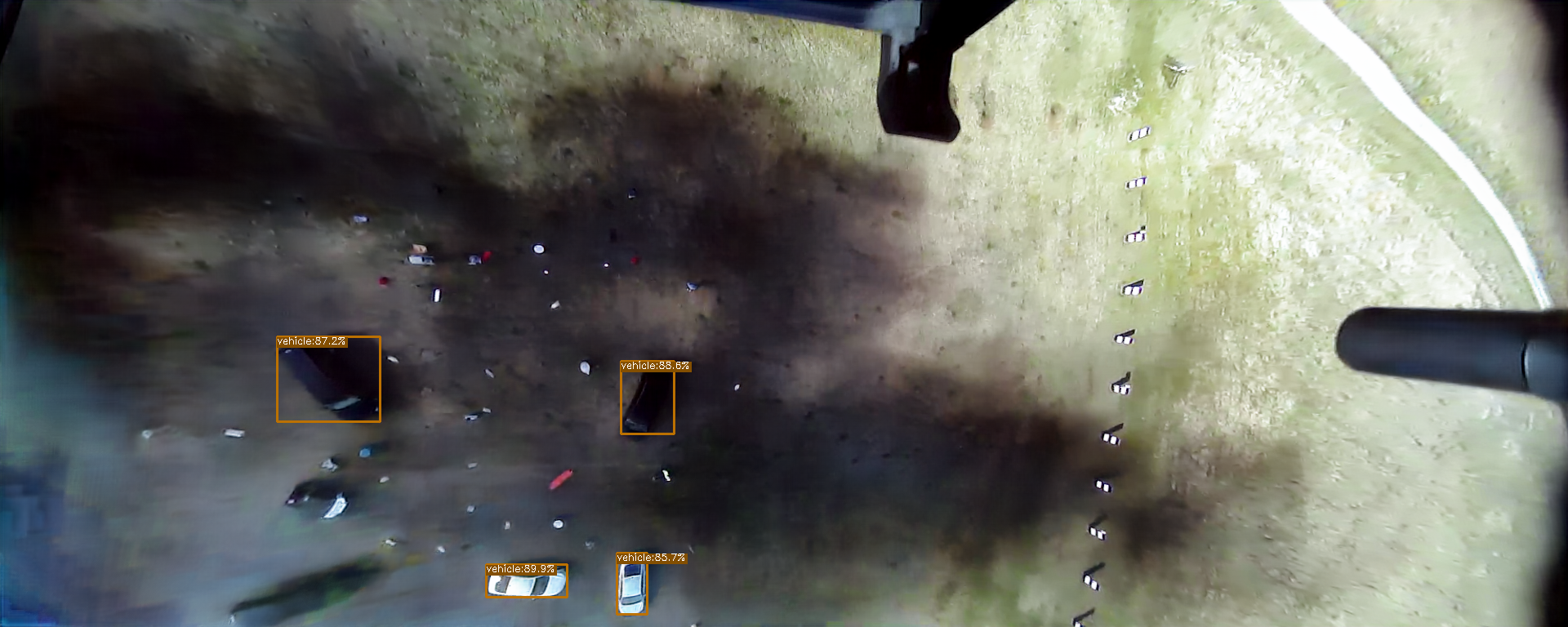}
    }
    \\
     \subfloat
    {
       \includegraphics[width=0.32\linewidth]{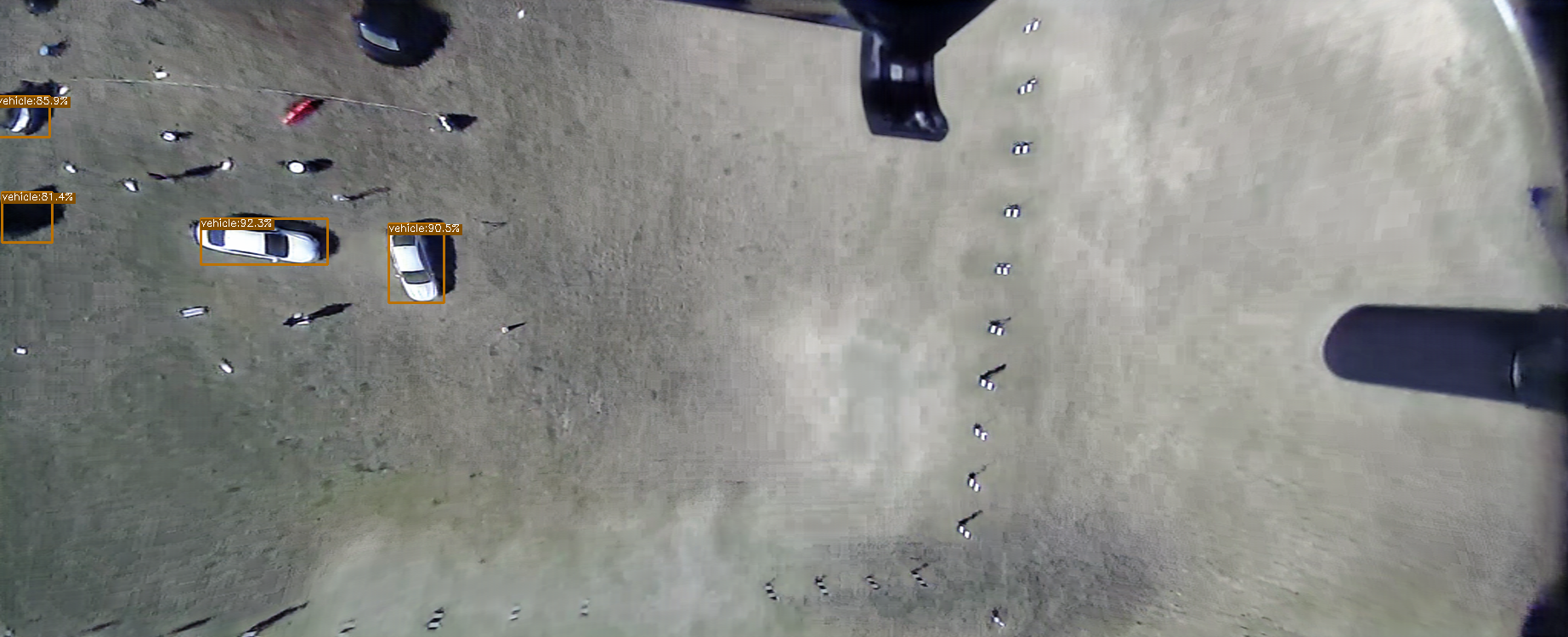}
    }
    \subfloat
    {
       \includegraphics[width=0.32\linewidth]{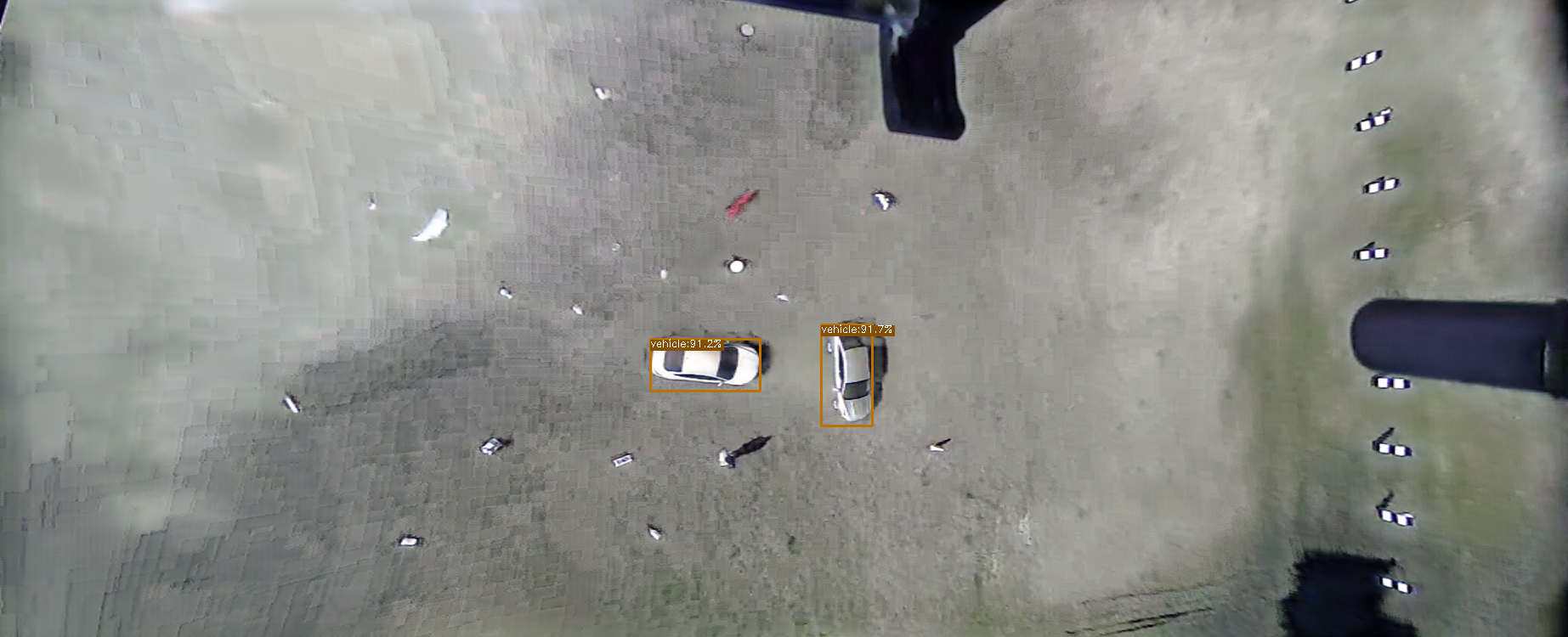}
    }
    \subfloat
    {
       \includegraphics[width=0.32\linewidth]{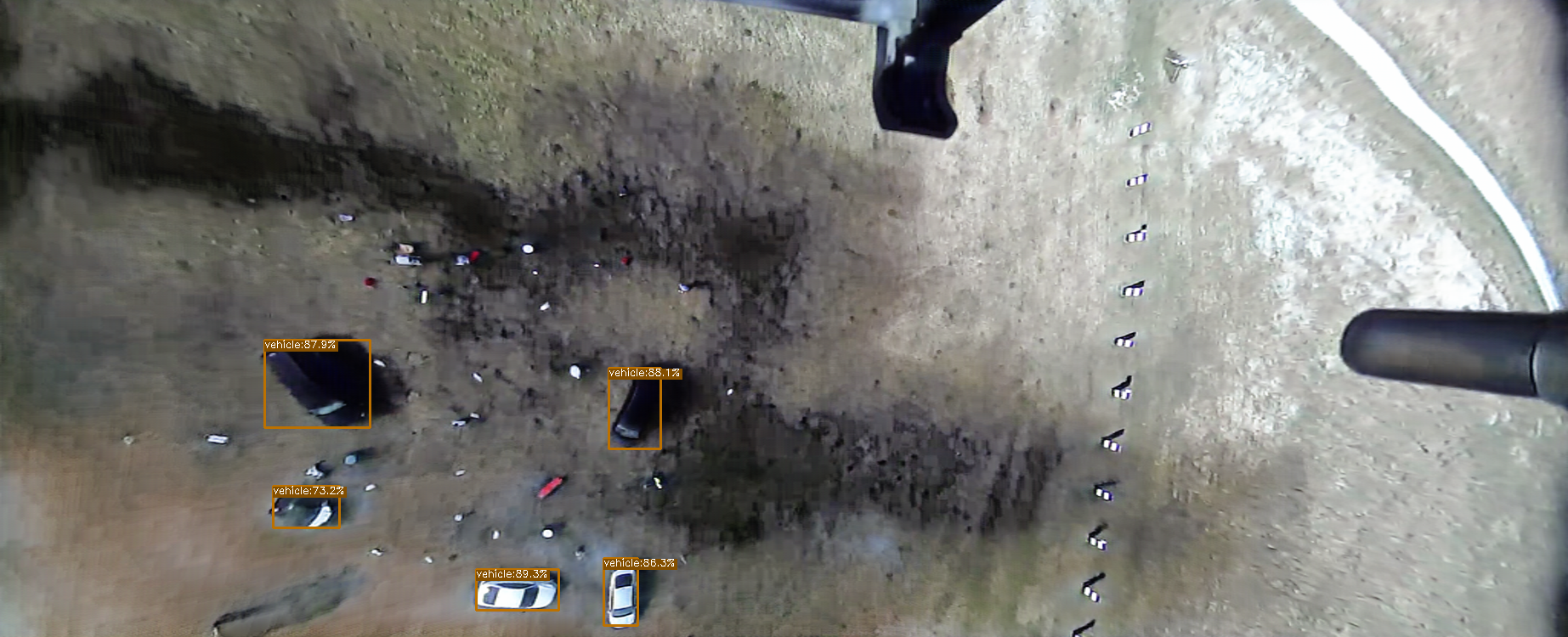}
    }
    \vspace{-0.5em}
    \caption{Examples of detected vehicles on hazy images and dehazed images. All detections are produced from YOLOX-M pretrained on a merged set of VisDrone2019-DET and UAVDT-M. The first row in red frame shows ground-truth bounding-box annotations. The second row in blue frame shows baseline results w/o dehazing. The detected vehicles using different dehazing approaches are arranged from the third row to the last row: FFA-Net, GCANet, MSBDN, SRKT, Trident, DWDehazing, and our proposed Cycle-DehazeNet.
    }
    \label{fig:detection-benchmarks}
\end{figure*}

\begin{figure}
    \centering
    \includegraphics[width=1\linewidth]{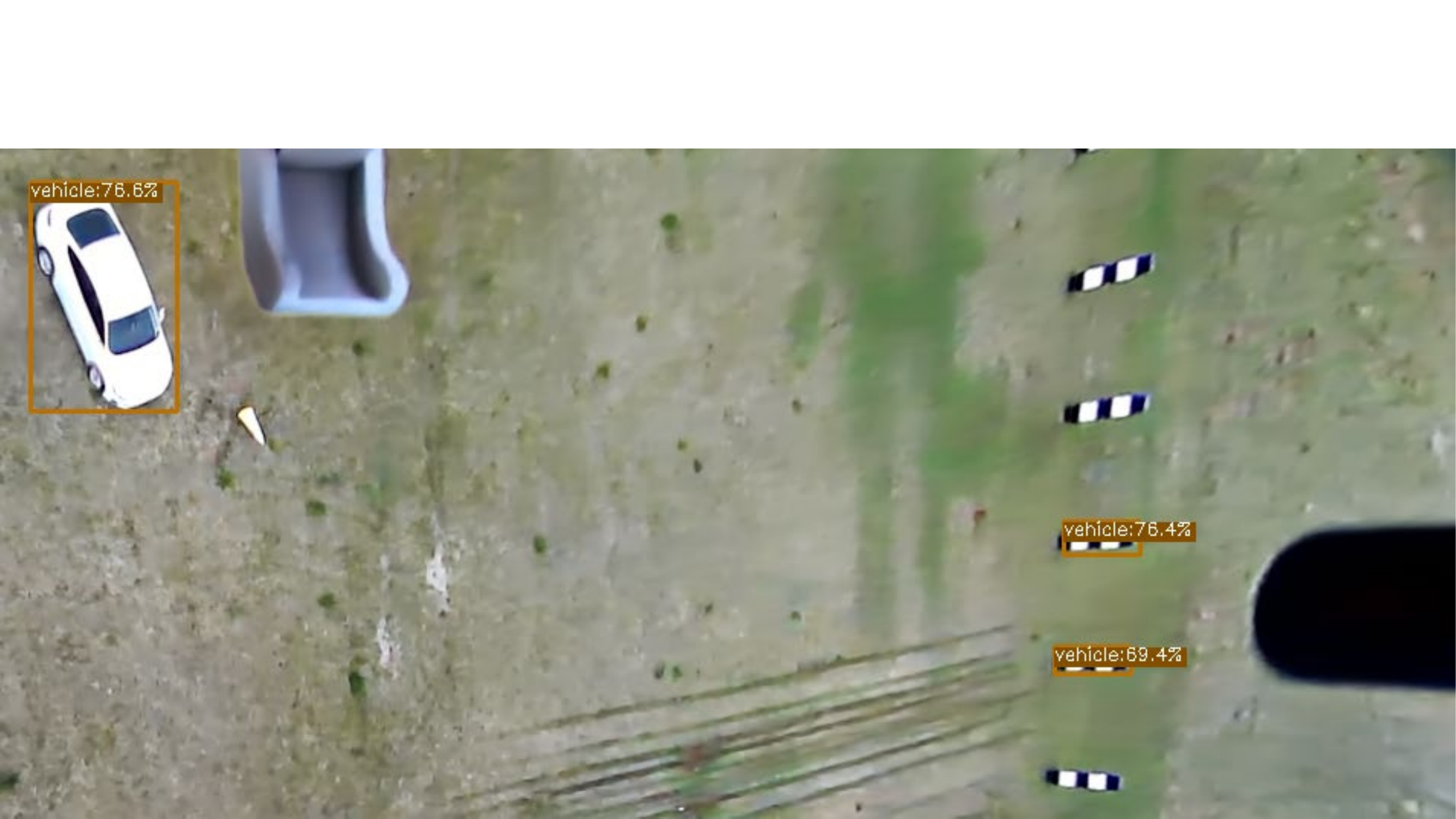}
    \caption{Failed examples on dehazed images (detected by YOLOX). YOLOX always mistakenly detect small contrast boards as vehicles (marked in orange on the middle right).}
    \label{fig:failed}
\end{figure}

\begin{table*}\normalsize
    \centering
    \caption{Vehicle Detection on A2I2-Haze w/o Dehazing. The first row of A2I2-UAV shows experimental results on A2I2-UAV w/o NDFT. The second row of A2I2-UAV$^+$ shows experimental results on A2I2-UAV w/ NDFT. The third row of A2I2-UGV shows experimental results on A2I2-UGV.}
    \resizebox{0.75\linewidth}{!}{
    \begin{tabular}{c|ccc|ccc|ccc|ccc}
        \hline
         & \multicolumn{3}{c|}{YOLOv5} & \multicolumn{3}{c|}{FasterRCNN} & \multicolumn{3}{c|}{CenterNet} & \multicolumn{3}{c}{YOLOX}\\
        \hline
         & AR &  $\text{AP}_{0.5}$ & $\text{AP}_{0.5:0.95}$ & AR &  $\text{AP}_{0.5}$ & $\text{AP}_{0.5:0.95}$ & AR &  $\text{AP}_{0.5}$ & $\text{AP}_{0.5:0.95}$ & AR & $\text{AP}_{0.5}$ & $\text{AP}_{0.5:0.95}$ \\
        \hline
        A2I2-UAV & 44.9 & 53.8 & 42.8 & 41.1 & 55.3 & 38.1 & 60.9 & 69.3 & 51.4 & 41.2 & 49.9 & 39.5\\
        \hline
        A2I2-UAV$^+$ & 46.5 & 56.1 & 44.9 & 43.6 & 57.6 & 40.7 & 63.5 & 71.9 & 53.8 & 44.1 & 51.6 & 41.8\\
        \hline
        A2I2-UGV & 36.5 & 45.5 & 34.2 & 47.8 & 58.0 & 44.3 & 52.9 & 69.3 & 47.9 & 24.0 & 28.7 & 20.9 \\
        \hline
    \end{tabular}
    }
    \label{tab:detection_nodehaze}
\end{table*}

\begin{table*}\normalsize
    \centering
    \caption{Vehicle Detection on A2I2-Haze w/ Dehazing. The upper and lower table shows experiments on A2I2-UAV and A2I2-UGV, respectively. The numbers in black shows the detection score (AR, AP$_{0.5}$, and AP$_{0.5:0.95}$) after applying dehazing on both training and testing images. The numbers in blue shows the relative increase over the detection w/o dehazing. The numbers in red show the relative decrease under the detection w/o dehazing.}
    \resizebox{\linewidth}{!}{
    \begin{tabular}{c|ccc|ccc|ccc|ccc}
        \hline
         \multirow{2}*{A2I2-UAV} & \multicolumn{3}{c|}{YOLOv5} & \multicolumn{3}{c|}{FasterRCNN} & \multicolumn{3}{c|}{CenterNet} & \multicolumn{3}{c}{YOLOX}\\
         \cline{2-13}
         ~ & AR &  $\text{AP}_{0.5}$ & $\text{AP}_{0.5:0.95}$ & AR &  $\text{AP}_{0.5}$ & $\text{AP}_{0.5:0.95}$ & AR &  $\text{AP}_{0.5}$ & $\text{AP}_{0.5:0.95}$ & AR & $\text{AP}_{0.5}$ & $\text{AP}_{0.5:0.95}$ \\
        \hline
        GCANet     &     50.9(+\blue{6.0})    &   59.7(+\blue{5.9})   &   49.0(+\blue{6.2})    & 47.3(+\blue{6.2}) & 61.1(+\blue{5.8}) & 43.5(+\blue{5.4}) & 65.3(+\blue{4.4}) & 75.2(+\blue{5.9}) & 55.7(+\blue{4.3}) & 40.5(-\red{0.7}) & 47.9(-\red{2.0}) & 38.5(-\red{1.0})\\ 
        FFA-Net     &     45.3(+\blue{0.4})    &  53.3(-\red{0.5})   &   42.9(+\blue{0.1})   & 44(+\blue{2.9}) & 54.4(-\red{0.9}) & 37.4(-\red{0.7}) & 61.8(+\blue{0.9}) & 69.6(+\blue{0.3}) & 51.5(+\blue{0.1}) & 40.6(-\red{0.6}) & 48.5(-\red{1.4}) & 38.6(-\red{0.9}) \\ 
        MSBDN     &   49.5(+\blue{4.6})   &   58.8(+\blue{5.0})   &   47.5(+\blue{4.7})  & 47.3(+\blue{6.2}) & 62.8(+\blue{7.5}) & 43.6(+\blue{5.5}) & 64.2(+\blue{3.3}) & 72.5(+\blue{3.2}) & 57.5(+\blue{6.1}) & 42.7(+\blue{1.5}) & 48.8(-\red{1.1}) & 39.0(-\red{0.5}) \\  
        Trident     &     56.2(+\blue{11.3})   &   67.3(+\blue{13.5})    &   53.4(+\blue{10.6})    & 55.1(+\blue{14.0}) & 71.6(+\blue{16.3}) & 49.7(+\blue{11.6}) & 65.5(+\blue{4.6}) & 77.3(+\blue{8.0}) & 57.0(+\blue{5.6}) & 40.1(-\red{1.1}) & 48.3(-\red{1.6}) & 38.4(-\red{1.1}) \\
        
        SRKT    &     52.4(+\blue{7.5})   &   63.5(+\blue{9.7})   &   49.9(+\blue{7.1}) & 50.7(+\blue{9.6}) & 65.7(+\blue{10.4}) & 45.8(+\blue{7.7}) & 62.9(+\blue{2.0}) & 74.1(+\blue{4.8}) & 54.2(+\blue{2.8}) & 39.4(-\red{1.8}) & 45.9(-\red{4.0}) & 37.0(-\red{2.5})   \\
        
        DWDehaze     &    54.6(+\blue{9.7})  &    65.5(+\blue{11.7})   &   52.0(+\blue{9.2})   & 53.1(+\blue{12.0}) & 67.7(+\blue{12.4}) & 48.4(+\blue{10.3}) & 65.5(+\blue{4.6}) & 76.3(+\blue{7.0}) & 56.5(+\blue{5.1}) & \textbf{44.1(+\blue{2.9})} & \textbf{51.6(+\blue{1.7})} & 41.8(+\blue{2.3})  \\
        
        Cycle     &    \textbf{57.1(+\blue{12.2})}  &    \textbf{68.1(+\blue{14.3})}   &   \textbf{53.9(+\blue{11.1})}   & \textbf{56.6(+\blue{15.5})} & \textbf{73.1(+\blue{17.8})} & \textbf{50.1(+\blue{12.0})} & \textbf{66.2(+\blue{5.3})} & \textbf{77.8(+\blue{8.5})} & \textbf{57.7(+\blue{6.3})} & 43.9(+\blue{2.7}) & 51.5(+\blue{1.6}) & \textbf{41.8(+\blue{2.3})}  \\
        \hline
                \hline
         \multirow{2}*{A2I2-UGV} & \multicolumn{3}{c|}{YOLOv5} & \multicolumn{3}{c|}{FasterRCNN} & \multicolumn{3}{c|}{CenterNet} & \multicolumn{3}{c}{YOLOX}\\
         \cline{2-13}
         ~ & AR &  $\text{AP}_{0.5}$ & $\text{AP}_{0.5:0.95}$ & AR &  $\text{AP}_{0.5}$ & $\text{AP}_{0.5:0.95}$ & AR &  $\text{AP}_{0.5}$ & $\text{AP}_{0.5:0.95}$ & AR & $\text{AP}_{0.5}$ & $\text{AP}_{0.5:0.95}$ \\
        \hline
        GCANet     &  38.2(+\blue{1.7})  &  47.5(+\blue{2.0})    &    36.0(+\blue{1.8})   & 54.5(+\blue{6.7}) & 67.6(+\blue{9.6}) & 50.0(+\blue{5.7}) & 54.1(+\blue{1.2}) & \textbf{73.0(+\blue{3.7})} & 49.3(+\blue{1.4}) & 30.2(+\blue{6.2}) & 33.7(+\blue{5.0}) & 24.7(+\blue{3.8})\\
        FFA-Net     &   37.3(+\blue{0.8})      & 46.5(+\blue{1.0})     &    35.0(+\blue{0.8})  & 48.1(+\blue{0.3}) & 58.0(-\red{0.0}) & 44.3(-\red{0.0}) & 52.1(-\red{0.8}) & 68.5(-\red{0.8}) & 47.1(-\red{0.8}) & 24.1(+\blue{0.1}) & 28.6(-\red{0.1}) & 20.8(-\red{0.1}) \\ 
        MSBDN     &   37.1(+\blue{0.6})   & 45.5(-\red{0.0})     & 34.7(+\blue{0.5})    & 50.1(+\blue{2.3}) & 60.9(+\blue{2.9}) & 46.4(+\blue{2.1}) & 53.3(+\blue{0.4}) & 69.7(+\blue{0.4}) & 48.4(+\blue{0.5}) & 32.0(+\blue{8.0}) & 36.2(+\blue{7.5}) & 27.8(+\blue{6.9}) \\  
        Trident     &   \textbf{42.4(+\blue{5.9})}    &  \textbf{53.4(+\blue{7.9})}     &     \textbf{40.2(+\blue{6.0})}  & 53.8(+\blue{6.0}) & 66.5(+\blue{8.5}) & 49.5(+\blue{5.2}) & \textbf{54.2(+\blue{1.3})} & 72.7(+\blue{3.4}) & 48.8(+\blue{0.9}) & 35.0(+\blue{11.0}) & 42.7(+\blue{14.0}) & 31.1(+\blue{10.2}) \\
        SRKT    &    39.9(+\blue{3.4})    & 49.5(+\blue{4.0})     & 36.9(+\blue{2.7})   & 53.6(+\blue{5.8}) & 65.8(+\blue{7.8}) & 49.0(+\blue{4.7}) & 53.4(+\blue{0.5}) & 71.6(+\blue{2.3}) & 48.7(+\blue{0.8}) & 35.7(+\blue{11.7}) & 43.0(+\blue{14.3}) & 31.4(+\blue{10.4})   \\
        DWDehaze     & 36.4(-\red{0.1})  & 45.5(-\red{0.0})   &   34.2(-\red{0.0})   & 50.6(+\blue{2.8}) & 62.0(+\blue{4.0}) & 46.6(+\blue{2.3}) & 51.8(-\red{1.1}) & 67.9(-\red{1.4})& 45.9(-\red{2.0}) & 30.0(+\blue{6.0}) & 38.3(+\blue{9.6}) & 26.4(+\blue{5.5})  \\
        Cycle     & 40.6(+\blue{4.1})  & 50.0(+\blue{4.5})   &   37.1(+\blue{2.9})   & \textbf{55.9(+\blue{8.1})} & \textbf{68.7(+\blue{10.7})} & \textbf{50.5(+\blue{6.2})} & 54.1(+\blue{1.2}) & 72.9(+\blue{3.6}) & \textbf{49.3(+\blue{1.4})} & \textbf{36.2(+\blue{12.2})} & \textbf{44.1(+\blue{15.4})} & \textbf{32.2(+\blue{11.3})} \\
        \hline
    \end{tabular}
    }
    \label{tab:A2I2-Haze_Dehaze}
\end{table*}

\begin{figure*}[!htb]
    \captionsetup[subfigure]{labelformat=empty,justification=centering,farskip=0pt,captionskip=1pt}
    \centering
    \subfloat
    {
    \includegraphics[width=0.475\linewidth]{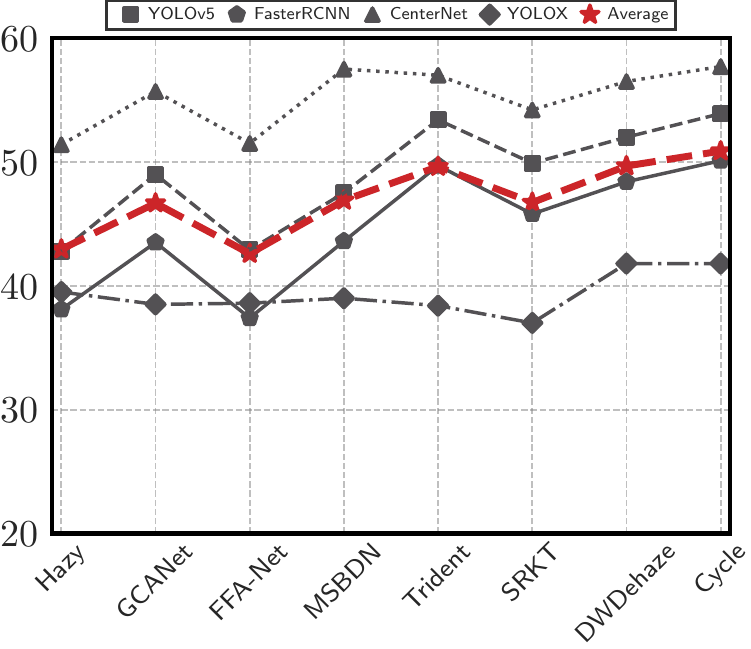}
    }
    \subfloat{
    \includegraphics[width=0.475\linewidth]{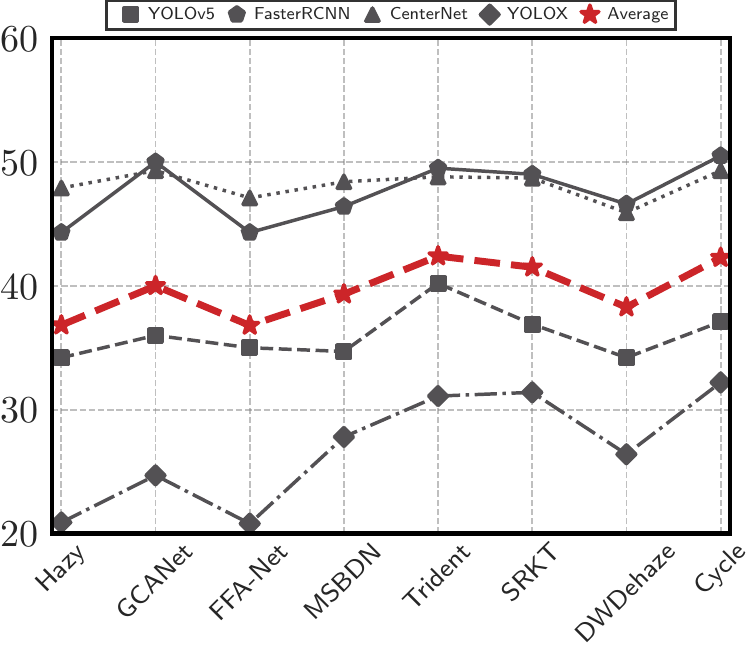}
    }
    \caption{Comparison of different detection and dehazing approaches. The left plot shows detection results on A2I2-UAV, and the right plot shows detection results on A2I2-UGV. Scatter points in these two plots correspond to AP$_{0.5:0.95}$ in Table~\ref{tab:A2I2-Haze_Dehaze}. It gives two perspective: object detectors (the red line) and dehazing approaches (the gray lines). In the left plot, the dashed gray line with triangles is clearly above the remaining three gray lines, which indicates that CenterNet is the best detector in non-homogeneous haze. In the dashed red line with stars, our proposed Cycle-DehazeNet has the best dehazing performance, indicated by the largest improvement on the four detectors' performance.}
    \label{fig:detection_dehaze}
\end{figure*}

\begin{figure}[!htb]
    \captionsetup[subfigure]{labelformat=empty,justification=centering,farskip=1pt,captionskip=1pt}
    \centering
    \subfloat
    {
    \includegraphics[width=1.0\linewidth]{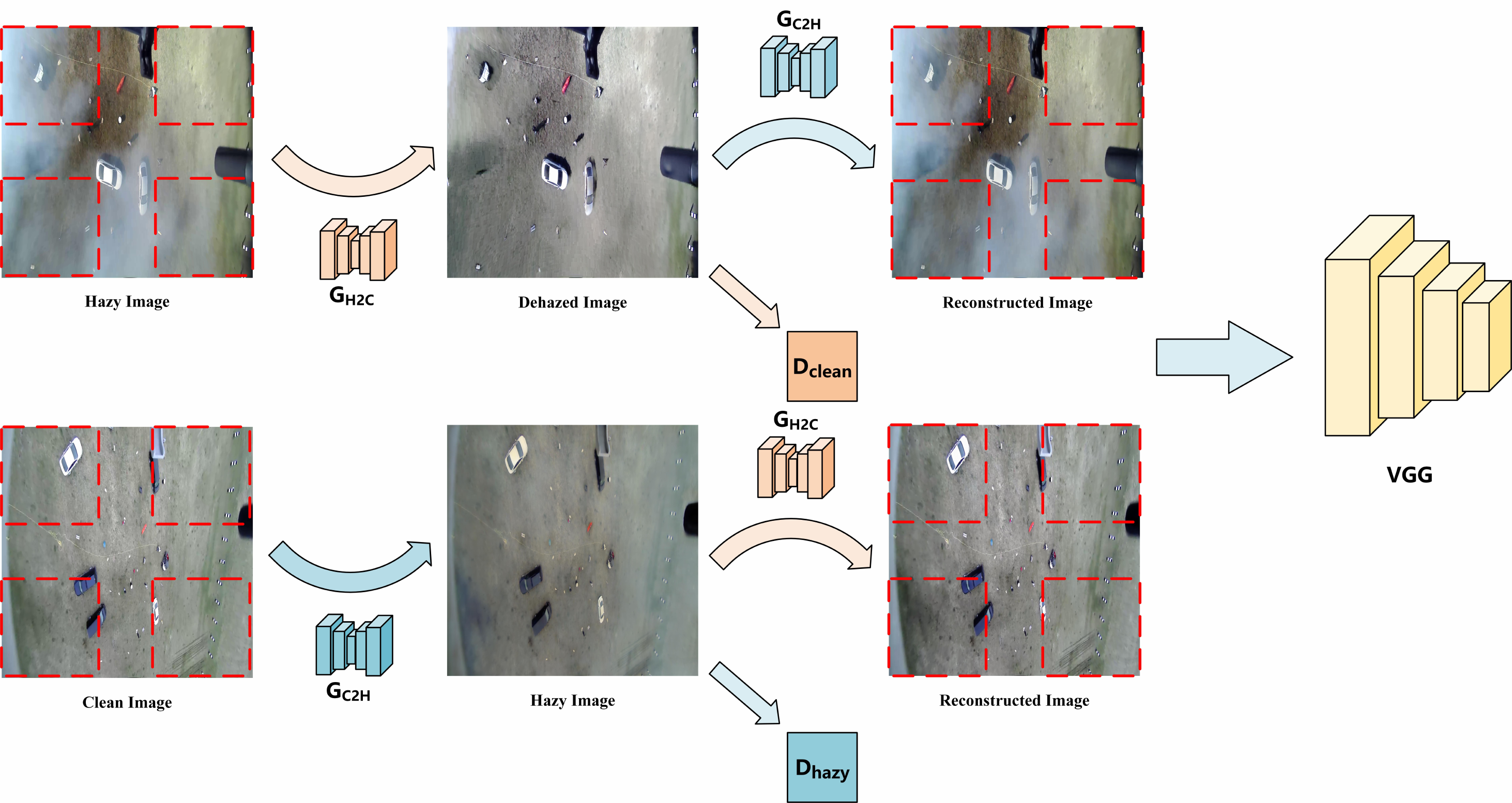}
    }
    \caption{Overview of Cycle-DehazeNet. $\rm G_{H2C}$ and $\rm G_{C2H}$ are generators, while $\rm D_{clean}$ and $\rm D_{hazy}$ are discriminators. Note that $\rm G_{H2C}$ translates image from hazy domain to haze-free domain while $\rm G_{C2H}$ translates image from haze-free domain to hazy domain. $\rm D_{clean}$ distinguishes real haze-free images from fake ones. $\rm D_{hazy}$ tells real hazy images from fake ones. VGG is used to give feature-level perceptual loss supervision.}
    \label{fig:cycle-dehaze}
\end{figure}

\section{Quantitative Evaluation}
Within A2I2-Haze dataset, we created two separate subsets, A2I2-UAV and A2I2-UGV, dedicated to UAV and UGV sets, respectively. 
A2I2-UAV includes a training set UAV-train with $224$ pairs of hazy and corresponding haze-free images, and an additional $240$ haze-free images. The testing set of UAV-test has $119$ hazy images. The training set of A2I2-UGV - UGV-train has $50$ pairs of hazy and corresponding haze-free images, and an additional $200$ haze-free images. UGV-test has $200$ hazy images.

\subsection{Object Detection}\label{det_eval}
\subsubsection{Detection Approaches}
We use four state-of-the-art detectors that are widely used both in industry and academia: (a) \textit{YOLOv5}~\cite{yolov5}, (b) \textit{YOLOX}~\cite{ge2021yolox}, (c) \textit{Faster R-CNN}~\cite{ren2016faster}, and (d) \textit{CenterNet}~\cite{zhou2019objects} to evaluate the proposed A2I2-Haze dataset. For experiments on A2I2-UAV, all detectors are pretrained on a merged set of VisDrone2019-DET~\cite{cao2021visdrone} and UAVDT-M~\cite{du2018unmanned} and adopt official COCO-based implementation. For experiments on A2I2-UGV, all detectors are pretrained on Cityscapes~\cite{cordts2016cityscapes} and adopt official COCO-based implementation.

\subsubsection{Results and Analyses} Table~\ref{tab:detection_nodehaze} shows the performance of various object detectors without dehazing on both A2I2-UAV and A2I2-UGV subsets of A2I2-Haze. In the first row, we infer that object detectors can be ordered by their detection performance on A2I2-UAV as follows: $\text{CenterNet} > \text{YOLOv5} > \text{YOLOX} > \text{FasterRCNN}$. In the second row, we show that utilizing the collected altitude metadata to learn an altitude variation robust detector by NDFT~\cite{wu2019delving} could improve the detection performance for all the four detectors. For A2I2-UGV (the third row), the detectors can be ordered based on their performance as $\text{CenterNet} > \text{FasterRCNN} > \text{YOLOv5} > \text{YOLOX}$. Among the four detectors, YOLOX consistently gives the worst performance on both A2I2-UAV and A2I2-UGV. It frequently mistakes small contrast boards as vehicles, as shown in Fig.~\ref{fig:failed}.
\subsection{Dehazing}
\subsubsection{Baseline Approaches}
We benchmark three state-of-the-art homogeneous dehazing approaches: (I) \textit{GCANet}~\cite{chen2019gated}, (II) \textit{FFA-Net}~\cite{qin2020ffa}, and (III) \textit{MSBDN}~\cite{dong2020multi}. In addition, we also benchmark another set of three state-of-the-art non-homogeneous dehazing approaches: (IV) \textit{SRKT}~\cite{chen2021srktdn}, (V) \textit{DWDehaze}~\cite{fu2021dw}, and (VI) \textit{Trident-Dehazing}~\cite{liu2020trident}. All dehazing models are pre-trained using official implementations.

\subsubsection{Proposed Cycle-DehazeNet}
Inspired by CycleGAN~\cite{engin2018cycle,you2019ct,zhu2017unpaired}, we propose a dehazing model that could be trained for image translation tasks on the hazy image domain and haze-free image domain without paired training samples. 
Cycle-DehazeNet is trained on $512\times 512$ patches (CycleGAN's constraint). It takes $512 \times 512$ as input resolution and restores $512 \times 512$ output images to minimize computational costs. In the inference, we first pad the input HR image from $1845 \times 1500$ to $2048 \times 1536$, so that the width and height are divisible by $512$. Then we extract overlapping $512 \times 512$ patches at a stride of $256$. Each target pixel $p_t$ in the image may be covered by multiple patches. Last, the target pixel value is computed as a weighted sum of all the collided pixels $p_i$ from different patches, where the weight is measured by the distance between the pixel and the patch center. The weight is decayed by a Gaussian distribution with controllable sigma parameters. 
It could be mathematically expressed below: 
\begin{equation}
    p_t = \frac{\sum_{i=1}^N w_i p_i}{\sum_{i=1}^N w_i},
\end{equation}
where $N$ is the number of patches that cover the subject pixel and $w$ follows a Gaussian distribution w.r.t. the geometric distance between the patch center and the target pixel.
Following \cite{engin2018cycle}, Cycle-DehazeNet has two losses: cycle-consistency loss~\cite{zhu2017unpaired} and perceptual loss~\cite{johnson2016perceptual}.
Cycle-consistency loss relieves the constraint of paired training samples. Perceptual loss remedies the local texture information that is heavily corrupted by haze and preserves the original image structure by analyzing the high and low-level features extracted from VGG Net.
%
% We observe that the hazy area is typically further away from the image center in A2I2-Haze.
%
We observe that the hazy areas in our dataset are mostly distributed around four corners instead of the central part. Based on that, we would like to force our proposed method to more \textbf{focus} on haze rather than clean parts. Thus the cropped four $512 \times 512$ patches from the corners of each image are used as ``focus'' (red bounding boxes in Fig.~\ref{fig:cycle-dehaze}). We emphasize more on the hazy area than the whole image by assigning larger weights for the ``focus'' patch in the final loss.

\subsubsection{Evaluation Metric}
We use object detection score on the dehazed images as the metric to quantitatively measure restoration capability of the dehazing technique on image semantics. We use the same four detectors as in Section.~\ref{det_eval} (a) \textit{YOLOv5}, (b) \textit{YOLOX}, (c) \textit{Faster R-CNN}, and (d) \textit{CenterNet}.
\subsubsection{Results and Analyses}
Fig.~\ref{fig:detection-benchmarks} shows detection results on original hazy images and dehazed images. Fig.~\ref{fig:detection_dehaze} shows AP$_{0.5:0.95}$ score as metric to compare various detectors and various dehazing approaches. Fig.~\ref{fig:detection_dehaze} gives two perspective: object detectors (the red line) and dehazing approaches (the gray lines). The red line shows the averaged AP$_{0.5:0.95}$ across the four detectors for different dehazing approaches. The gray lines show the AP$_{0.5:0.95}$ across different dehazing approaches for the four detectors.
Fig.~\ref{fig:detection-benchmarks} shows the visual effect of various dehazing approaches on three randomly selected images. Among the seven dehazing approaches used, FFA-Net leads to the worst detection performance, and our proposed Cycle-DehazeNet gives the best mAP on YOLOv5, Faster R-CNN and CenterNet. In general, we observe that non-homogeneous dehazing algorithms have better mAP compared to the homogeneous dehazing algorithms.
\section{Discussions and Future Works}
A2I2-Haze being the first attempt to develop a real-world hazy dataset with in-situ measurements, we acknowledge several limitations in the data collection procedure. We propose the following research directions to address the gaps:
\begin{itemize}[leftmargin=*]

\item \textbf{Object Diversity:} Both UAV and UGV datasets have a limited number of object categories and instances. Improving the diversity in objects will be a crucial step towards developing a more comprehensive dataset. 
\item \textbf{Background Diversity:} A2I2-Haze was primarily collected in two different scenarios: with concrete and grass backgrounds. Including additional scenes with diverse backgrounds will make the dataset a more challenging benchmark. 
\item \textbf{Global Haze Measure:} The smoke measurement for A2I2-Haze was made using two laser transmissometers. These measurements can be highly localized in a non-homogeneous setting. A novel sensor network for grid-based measurement is required to improve the accuracy of the in-situ data. 
\item \textbf{Synchronized Aerial-Ground View:} A2I2-Haze could be extended to a multi-view dataset using air-ground coordination and multi-source image matching.
\item \textbf{Jointly Optimized Dehazing and Detection Pipeline:}
If jointly optimized, low-level image processing could be made beneficial for high-level semantic tasks~\cite{liu2020connecting,li2017aod}. In this study, dehazing can be considered a pre-processing step for the subsequent detection task. Thus, as a future research direction, detection and dehazing could be jointly designed and trained to optimize detection performance in the presence of haze. This could be particularly significant in the case of aerial-view imagery. Furthermore, as this approach removes the constraint of maintaining the aesthetic quality of the image, dehazing could be utilized to purely restore image semantics for object detection.  
\item \textbf{Metadata Utilization:}
The annotated UAV-specific nuisance or metadata in A2I2-Haze could be utilized to improve the robustness of the learned features via adversarial training as discussed in NDFT~\cite{wu2019delving}. In Tab.~\ref{tab:detection_nodehaze}, we have shown some preliminary success by learning a detector robust to altitude variation on A2I2-UAV. Utilizing the haze density metadata to train a dehazing-aware detector better could be a future research direction.
\end{itemize}
\section{Conclusion }
In this paper we develop a challenging UAV and UGV dataset with realistic haze for foundational research in scene understanding in obscured conditions. Advanced smoke-generation and measurement techniques were used to develop this dataset for object detection and dehazing tasks. All images were curated precisely using a combination of both manual and partially-automated techniques. Further, the images are synchronized with haze density measurements and altitude of the UAV. Quantitative evaluation was performed using state-of-the-art object detectors and de-hazing models. Overall the baseline approaches were found to perform poorly on A2I2-Haze and we hope the dataset will serve as a good benchmark for future algorithmic advances. As discussed in the paper, we also plan to address the limitations of A2I2-Haze in future research efforts.

% or


%\appendices
%\section{}

%\section{}

\section*{Acknowledgment}

The authors would like to thank US Army Artificial Innovation Institute (A2I2) for funding the data collection and annotation, and financially supporting graduate students at Texas A \& M University for this project. The authors also acknowledges experimentation support from DEVCOM Chemical \& Biological Center (CBC) during the Data collection process.

\ifCLASSOPTIONcaptionsoff
  \newpage
\fi



%\bibliographystyle{IEEEtran}
\bibliographystyle{IEEEtran}
\bibliography{IEEEabrv}
% {\small
%     \input{bare_jrnl.bbl}
\begin{IEEEbiography}[{\includegraphics[width=1in,height=1.25in,clip,keepaspectratio]{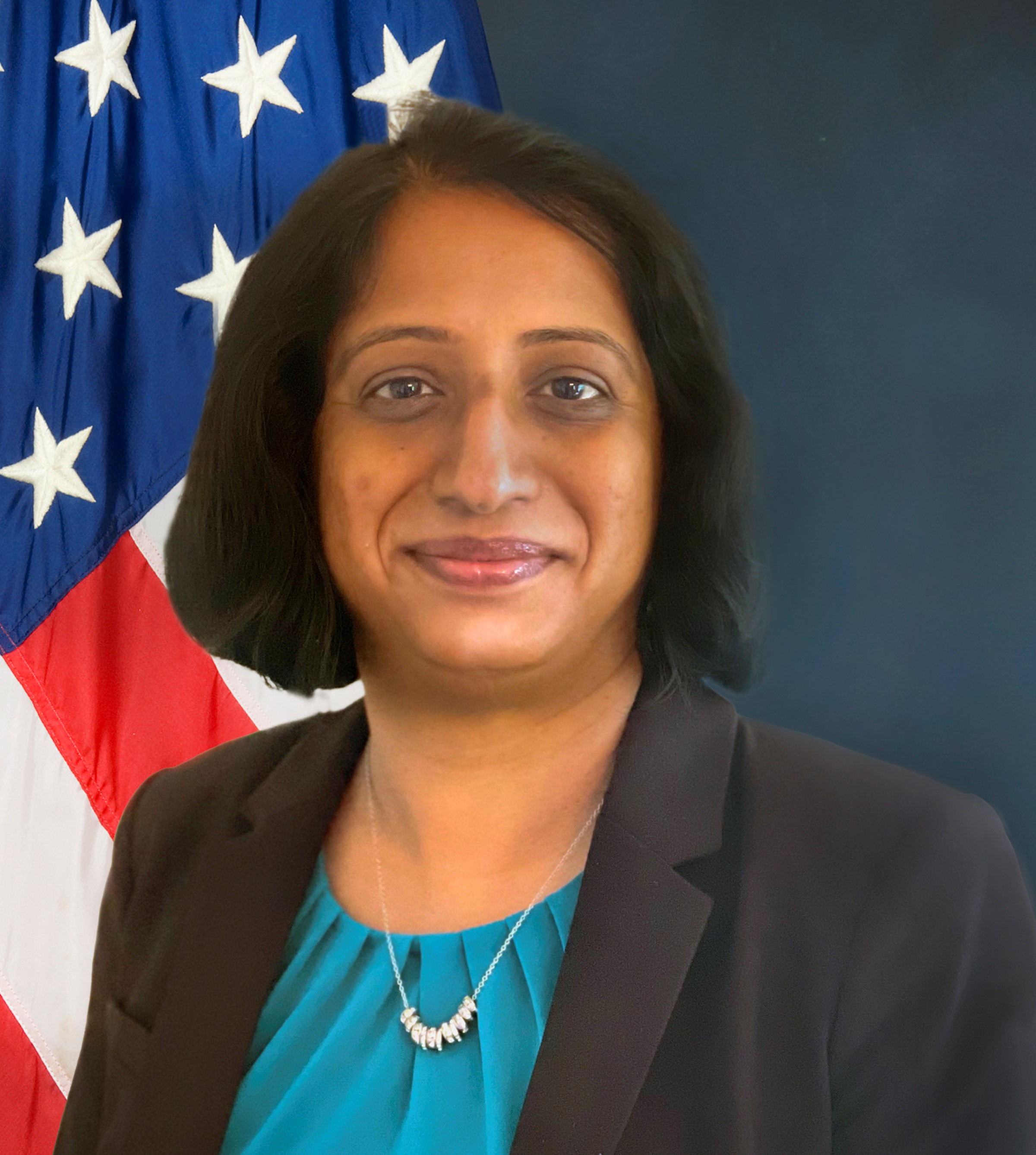}}]{Priya Narayanan} received her Ph.D. from the Mechanical Engineering Department at University of Maryland Baltimore County (UMBC). She was a National Research Council (NRC) Fellow in the Navy Center for Applied Research in Artificial Intelligence (NCARAI) at the Naval Research Laboratory (NRL). Currently she is a Engineering Researcher at DEVCOM Army Research Laboratory. She has led key research programs in aerial and ground robotics including perception integration activities for the final capstone experimentation of Robotic Collaborative Technology Alliance (Robotic CTA) Program.
\end{IEEEbiography}

\begin{IEEEbiography}[{\includegraphics[width=1in,height=1.25in,clip,keepaspectratio]{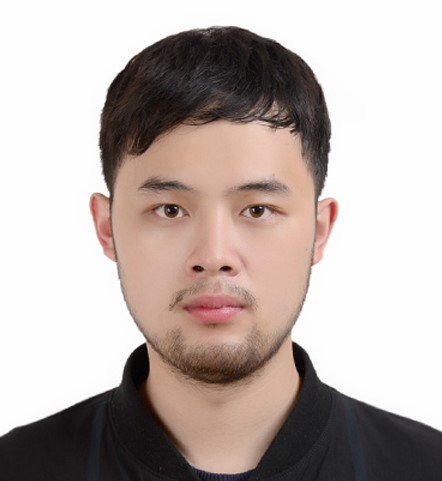}}]{Xin Hu}
received the B.Eng. from Huazhong University of Science and Technology in 2019 and M.S.
from Texas A\&M University in 2021. He is currently a Ph.D. student advised by Dr. Zhengming (Allan) Ding at Tulane University. His research interests include decision-making in autonomous vehicles, video understanding, and multi-modality.
\end{IEEEbiography}

\begin{IEEEbiography}[{\includegraphics[width=1in,height=1.25in,clip,keepaspectratio]{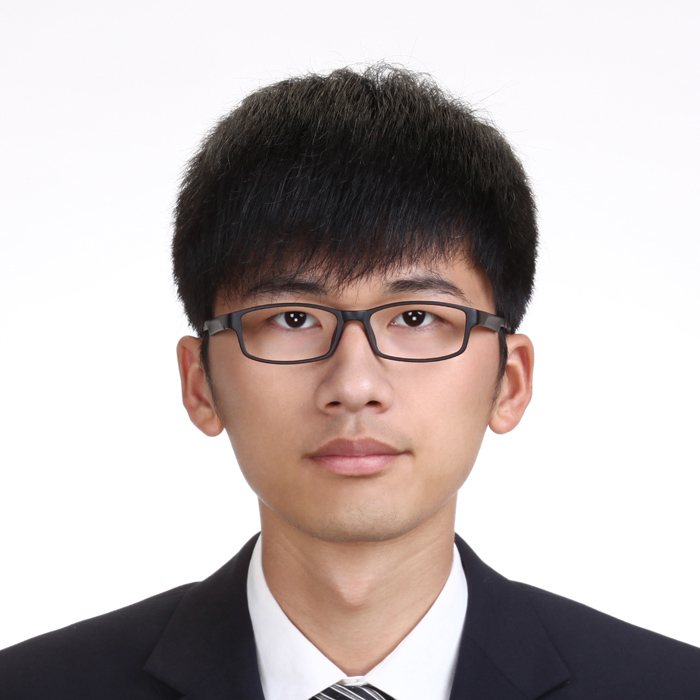}}]{Zhenyu Wu}
received the B.Eng. from Shanghai Jiao Tong University in 2015 and Ph.D. from Texas A\&M University in 2021. His Ph.D. advisor is Dr. Zhangyang Wang. His research interests include object detection, video understanding, efficient vision on embedded systems, and adversarial machine learning.
He is currently a researcher at Wormpex AI Research. He is closely working with Dr. Zhou Ren, Dr. Yi Wu and Dr. Gang Hua.
\end{IEEEbiography}

%\newpage

\begin{IEEEbiography}[{\includegraphics[width=1in,height=1.25in,clip,keepaspectratio]{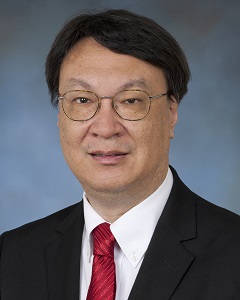}}] {Matthew Thielke}
received his B.S. in Physics from Virginia Tech. He worked at the DEVCOM C5ISR’s Night Vision and Electronic Sensors Directorate in Electro-Optic measurements and processing. Currently at DEVCOM U.S. Army Research Laboratory, he has extensive measurements experience with Electro-Optical sensors including infrared imagers, and hyper spectral imagers. He has led data collection efforts in thermal and visible face recognition including a ``Large-Scale, Time-Synchronized Visible and Thermal Face Dataset.''
\end{IEEEbiography}

\begin{IEEEbiography}[{\includegraphics[width=1in,height=1.25in,clip,keepaspectratio]{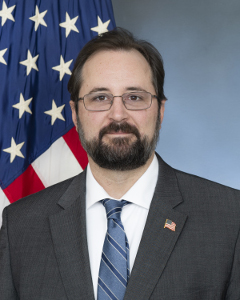}}] {John G Rogers}
received his Ph.D. in Robotics at the Georgia Institute of Technology, his masters in Computer Science at Stanford, and his masters and bachelors degrees in Electrical and Computer Engineering at Carnegie Mellon. He is currently a senior research scientist at the US DEVCOM Army Research Laboratory (ARL). He is currently leading research efforts in coordinated tactical maneuver in complex and contested terrain for teams of ground robots at ARL.
\end{IEEEbiography}

\begin{IEEEbiography}[{\includegraphics[width=1in,height=1.25in,clip,keepaspectratio]{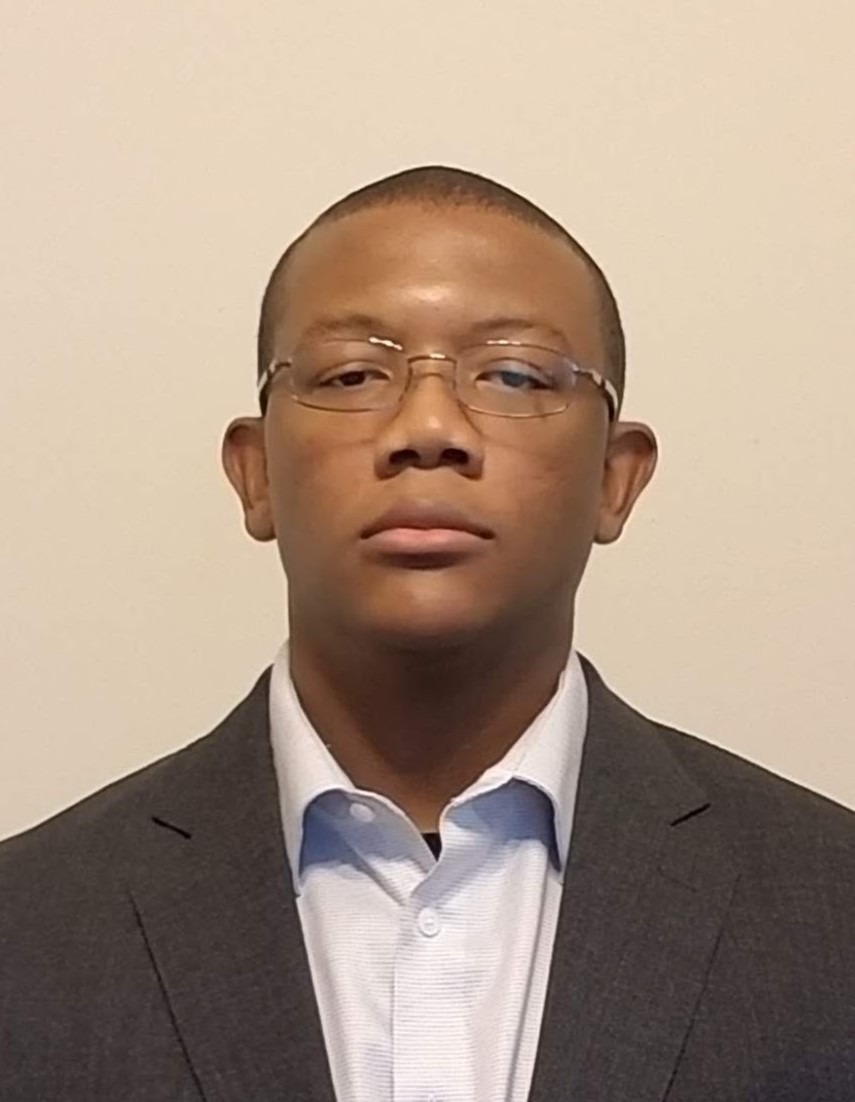}}]{Andre V. Harrison}
received his Ph.D. in Electrical and Computer Engineering from Johns Hopkins University, his M.E. and B.E. in Electrical and Computer Engineering from Cornell University.  He is currently a Computer Engineer with the US DEVCOM Army Research Laboratory (ARL). He has led key research efforts in visual perception for ground vehicles as part of the A.I. for Maneuver and Mobility Essential Research Program.  He has also led research projects estimating visual salience and predicting user state.  
\end{IEEEbiography}

\begin{IEEEbiography}[{\includegraphics[width=1in,height=1.25in,clip,keepaspectratio]{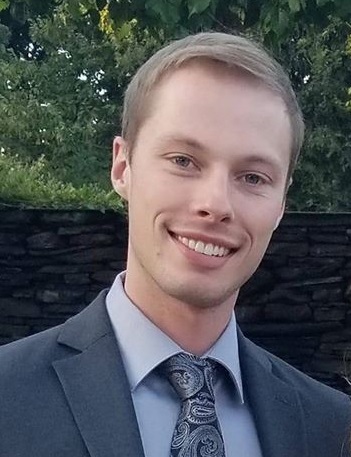}}] {John D'Agostino}
received his B.S. from The Pennsylvania State University in Mechanical Engineering. He is currently a Mechanical Engineer with the U.S Army DEVCOM Chemical and Biological Center in the Research and Technology Directorate. He leads a variety of countermeasure activities including the testing of sensors in various degraded battlefield conditions.   
\end{IEEEbiography}

\begin{IEEEbiography}[{\includegraphics[width=1in,height=1.25in,clip,keepaspectratio]{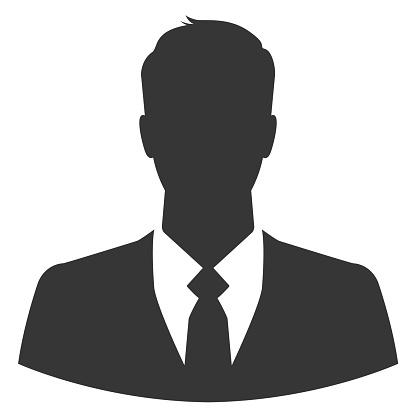}}] {James D Brown}
is a Mechanical Engineer and contractor supporting the US Army DEVCOM-CBC Advanced Design and Manufacturing division. He has supported a variety of UAV and UGV projects from design to implementation, and served as the primary ground control station operator during DVE testing.
\end{IEEEbiography}

\begin{IEEEbiography}[{\includegraphics[width=1in,height=1.25in,clip,keepaspectratio]{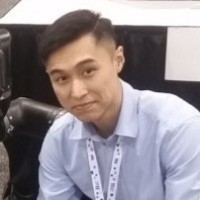}}] {Long Quang}
acquired a Bachelor of Science in Electrical Engineering at the University of Texas at Dallas. He is currently an Electronics Engineer with the U.S. Army DEVCOM Computational and Information Sciences Directorate, where he actively facilitates robotics research and systems integration.
\end{IEEEbiography}

\begin{IEEEbiography}[{\includegraphics[width=1in,height=1.25in,clip,keepaspectratio]{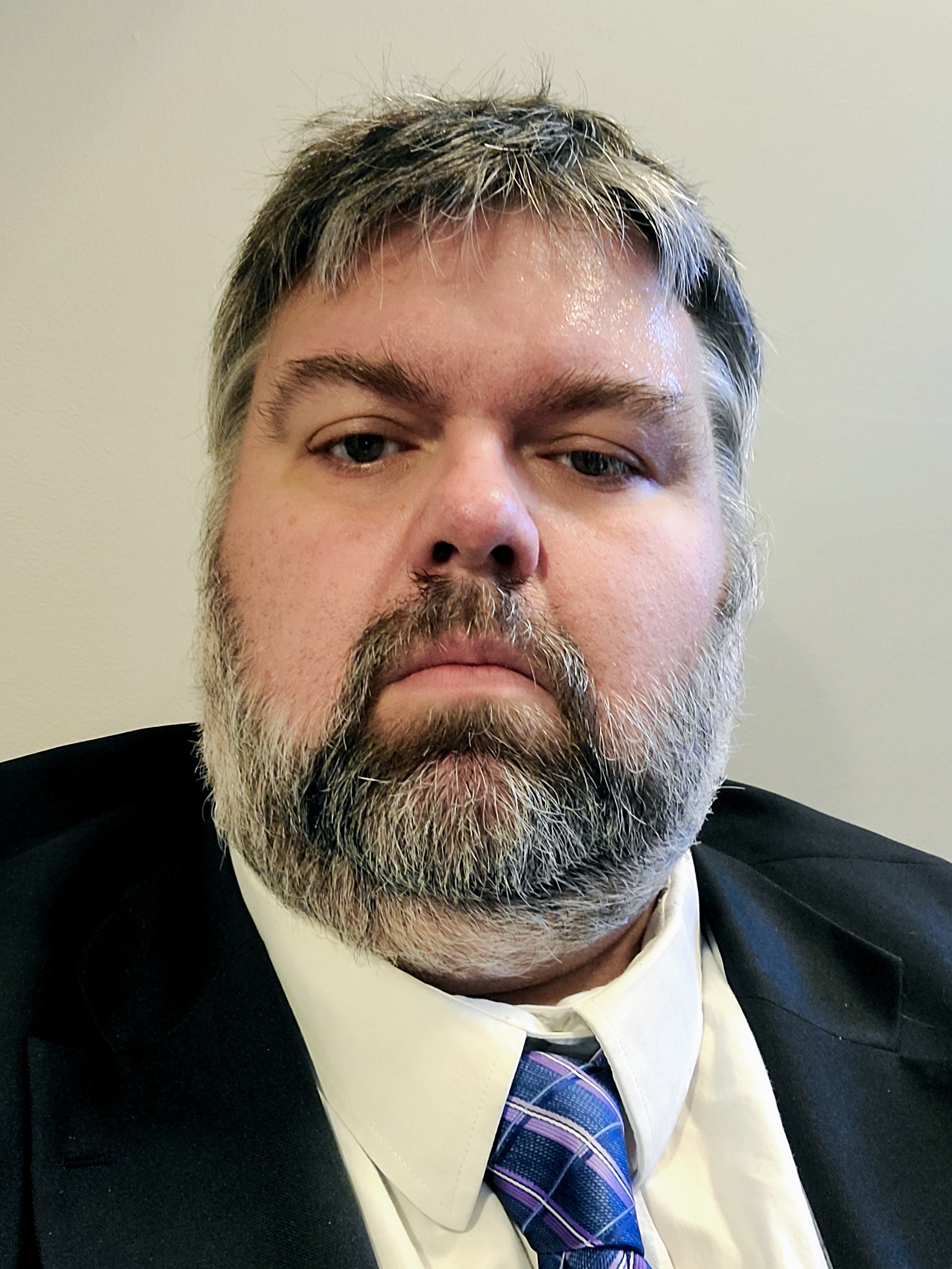}}] {James Uplinger}
received his doctorate and masters from the Physics
Department at the University of Arkansas, and bachelors from the Physics
Department at the Rochester Institute of Technology (RIT). Currently he is
an Associate Principal Data Scientist at Huntington Ingalls Industries, supporting DEVCOM Army Research Laboratory. He has led
research programs in synthetic infrared (IR) image generation, Radio
Frequency (RF) side channel analytics, and biotechnology development.
\end{IEEEbiography}

\begin{IEEEbiography}[{\includegraphics[width=1in,height=1.25in,clip,keepaspectratio]{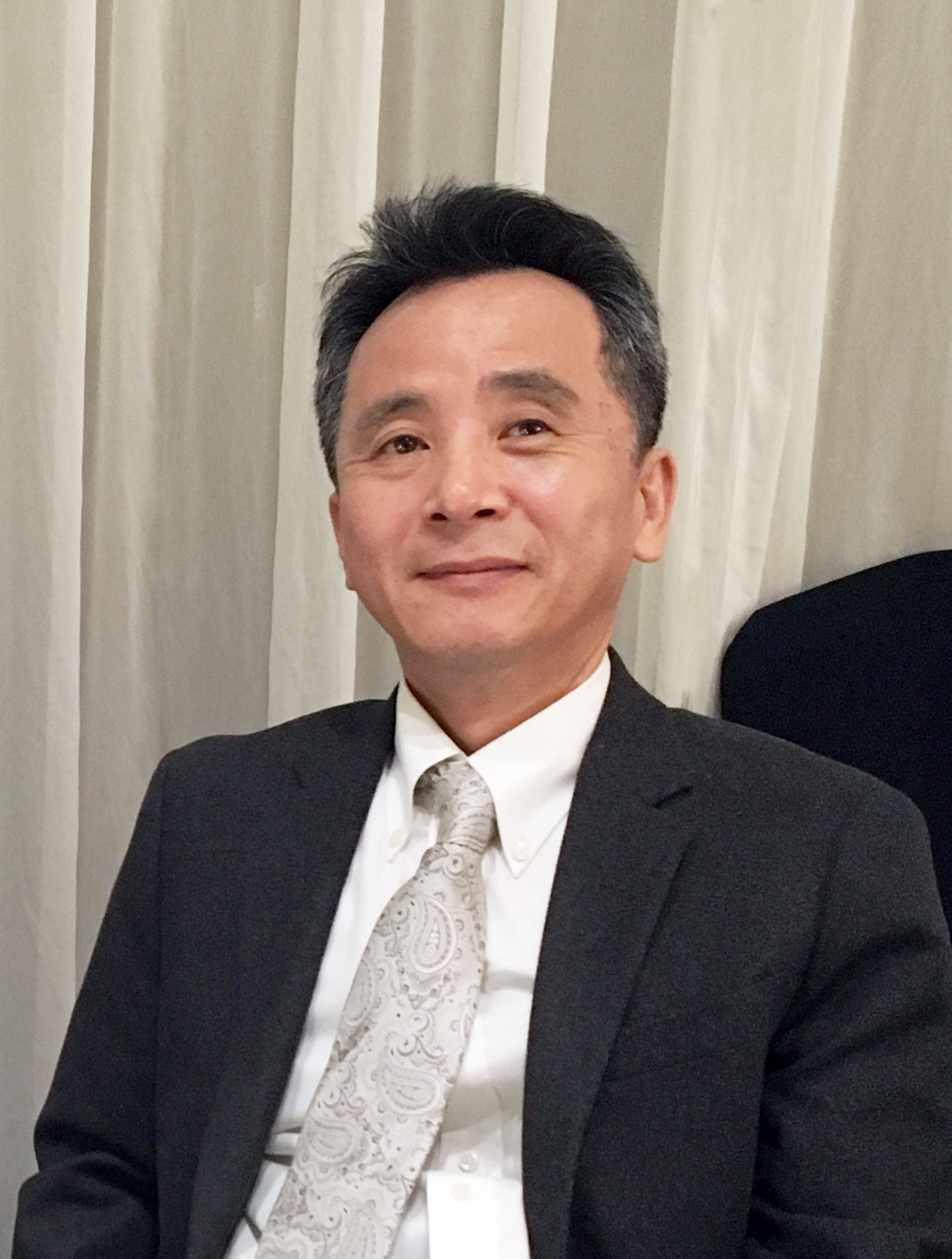}}] {Heesung Kwon}
is Senior Researcher and Team Lead at the DEVCOM Army Research Laboratory (ARL). He received the B.Sc. degree in Electronic Engineering from Sogang University, Seoul, Korea, in 1984, and the MS and Ph.D. degrees in Electrical Engineering from the State University of New York at Buffalo in 1995 and 1999, respectively. Dr. Kwon rejoined ARL in August 2007 as Team Lead and has been leading various AI/ML efforts pertaining to semantic scene understanding in resource-constrained environments, unmanned aerial systems (UAS)-based perception and action/activity recognition at the edge, etc., primarily leveraging machine learning-based approaches.
\end{IEEEbiography}

\begin{IEEEbiography}[{\includegraphics[width=1in,height=1.25in,clip,keepaspectratio]{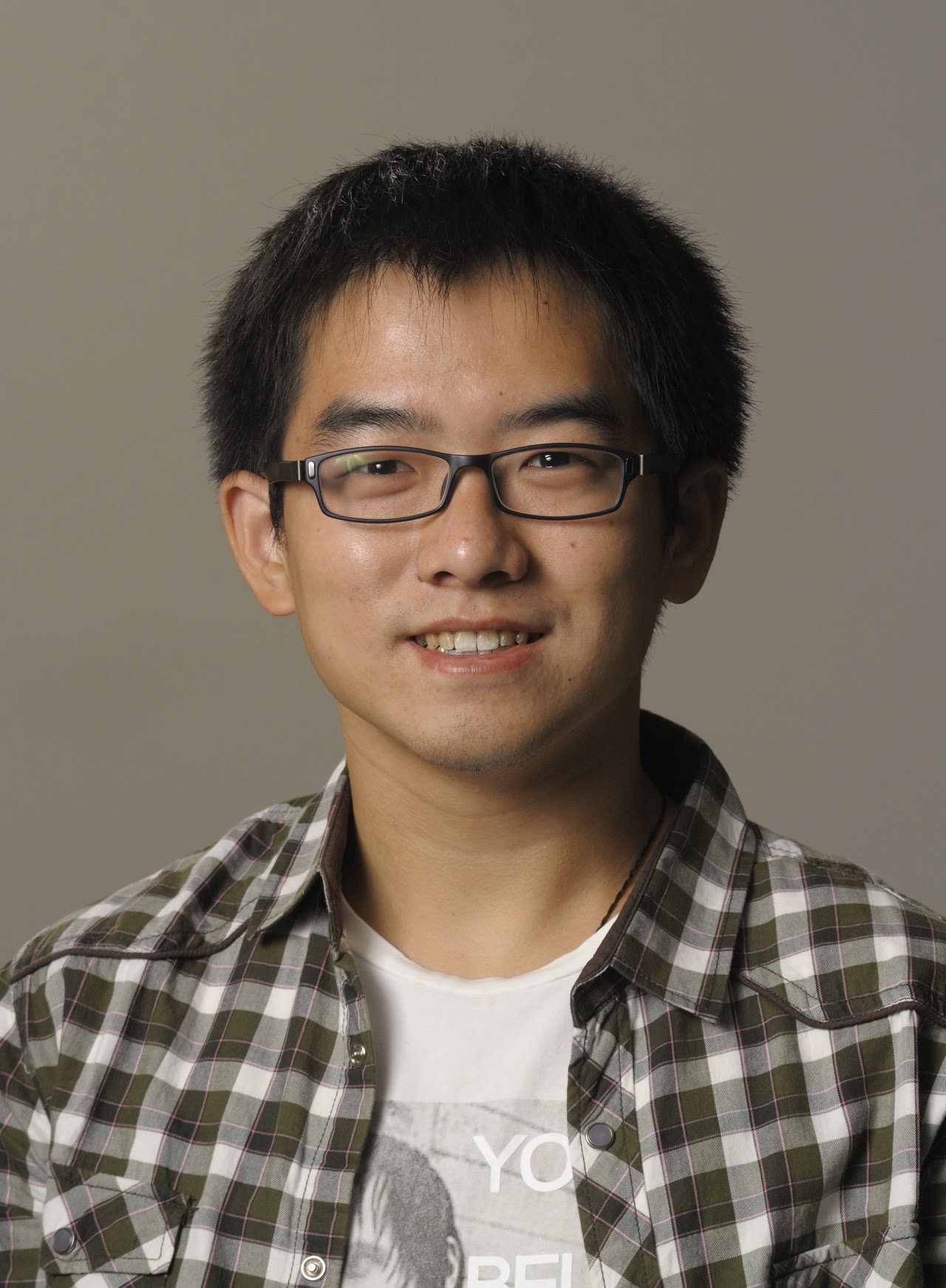}}]{Zhangyang Wang}
is currently an Assistant Professor of ECE at UT Austin. He received his Ph.D. in ECE from UIUC in 2016, and his B.E. in EEIS from USTC in 2012. Prof. Wang is broadly interested in the fields of machine learning, computer vision, optimization, and their interdisciplinary applications. His latest interests focus on automated machine learning (AutoML), learning-based optimization, machine learning robustness, and efficient deep learning.
\end{IEEEbiography}


%\vfill


\end{document}